\documentclass[11pt]{article}

\usepackage[final]{acl}

\usepackage{times}
\usepackage{latexsym}

\usepackage[T1]{fontenc}

\usepackage[utf8]{inputenc}

\usepackage{microtype}

\usepackage{inconsolata}

\usepackage{graphicx}

\usepackage{comment}
\usepackage{tcolorbox}
\usepackage{xspace}
\usepackage{hyperref}       
\usepackage{url}            
\usepackage{booktabs}       
\usepackage{tabularx}
\usepackage{amsfonts}       
\usepackage{nicefrac}       
\usepackage{microtype}      
\usepackage{pifont}
\usepackage{graphicx} 
\usepackage{multirow}
\usepackage{tikz}
\usepackage{pgfplots}
\usepackage{pgf-pie}
\usepackage[table]{xcolor}
\usetikzlibrary{pgfplots.groupplots, calc} 
\usepackage[capitalize,noabbrev]{cleveref} 

\pgfplotsset{compat=1.18} 
\usepackage{makecell}
\usepackage{pgfplotstable}
\usepackage{caption}
\usepackage{subcaption}
\usepackage{ragged2e}
\usepackage{longtable}
\usepackage{array}
\usepackage{tcolorbox}
\usepackage{needspace}
\tcbuselibrary{breakable}

\newcommand{\cmark}{{\color{green!60!black}\ding{51}}} 
\newcommand{\xmark}{{\color{red!70!black}\ding{55}}}   

\definecolor{single-yellow}{RGB}{240, 204, 162}    
\definecolor{single-yellow-border}{RGB}{200, 170, 130}   
\definecolor{global-green}{RGB}{189, 227, 214}    
\definecolor{global-green-border}{RGB}{150, 190, 180}   
\definecolor{multi-blue}{RGB}{151, 198, 247}    
\definecolor{multi-blue-border}{RGB}{115, 160, 210}   
\definecolor{c4}{RGB}{199, 197, 244}    
\definecolor{c4b}{RGB}{160, 150, 210}   
\definecolor{c5}{RGB}{245, 198, 214}    
\definecolor{c5b}{RGB}{200, 160, 175}   
\definecolor{c6}{RGB}{169, 226, 243}    
\definecolor{c6b}{RGB}{130, 185, 205}   
\definecolor{c7}{RGB}{222, 180, 235}    
\definecolor{c7b}{RGB}{180, 140, 195}   
\definecolor{gray}{RGB}{214, 204, 194}  
\definecolor{grayborder}{RGB}{180, 170, 160}  
\definecolor{event-color}{RGB}{212, 163, 115}
\definecolor{lightgrayrow}{gray}{0.96}

\newcommand{\acro}{MF$^2$\xspace}
\newcommand{\fact}{\textit{fact}}
\newcommand{\fib}{\textit{fib}}
\newcommand{\repo}{\url{https://github.com/deep-spin/MF2-bench-can-vlms-understand-movies}}

\title{How Well Do Vision-Language Models Understand Movies?}

\author{
  \textbf{Emmanouil Zaranis\textsuperscript{1,2}}\thanks{Equal contribution.},
  \textbf{António Farinhas\textsuperscript{18}*},
  \textbf{Saul Santos\textsuperscript{1,2}*},
  \textbf{Beatriz Canaverde\textsuperscript{1,2}*},
  \\
  \textbf{Miguel Moura Ramos\textsuperscript{1,2}*},
  \textbf{Aditya K Surikuchi\textsuperscript{3}},
  \textbf{André Viveiros\textsuperscript{1,2}},
  \textbf{Baohao Liao\textsuperscript{4}},
  \\
  \textbf{Elena Bueno-Benito\textsuperscript{5}},
  \textbf{Nithin Sivakumaran\textsuperscript{6}},
  \textbf{Pavlo Vasylenko\textsuperscript{1,2}},
  \textbf{Shoubin Yu\textsuperscript{6}},
  \\
  \textbf{Sonal Sannigrahi\textsuperscript{1,2}},
  \textbf{Wafaa Mohammed\textsuperscript{4}},
  \textbf{Ben Peters\textsuperscript{11}},
  \textbf{Danae Sánchez Villegas\textsuperscript{7}},
  \\
  \textbf{Elias Stengel-Eskin\textsuperscript{17}},
  \textbf{Giuseppe Attanasio\textsuperscript{2}},
  \textbf{Jaehong Yoon\textsuperscript{16}},
  \textbf{Stella Frank\textsuperscript{7,8}},
  \\
  \textbf{Alessandro Suglia\textsuperscript{9}},
  \textbf{Chrysoula Zerva\textsuperscript{2,13,19}},
  \textbf{Desmond Elliott\textsuperscript{7,8}},
  \textbf{Mariella Dimiccoli\textsuperscript{5,15}},
  \\
  \textbf{Mohit Bansal\textsuperscript{6}},
  \textbf{Oswald Lanz\textsuperscript{10,14}},
  \textbf{Raffaella Bernardi\textsuperscript{10,14}},
  \textbf{Raquel Fernández\textsuperscript{3,12}},
  \\
  \textbf{Sandro Pezzelle\textsuperscript{3,12}},
  \textbf{Vlad Niculae\textsuperscript{4,12}},
  \textbf{André F. T. Martins\textsuperscript{1,2,11,13}}
  \\[1ex]
  \textsuperscript{1}Instituto Superior Técnico, Universidade de Lisboa,
  \textsuperscript{2}Instituto de Telecomunicações,
  \\
  \textsuperscript{3}ILLC, University of Amsterdam,
  \textsuperscript{4}Language Technology Lab, University of Amsterdam,
  \\
  \textsuperscript{5}Institut de Robòtica i Informàtica Industrial, CSIC-UPC,
  \textsuperscript{6}UNC Chapel Hill,\\
  \textsuperscript{7}University of Copenhagen,
  \textsuperscript{8}Pioneer Center for AI,
  \textsuperscript{9}University of Edinburgh,\\
  \textsuperscript{10}Free University of Bozen-Bolzano,
  \textsuperscript{11}TransPerfect,
  \textsuperscript{12}ELLIS Unit Amsterdam,\\
  \textsuperscript{13}ELLIS Unit Lisbon,
  \textsuperscript{14}ELLIS Unit Trento,
  \textsuperscript{15}ELLIS Unit Barcelona,\\
  \textsuperscript{16}Nanyang Technological University,
  \textsuperscript{17}The University of Texas at Austin,
  \textsuperscript{18}Sword Health,\\
  \textsuperscript{19}National Technical University of Athens\\
  \small{
    \textbf{Correspondence:}
    \href{mailto:emmanouil.zaranis@tecnico.ulisboa.pt}{emmanouil.zaranis@tecnico.ulisboa.pt},
    \href{mailto:andre.t.martins@tecnico.ulisboa.pt}{andre.t.martins@tecnico.ulisboa.pt}
  }
}

\begin{document}
\maketitle
\begin{abstract}
Vision-language models (VLMs) can increasingly handle extended contexts and answer questions about long videos---but can they understand the narrative structure of movies as humans do? To investigate this question, we introduce {MF}$^2$ (\textit{Movie Facts and Fibs}), a fully manually curated benchmark designed to evaluate whether models can identify, recall, and comprehend information central to the movie's storyline, using both the visual and textual modalities. {MF}$^2$ includes over 50 \textit{open-licensed} movies, each paired with manually constructed sets of claim pairs: one true claim (\textit{fact}) and one plausible but false (\textit{fib}), totaling over 850 pairs. Our analysis across 20 open-source and 2 proprietary models reveals a substantial gap relative to human performance. While subtitles provide a strong signal for narrative understanding, other textual sources (e.g., synopses) are less informative. In contrast, when visual information is distributed across the entire movie and jointly embedded with subtitles, models struggle to identify and effectively use the relevant content. However, when the visual input is restricted to claim-relevant visual content, models improve significantly, indicating substantial room for improvement in processing large multimodal contexts and distinguishing relevant from irrelevant information. Overall, our findings highlight persistent challenges in multimodal narrative understanding for current VLMs. We release our data and code to support further research in this direction.
\end{abstract}

\section{Introduction}
The past few years have seen remarkable advances in vision-language models (VLMs), particularly in their capacity to process extended multimodal context and perform question answering across long video streams~\cite{team2023gemini,llava_video,Qwen3-VL,internvl3}. Yet their ability to understand movie narratives in a human-like manner remains underexplored. This is an important gap, as narratives provide a natural testbed for studying the long-context capabilities of current models: storylines are shaped by how events, entities, emotional arcs, and causal interactions between characters unfold over time. As a result, successful understanding goes beyond local perception or step-by-step reasoning, and instead depends on integrating information across extended spans, while maintaining coherence over evolving relationships between characters and events.

In textual domains, 
literal and narrative settings have increasingly emerged as an important aspect of long-context evaluation 
~\citep{hudson-al-moubayed-2022-muld,an-etal-2024-l,piper-bagga-2024-using,karpinska-etal-2024-one,hamilton-etal-2026-narrabench}. 
Similar extensions have begun to arise in multimodal settings, particularly focusing on videos, with recent benchmarks shifting towards longer content~\citep{,ataallah2024infinibenchcomprehensivebenchmarklarge,wang2025lvbench,fu2025videomme,wu2024longvideobench}. However, the majority of these efforts rely on general-purpose video content (e.g., egocentric recordings, instructional videos, or vlogs), which, while suitable for assessing perception and reasoning capabilities over extended sequences, are not explicitly designed to capture narrative structure and  assess the relationships that shape a storyline. A smaller subset of works instead leverages narrative-rich content, such as movies or TV shows, aligning better with the assessment of storyline development~\citep{huang2020movienet,song2023moviechat,song2024moviechat+,ataallah2024infinibenchcomprehensivebenchmarklarge}. However, these are often constrained by \textit{limited temporal coverage} or \textit{restricted accessibility due to copyright considerations}, often providing only partial spans of the original content. Moreover, some rely on automatically generated questions~\cite{ataallah2024infinibenchcomprehensivebenchmarklarge,chandrasegaran2024hourvideo,faure-etal-2025-moviecore}, which evaluate content uniformly and may place disproportionate emphasis on non-narrative details rather than on central storyline events.

\begin{figure*}[ht]
\centering
\includegraphics[width=\textwidth]{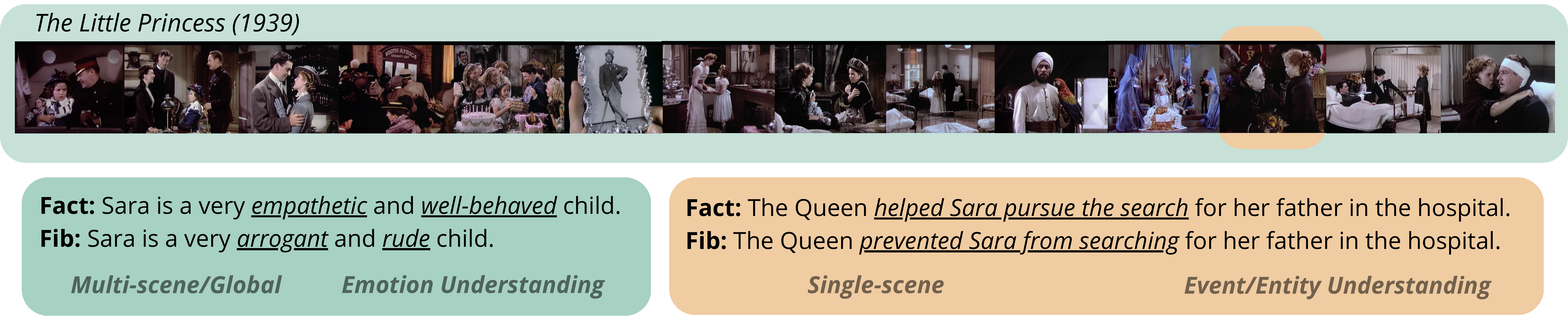}
\caption{Illustration of two claim pairs (each with a \textit{fact} and a \textit{fib}) from the movie ``The Little Princess'' (1939). Our claims target memorable events, focusing on key points of the narrative such as emotional arcs and causal relationships between characters, and require reasoning across different granularities (\textcolor{single-yellow-border}{single-scene} and \textcolor{global-green-border}{multi-scene/global}).} 
\label{fig:example-movie-claims}
\end{figure*}

In this work, we investigate the extent to which VLMs can understand and reason over movie narratives. Building on prior research in narrative understanding across textual and multimodal domains~\citep{papalampidi-etal-2019-turning-points,karpinska-etal-2024-one,hamilton-etal-2026-narrabench}, we focus on whether models can answer questions related to \textit{core narrative elements}, which shape the narrative trajectory~\citep{papalampidi-etal-2019-turning-points,papalampidi-etal-2020-screenplay}.  Rather than assessing models uniformly across all parts of the content, as is typically done with automatically generated questions, our analysis targets events central to the storyline that are retained in \textit{human memory}, namely those that viewers can \textit{recall without rewatching the video due to narrative importance}. 

To support this analysis, we construct \acro (\textit{Movie Facts and Fibs}), a \textit{manually annotated} evaluation set of 868 claim pairs grounded in 53 open-licensed movies. Each pair consists of a true statement (a \fact) and a plausible but false counterpart (a \fib) (see~\cref{fig:example-movie-claims} and Examples in Appendix~\ref{app:examples}). Following prior work on narrative understanding~\citep{karpinska-etal-2024-one}, we adopt a claim-based evaluation protocol instead of multiple-choice questions. This design is motivated by two considerations: (i) it enables isolated statement-level truth assessment without reliance on contrastive options and (ii) it reduces biases from poorly constructed distractors in multiple-choice setups \citep{li2024anchored_gpt2_mcq_positional_bias,loginova2024_addressing_blind_guessing_selection_bias,singh2025_too_many_options_pitfalls_mcq,molfese2025_right_answer_wrong_score_multiple_choice}.

Our analysis across 20 open-source and 2 proprietary models reveals several key findings: (i) models substantially underperform humans on the task, with the best-performing model lagging behind human performance by 20.3\% absolute; (ii) subtitles help models reason better about narratives, while other text forms (e.g., synopses) are not equally informative; and (iii) although claim-relevant visual evidence provides a strong signal when directly provided as input, models struggle to effectively retrieve and utilize it when it is embedded within full-movie  context. Together, these findings highlight an important challenge for current VLMs: identifying and reasoning over the relevant information in long-form movies, while ignoring irrelevant details. As part of our contribution to support reproducibility and future research, we release \acro's data and code under a CC-BY 4.0 license, including raw video content to mitigate link rot and ensure long-term accessibility.\footnote{\repo}

\section{Dataset Construction} \label{sec:mf2_dataset_construction}

\begin{figure*}
\centering
\includegraphics[width=1.0\textwidth]{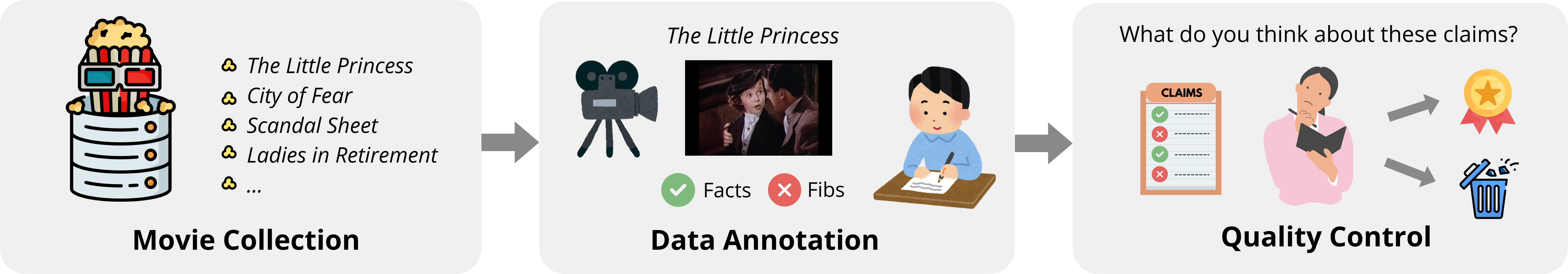}
\caption{Dataset construction process involving three main stages: movie collection, data annotation, and quality control.}
\label{fig:pipeline}
\end{figure*}

\acro includes 53 full-length, open-licensed movies, each accompanied by subtitles, and 868 human-authored contrastive claim pairs.
Each pair tests whether a model can distinguish true from false information based on its understanding of the story.
\cref{fig:example-movie-claims} shows some examples (see more in Appendix~\ref{app:examples}).
We now describe the dataset construction process in detail, covering movie selection (\cref{subsec:movie_selection}), the annotation protocol (including claim categorization and reasoning granularities; \cref{subsec:data_annotation}), and human quality control procedures used to filter ambiguous or low-quality claims (\cref{subsec:quality_control}). \cref{fig:pipeline} illustrates these three stages.

\subsection{Movie Selection and Subtitles}\label{subsec:movie_selection}

We started by collecting a pool of movies from the Internet Archive,\footnote{\url{https://archive.org}} an online repository of open-licensed media. We specifically selected titles released under the Public Domain 1.0 license to ensure legal reusability and support open-access research. We focused on older films, prioritizing those with limited online visibility, measured by the number of user reviews on IMDb, to reduce the risk of data contamination.
We sourced original-language subtitles---the majority of which are in  English---from OpenSubtitles.org,\footnote{\url{https://www.opensubtitles.org}} a widely used platform that provides subtitles for a large collection of movies, TV shows, and other video content. This process yielded a final collection of 53 full-length movies with an average duration of 88.33 minutes, each accompanied by audio and aligned subtitles. The collection spans a wide range of genres as we detail in Appendix~\ref{app:details_dataset}.

\subsection{Data Annotation}\label{subsec:data_annotation}
In contrast to approaches that rely on semi-automatically generated questions, often produced by language models and potentially reflecting model-specific biases, we opted for fully manual annotations, \textit{prioritizing quality over quantity}.
The annotation process involved 26 annotators  (see Appendix~\ref{app:details_dataset} for details) who watched the full movies, identified key narrative elements, and constructed pairs of contrastive claims: one factually correct statement (\fact) and one minimally altered, false counterpart (\fib). Following~\citet{karpinska-etal-2024-one}, annotators were instructed to minimize lexical differences between the \textit{fact} and the \textit{fib}, changing only the parts needed to flip the truth value. The data annotation guidelines are presented in Appendix~\ref{appendix:guidelines_human}.


\paragraph{Claim granularity.}
To capture different levels of reasoning, annotators labeled each \textit{fact} according to the granularity required to verify its truth: \textit{(i)} \textit{single-scene}: answerable using information from one scene; or \textit{(ii)} \textit{multi-scene/global}: requiring integration of information across multiple scenes or higher-level understanding that may not be easily attributable to specific scenes, including cases involving accumulated or inferred information across the movie.
The dataset exhibits a balanced distribution of single-scene and multi-scene/global facts, comprising 49.4\% and 50.6\%, respectively. 

\paragraph{Comprehension dimensions.}
In addition to the reasoning granularity, annotators labeled each claim pair with one or more comprehension dimensions, indicating the specific aspects of narrative understanding being tested. These dimensions, informed by prior work \citep{xiao2021next,zhang2023movqabenchmarkversatilequestionanswering,wang2025lvbench}, include event/entity understanding, temporal perception, emotion understanding, and causal reasoning. We provide detailed definitions of each dimension and their distribution in Appendix~\ref{app:details_dataset}.

\paragraph{Grounding Timeframes.}
For a subset of single-scene claim pairs (12\% of the total pairs), annotators were instructed to mark the visual segment (timespan) of the movie required to correctly answer the claim pair. This provides a basis for evaluating models’ use of relevant visual content.

\subsection{Quality Control}\label{subsec:quality_control}

To ensure dataset quality and support the identification of problematic claims, we include a quality-control stage in which annotators answer existing claims and provide constructive feedback. Feedback focused on claims that were ambiguous, poorly phrased, open to interpretation, or too fine-grained to be meaningfully tied to narrative understanding (e.g., needle-in-a-haystack claims). Specifically, annotators first selected a subset of movies they had not previously seen during the data annotation stage. After watching a movie,  they classified the corresponding claims as either true or false using a custom annotation interface (see Appendix~\ref{appendix:guidelines_human} for an example and full guidelines). Claims were presented one at a time, and annotators were required to respond based solely on memory, without rewatching the movie or using external tools or notes. Annotators were encouraged to leave comments whenever a claim appeared ambiguous, poorly phrased, open to interpretation, or excessively fine-grained. After this stage, annotators could optionally revisit their earlier responses with access to the movie in order to provide additional validation comments.

Finally, two annotators reviewed all collected comments from the stages described above. Based on the provided feedback, and after discussion to reach consensus, they identified problematic claims and removed them from the dataset. Importantly, no claims were rewritten during this process; they were either accepted or discarded. This filtering stage resulted in the removal of 104 pairs of claims, yielding a cleaner set of 868 high-quality pairs.

\section{Experimental Setup}
\label{sec:experimental_setup}

We provide an overview of the models included in our study, as well as important information regarding the input modalities used.

\paragraph{Input Modalities.}
We evaluate all models in a vision-language setting, where visual input is provided through sampled movie frames. In addition, we investigate the effect of language by providing subtitles either independently or jointly with the visual input.\footnote{Subtitles are appended to the visual input.} Importantly, models always receive frames sampled \textit{uniformly across entire movie}, rather than from targeted scenes. Consequently, models must process the full provided content to identify the relevant information and determine whether a claim is true or false.

\paragraph{Models.} 
Our evaluation suite covers 2 proprietary and 20 open-weight VLMs. Among the proprietary models, we include GPT-4o \citep{openai2024gpt4ocard}, Gemini 2.5 Pro \citep{team2023gemini}. We cover several open-weight VLM families, including  QwenVL~\citep{Qwen2.5-VL,Qwen3-VL}, InternVL~\citep{internvl3,wang2025internvl35advancingopensourcemultimodal}, Ovis2~\citep{Ovis}, Gemma3~\citep{gemmateam2025gemma3technicalreport}, GLM-4.5V~\citep{vteam2026glm45vglm41vthinkingversatilemultimodal}, among others. We additionally include models specialized for long videos, such as LongVILA-R1~\citep{chen2025longvila-r1} and MiMo-VL-7B-RL~\citep{coreteam2025mimovltechnicalreport}. We include all prompt, frame sampling and answer parsing details in Appendix~\ref{app:experimental_settings} for reproducibility. 

\paragraph{Human baseline.} 
To establish a human baseline for comparison, 24 annotators evaluated the full set of constructed claim pairs, in contrast to prior work that typically relies on smaller subsets. Annotators selected movies they had not previously seen during the data annotation stage. After watching a movie, with optional access to subtitles, they classified the corresponding claims as either true or false using the same interface employed during quality control (see Appendix~\ref{appendix:guidelines_human} for an example and full guidelines). Each movie was evaluated by a single annotator, claims were presented one at a time, and annotators were required to answer based solely on memory, without rewatching the movie, using external tools, or taking notes.\footnote{Due to the labor-intensive nature of the process, we do not repeat the experiment under alternative input conditions (e.g., video-only), and instead use this setting as a \textit{reference estimate of human performance} for movie narrative understanding.}

\paragraph{Evaluation protocol.} 
We report two metrics: (i) pairwise accuracy, which measures how often models correctly classify both the true and the false claim in a pair (i.e., credit is given only if both are labeled correctly; no points are awarded for partial correctness); and (ii) standard accuracy, which is computed over individual claims. The random baselines are 25\% and 50\%, respectively. 
Following prior work \citep{karpinska-etal-2024-one}, \textit{models answer each claim independently, without access to the paired structure during prediction}. Pairwise accuracy is computed post-hoc by grouping predictions from the same pair. This particular design (i) allows for assessing isolated statement-level reasoning and truth prediction, and does not involve contrastive reasoning between paired alternatives, (ii) reduces biases caused by poorly constructed distractor options in multiple-choice questions \citep{loginova2024_addressing_blind_guessing_selection_bias,singh2025_too_many_options_pitfalls_mcq,molfese2025_right_answer_wrong_score_multiple_choice}, and (iii) avoids the use of automatic judges, which may struggle with rich long-form video \citep{bavaresco2024llms,liu2025_is_your_vlm_reliable_judge,ye2025justice}. 
In Appendix~\ref{app:paiwise_acc_vs_acc}, we discuss this design choice and the use of pairwise accuracy as our primary metric.


\section{Empirical Findings}

\begin{table*}[t]
\centering
\small
\renewcommand{\arraystretch}{1.05}

\begin{tabular*}{\textwidth}{@{\extracolsep{\fill}}l c r r r r r r}
\toprule

\multirow{2}{*}{\textbf{Method}} &
\multirow{2}{*}{\textbf{\#Params}} &
\multicolumn{3}{c}{\textbf{Pairwise Accuracy (\%)}} &
\multicolumn{3}{c}{\textbf{Accuracy (\%)}} \\
\cmidrule(lr){3-5} \cmidrule(lr){6-8}
 &  & Subs & Video & Video+Subs & Subs & Video & Video+Subs \\
\midrule

\multicolumn{8}{c}{\textit{\textbf{Baselines}}} \\
\midrule
Random & - & 25.0 & 25.0 & 25.0 & 50.0 & 50.0 & 50.0 \\
Human  & - & -- & -- & 84.1 & -- & -- & 90.5 \\
\midrule

\multicolumn{8}{c}{\textit{\textbf{Closed Models}}} \\
\midrule
GPT-4o & - & 48.8 & 19.5 & 49.8  & 72.9 & 55.5 & 73.2 \\
Gemini 2.5 Pro & - & \textbf{\underline{58.9}} & \textbf{\underline{37.9}} & \textbf{\underline{63.8}} & \textbf{\underline{77.5}} & \textbf{\underline{64.4}} & \textbf{\underline{79.8}} \\
\midrule

\multicolumn{8}{c}{\textit{\textbf{Open-weight Models}}} \\
\midrule
mPLUG-Owl3*  & 7B & -- & -- & -- & -- & -- & --
\\
VideoLLaMA3 & 7B & 36.3 & 21.1 & 36.1 & 64.0 & 56.8 & 64.2 \\
Qwen2.5-VL  & 7B & 31.7 & 24.5 & 32.8 & 62.1 & 56.7 & 62.0 \\
LLaVA-Video & 7B & 20.5 & 6.5 & 19.1 & 58.2 & 51.4 &  57.9 \\
LongVILA-R1 & 7B & 9.8 & 11.5 & 16.9 & 54.0 & 50.1 & 56.6 \\
MiMo-VL-7B-RL & 7B & 36.4 & 25.8 & 38.6 & 62.1 & 57.5 & 65.2 \\
Kangaroo*  & 8B & -- & -- & -- & -- & -- & -- \\
InternVL3.5  & 8B & 39.2 & 18.9 & 37.2 & 65.4 & 55.1 & 64.5 \\
LLaVAOneVision1.5  & 8B & 37.4 & 4.9 & 26.4 & 65.2 & 51.0 & 60.9 \\
Molmo2  & 8B & 41.0 & 28.3 & 43.0 & 67.2 & 59.0 & 67.7 \\
Phi4* & 14B & 26.8 & -- & -- & 58.8 & -- & -- \\
Aria & 25B (3.9B) & 36.2 & 2.5 & 33.4 & 64.1 & 50.0 & 63.2 \\
Gemma3 & 27B & 42.3 & \underline{31.5} & 39.7 & 67.8 & \underline{61.2} & 65.3  \\
Qwen3VL-IT  & 30B & 33.5 & 19.2 & 43.8 & 64.7 & 56.9 & 69.5 \\
Ovis2 & 34B & 47.2 & 20.7 & 48.5 & 70.3 & 56.5 & 71.1\\
InternVL3.5 & 38B & 47.5 & 26.6 & 46.2 & 70.8 & 60.0 & 70.3  \\
Qwen2.5-VL & 72B & 37.6 & 29.7 & 47.5 & 66.8 & 58.8 & 69.0 \\
LLaVA-Video & 72B & 41.7 & 15.6 & 41.8 & 68.8 & 54.6 & 69.1  \\
InternVL3 & 78B & \underline{49.0} & 22.1 & \underline{51.3} & \underline{72.0} & 58.0 & \underline{72.7} \\
GLM-4.5V-FP8  & 108B & 38.4 & 26.6 & 31.4 & 64.7 & 55.0 & 62.4 \\

\bottomrule
\end{tabular*}
\caption{Performance (\%) of open-weight and closed models on \acro under different input modalities. Best values overall are \textbf{bolded}; best per group are \underline{underlined}. Asterisk (*) denotes models that failed to follow the instruction prompt in a substantial fraction of samples and their results are therefore not reported. Numbers in parentheses indicate the number of active parameters.}
\label{tab:main_results}
\end{table*}

We now present our empirical findings, starting with the main comparison between models and humans and an analysis across input modalities (Section~\ref{subsec:main_resutls}). This is followed by a more detailed study of Gemini 2.5 Pro, the best-performing model among those evaluated, in Section~\ref{subsec:gemini25}.

\subsection{The Role of Language and Vision}\label{subsec:main_resutls}
In this section, we analyze model performance in comparison to human performance, as reported in Table~\ref{tab:main_results}, and examine how different input modalities (subtitles-only, video-only, and video with subtitles) affect performance across open-weight and closed models, identifying several key challenges for VLMs in movie narrative understanding.


\paragraph{Models perform significantly worse than humans.}
Among closed models, Gemini 2.5 Pro achieves the highest scores, with a pairwise accuracy of $63.8$\%, followed by  InternVL3-72B, which performs $12.5\%$ lower on video and subtitles. Despite their fair performance, \textit{both models rank significantly behind humans}. Among small-scale models ($\leq$9B), Molmo2 achieves the highest performance, outperforming the random baseline by 18\% in the video with subtitles setting, while Ovis2 performs best in the medium-scale category (25B–38B). These results underscore the difficulty of the task for current models, while highlighting humans' superior ability to retain, integrate, and reason over critical narrative content.

\paragraph{Subtitles provide a stronger narrative signal than visual information alone.}
Across nearly all models, performance with subtitles only is remarkably higher than when relying exclusively on visual information sampled across the entire movie. Since subtitles play a central role in constructing the storyline through dialogue, character interactions, and explicit descriptions of events and actions, this result confirms that such textual signals provide a particularly strong source of narrative information for the task.

\paragraph{Coupling subtitles with visual information yields no consistent benefits over subtitles alone for most models.}
One might expect visual information to provide complementary signals to subtitles, leading to substantial gains when both modalities are available. However, as shown in Table~\ref{tab:main_results}, adding visual information yields no consistent improvements across the evaluated models. This suggests that either visual cues are not consistently helpful, or that models do not effectively identify and attend to the relevant visual information when embedded in long multimodal contexts. In the following paragraph, we further examine this hypothesis by investigating whether visual information becomes beneficial when the relevant visual content is explicitly provided to the models in isolation.

\paragraph{Explicit visual cues provide a useful signal.} We evaluate whether performance changes when the visual input is restricted to the parts of the movie that are directly relevant to each claim. Specifically, we use a subset of single-scene claims with annotated time spans identifying the relevant visual spans (see Section~\ref{subsec:data_annotation}). We extract targeted frames from these spans instead of sampling uniformly from the full movie. We evaluate Gemini 2.5 Pro alongside five mid and large scale open-weight models. In Table~\ref{tab:cropped_results}, we report performance on this subset under subtitles-only, video-only, and video-with-subtitles settings, comparing results between full-movie and targeted-frame inputs. Across most models, performance improves when given targeted visual content, with and without subtitles.\footnote{Qwen2.5-VL-72B is the only exception, showing a 4.9-point drop in the video-only setting. This behavior appears to be associated with a more conservative prediction tendency, with the model more frequently rejecting both claims; a similar bias toward rejecting claims is also observed across other models (see Appendix~\ref{app:paiwise_acc_vs_acc}). This tendency disappears when subtitles are provided, suggesting that textual information helps interpret the visual cues.} The consistent improvements observed with targeted visual content demonstrate that \textit{visual cues are indeed useful when explicitly provided to the model}. Therefore, the observed behavior likely stems from difficulties in retrieving and effectively integrating relevant visual information with textual cues in long multimodal contexts, highlighting substantial room for improvement in these aspects.
\definecolor{DarkGreen}{RGB}{0,100,0}   
\definecolor{DarkRed}{RGB}{139,0,0}     

\begin{table*}[t]
\centering
\small
\begin{tabular}{lrlrl}
\toprule
\textbf{Model} &
\multicolumn{2}{c}{\textbf{Video-only}} &
\multicolumn{2}{c}{\textbf{Video+Subs}} \\
\cmidrule(lr){2-3} \cmidrule(lr){4-5}
&
\textbf{(full)} &
\textbf{(targeted)} &
\textbf{(full)} &
\textbf{(targeted)} \\
\midrule

Gemini 2.5 Pro &
38.6 &
49.4 {\color{DarkGreen}(+10.8)} &
59.4 &
74.0 {\color{DarkGreen}(+14.6)} \\

Gemma3-27B &
27.2 &
36.9 {\color{DarkGreen}(+9.7)} &
42.7 &
49.5 {\color{DarkGreen}(+6.8)} \\

Qwen3VL-30B &
12.6 &
22.3 {\color{DarkGreen}(+9.7)} &
40.8 &
52.4 {\color{DarkGreen}(+11.6)} \\

InternVL3.5-38B &
29.1 &
41.7 {\color{DarkGreen}(+12.6)} &
51.5 &
53.4 {\color{DarkGreen}(+1.9)} \\

Qwen2.5VL-72B &
34.0 &
29.1 {\color{DarkRed}(-4.9)} &
46.6 &
53.4 {\color{DarkGreen}(+6.8)} \\

InternVL3-78B &
27.2 &
40.8 {\color{DarkGreen}(+13.6)} &
49.5 &
66.0 {\color{DarkGreen}(+16.5)} \\

\bottomrule
\end{tabular}

\caption{Pairwise accuracy when models are provided with ``targeted'' frames as visual input versus frames sampled uniformly from the entire movie. ``Targeted'' frames refer to frames extracted from the annotated time intervals that contain the visual content relevant to each claim. Results are reported on a single-scene claim subset under video-only and video with subtitles settings. Values in parentheses indicate the performance difference relative to the full-movie setting. {\color{DarkGreen}Positive} and {\color{DarkRed}negative} deltas denote performance gains and drops, respectively.}
\label{tab:cropped_results}
\end{table*}

\paragraph{The key challenge:} Overall, our results suggest that models substantially lag behind human performance in movie narrative understanding. Beyond this gap, (i) subtitles provide a strong signal that significantly helps reasoning in narrative understanding, and (ii) the relevant visual cues are nevertheless informative when directly provided as input, but are less effectively utilized in long-context settings (as also illustrated by the qualitative Examples 5 and 6, in Appendix~\ref{app:examples}). Taken together, these findings point to a central challenge for current VLMs: effectively identifying, retrieving, and integrating relevant visual evidence with textual information within extended multimodal contexts.
Interestingly, our results align well with prior findings from core “vision-centric” tasks, where vision-language models have been shown to rely heavily on language priors and textual inputs \citep{fu2025hidden,zheng2025mllmsdeeplyaffectedmodality}.

\subsection{A Closer Look at Gemini 2.5 Pro}\label{subsec:gemini25}
\begin{table*}[t]
\centering
\small
\begin{tabularx}{\textwidth}{
  >{\centering\arraybackslash}X
  >{\centering\arraybackslash}X
  >{\centering\arraybackslash}X
  >{\centering\arraybackslash}X
  >{\centering\arraybackslash}X
  >{\centering\arraybackslash}X
}
\toprule
 & \multicolumn{5}{c}{\textbf{Input Modality}} \\
\cmidrule(lr){2-6}
\textbf{Metric} & \textbf{Video} & \textbf{Subs} & \textbf{Video+Subs} & \textbf{Synopsis} & \textbf{Movie Title} \\
\midrule
Pairwise Acc. & 37.9 & 58.9 & 63.8 & 27.9 & 45.4 \\
\bottomrule
\end{tabularx}
\caption{Performance of Gemini 2.5 Pro across different input modalities. \textit{Video} uses only the visual stream; \textit{Subs} includes only subtitle information; \textit{Synopsis} relies only on the synopsis of the movie obtained from Wikipedia; \textit{Video+Subs} combines both visual and subtitles inputs; and \textit{Movie Title} uses only the claim, along with the movie title and release year, without access to movie content.}
\label{tab:gemini_ablation}
\end{table*}

We now turn to Gemini 2.5 Pro, the best-performing model, which we use as a representative  model to examine its reliance on parametric (world) knowledge and how different forms of textual content (i.e., synopses) affect performance. Lastly, we analyze its behavior across reasoning granularities and comprehension dimensions.

\paragraph{Beyond vision and subtitles: the role of summarized content (synopsis) and world knowledge.}
Table~\ref{tab:gemini_ablation} presents an ablation study examining reliance on textual context beyond subtitles. When provided only with the movie title and release year, the model achieves above-random performance, suggesting an influence of parametric (pretrained) knowledge (e.g., awareness of well-known films and their plots). In contrast, performance drops to near-random levels with only the movie synopsis, indicating that compressed narrative descriptions provide little useful signal for reasoning on the task (a trend that also holds for large-scale open-weight models; see Appendix~\ref{appendix:ablation_studies}). Importantly, both settings remain substantially below human performance, with absolute gaps of 38.7\% and 56.2\%, respectively, suggesting that \textit{neither memorization nor high-level summarization is sufficient to solve the task}.
\paragraph{Input modality contributions across comprehension dimensions and reasoning granularities.}
In Figure~\ref{Fig:modalities_cat_gemini}, we present ablation studies examining how different input modalities contribute to performance across comprehension dimensions and reasoning granularities. We observe that the presence of visual information particularly benefits claims involving emotional understanding, with a noticeable gap relative to other comprehension dimensions, especially in the video-with-subtitles setting (a trend that also holds for large-scale open-weight models as well-see Appendix~\ref{appendix:ablation_studies}). In contrast, causal reasoning is notably weaker under video-only conditions, likely due to the need for linguistic disambiguation. This limitation becomes evident when subtitles are introduced: the largest improvement is observed for causal reasoning, highlighting the complementary role of textual context. Beyond comprehension dimensions, reasoning performance remains relatively consistent across reasoning granularities, although single-scene reasoning appears slightly more challenging under the video-only setting.


\definecolor{c1}{RGB}{240, 204, 162}    
\definecolor{c1b}{RGB}{200, 170, 130}   
\definecolor{c2}{RGB}{189, 227, 214}    
\definecolor{c2b}{RGB}{150, 190, 180}   
\definecolor{c3}{RGB}{151, 198, 247}    
\definecolor{c3b}{RGB}{115, 160, 210}   
\definecolor{c4}{RGB}{199, 197, 244}    
\definecolor{c4b}{RGB}{160, 150, 210}   
\definecolor{c5}{RGB}{245, 198, 214}    
\definecolor{c5b}{RGB}{200, 160, 175}   
\definecolor{c6}{RGB}{169, 226, 243}    
\definecolor{c6b}{RGB}{130, 185, 205}   
\definecolor{c7}{RGB}{222, 180, 235}    
\definecolor{c7b}{RGB}{180, 140, 195}   

\begin{figure*}
\centering
\resizebox{0.97\textwidth}{!}{%
\begin{tikzpicture}[baseline]

\begin{groupplot}[
    group style={
        group name=barplotsA,
        group size=3 by 1,
        horizontal sep=0.35cm,
        xlabels at=edge bottom
    },
    width=3.2cm,
    height=2.9cm,
    legend style={
        at={(0.05,-0.25)},
        anchor=north,
        legend columns=4,
        column sep=3pt,
        font=\scriptsize
    },
    ybar,
    ymin=0, ymax=72,
    enlarge x limits=19,
    axis lines=none,
    xtick=data,
    symbolic x coords={ModA, ModB, ModC, ModD, ModE, ModF, ModG, ModH},
    xticklabels={},
    title style={yshift=2mm, font=\scriptsize, text depth=0.5ex},
    nodes near coords,
    nodes near coords align={vertical},
]

\nextgroupplot[title={Video}]
\addplot+[fill=c5, draw=c5b, bar width=9.5pt, nodes near coords,
    point meta=explicit symbolic, font=\scriptsize, text=black]
    coordinates {(ModA, 37.84) [38]};
\addplot+[fill=c6, draw=c6b, bar width=9.5pt, nodes near coords,
    point meta=explicit symbolic, font=\scriptsize, text=black]
    coordinates {(ModB,40.35) [40]};
\addplot+[fill=c7, draw=c7b, bar width=9.5pt, nodes near coords,
    point meta=explicit symbolic, text=black, font=\scriptsize]
    coordinates {(ModC,31.57) [32]};
\addplot+[fill=c4, draw=c4b, bar width=9.5pt, nodes near coords,
    point meta=explicit symbolic, text=black, font=\scriptsize]
    coordinates {(ModD,43.63) [44]};

\nextgroupplot[title={Subs},
    legend style={at={(0.5,-0.1)}, anchor=north, legend columns=4, font=\scriptsize}]
\addplot+[fill=c5, draw=c5b, bar width=9.5pt, nodes near coords,
    point meta=explicit symbolic, font=\scriptsize, text=black]
    coordinates {(ModA, 58.04) [58]};
\addlegendentry{Event/Entity}
\addplot+[fill=c6, draw=c6b, bar width=9.5pt, nodes near coords,
    point meta=explicit symbolic, font=\scriptsize, text=black]
    coordinates {(ModB,69.29) [69]};
\addlegendentry{Temporal}
\addplot+[fill=c7, draw=c7b, bar width=9.5pt, nodes near coords,
    point meta=explicit symbolic, text=black, font=\scriptsize]
    coordinates {(ModC,60.76) [61]};
\addlegendentry{Causal}
\addplot+[fill=c4, draw=c4b, bar width=9.5pt, nodes near coords,
    point meta=explicit symbolic, text=black, font=\scriptsize]
    coordinates {(ModD,60.9) [61]};
\addlegendentry{Emotion}

\nextgroupplot[title={Video w/ Subs}]
\addplot+[fill=c5, draw=c5b, bar width=9.5pt, nodes near coords,
    point meta=explicit symbolic, font=\scriptsize, text=black]
    coordinates {(ModA, 62.8) [63]};
\addplot+[fill=c6, draw=c6b, bar width=9.5pt, nodes near coords,
    point meta=explicit symbolic, font=\scriptsize, text=black]
    coordinates {(ModB,64.03) [64]};
\addplot+[fill=c7, draw=c7b, bar width=9.5pt, nodes near coords,
    point meta=explicit symbolic, text=black, font=\scriptsize]
    coordinates {(ModC,63.15) [63]};
\addplot+[fill=c4, draw=c4b, bar width=9.5pt, nodes near coords,
    point meta=explicit symbolic, text=black, font=\scriptsize]
    coordinates {(ModD,70.9) [71]};

\end{groupplot}

\node[font=\scriptsize, above=0.85cm] at
    ($(barplotsA c1r1.north west)!0.5!(barplotsA c3r1.north east)$)
    {Comprehension Dimension};

\end{tikzpicture}%
\hspace{1.2cm}%
\begin{tikzpicture}[baseline]

\begin{groupplot}[
    group style={
        group name=barplotsB,
        group size=3 by 1,
        horizontal sep=0.35cm,
        xlabels at=edge bottom
    },
    width=3.2cm,
    height=2.9cm,
    legend style={
        at={(0.05,-0.25)},
        anchor=north,
        legend columns=4,
        column sep=3pt,
        font=\scriptsize
    },
    ybar,
    ymin=0, ymax=72,
    enlarge x limits=19,
    axis lines=none,
    xtick=data,
    symbolic x coords={ModA, ModB, ModC, ModD, ModE, ModF, ModG, ModH},
    xticklabels={},
    title style={yshift=2mm, font=\scriptsize, text depth=0.5ex},
    nodes near coords,
    nodes near coords align={vertical},
]

\nextgroupplot[title={Video}]
\addplot+[fill=c1, draw=c1b, bar width=9.5pt, nodes near coords,
    point meta=explicit symbolic, font=\scriptsize, text=black]
    coordinates {(ModA,35.66) [36]};
\addplot+[fill=c2, draw=c2b, bar width=9.5pt, nodes near coords,
    point meta=explicit symbolic, font=\scriptsize, text=black]
    coordinates {(ModB,40.09) [40]};

\nextgroupplot[title={Subs},
    legend style={at={(0.5,-0.1)}, anchor=north, legend columns=2, font=\scriptsize},
    width=2.4cm]
\addplot+[fill=c1, draw=c1b, bar width=9.5pt, nodes near coords,
    point meta=explicit symbolic, font=\scriptsize, text=black]
    coordinates {(ModA,57.57) [58]};
\addlegendentry{Single-scene}
\addplot+[fill=c2, draw=c2b, bar width=9.5pt, nodes near coords,
    point meta=explicit symbolic, font=\scriptsize, text=black]
    coordinates {(ModB,60.13) [60]};
\addlegendentry{Multi-scene/Global}

\nextgroupplot[title={Video w/ Subs}]
\addplot+[fill=c1, draw=c1b, bar width=9.5pt, nodes near coords,
    point meta=explicit symbolic, font=\scriptsize, text=black]
    coordinates {(ModA,62.93) [63]};
\addplot+[fill=c2, draw=c2b, bar width=9.5pt, nodes near coords,
    point meta=explicit symbolic, font=\scriptsize, text=black]
    coordinates {(ModB,64.6) [65]};

\end{groupplot}

\node[font=\scriptsize, above=0.85cm] at
    ($(barplotsB c1r1.north west)!0.5!(barplotsB c3r1.north east)$)
    {Reasoning Granularity};

\end{tikzpicture}%
}
\caption{Pairwise accuracy for Gemini 2.5 Pro per comprehension dimension and reasoning granularity when varying the input modalities.}
\label{Fig:modalities_cat_gemini}
\end{figure*}

\paragraph{Success and failure patterns.} We analyze representative success and failure cases of Gemini 2.5 Pro. The model performs well on clearly grounded and explicitly depicted information, where visual and textual cues are easy to align (see Examples 3 and 4 in Appendix~\ref{app:examples}). In contrast, failures are more pronounced when the model must rely on implicit cues beyond surface alignment. Specifically, cases involving causal reasoning (e.g., inferring intentions) and implicit event/entity understanding, where relevant event instances must be identified and aggregated over time to form a coherent interpretation (see Examples 1, 2, 5 and 6 in App.~\ref{app:examples}).

\section{Related Work}

\paragraph{Vision and long context LLMs.}
The field of VLMs has seen rapid progress, with models becoming increasingly effective at video-language understanding \citep{molmo2024, Qwen2.5-VL, internvl3}. Early methods focused on short clips and relied on complex spatio-temporal modules, such as Q-formers \citep{damonlpsg2023videollama, li2023videochat}, or temporal pooling techniques \citep{Maaz2023VideoChatGPT, luo2023valley, xu2024pllava}.
While not new, projection layers \citep{li2023llamavidimageworth2, liu2023llava, li2023videochat, liu2024nvilaefficientfrontiervisual} have gained popularity as a simpler and increasingly effective alternative for aligning video and language representations \citep{Qwen2.5-VL, damonlpsg2025videollama3, internvl3}, largely driven by advancements in visual encoders \citep{clip, tschannen2025siglip}. In the domain of long video understanding, current approaches primarily focus on compressing tokens \citep{li2023llamavidimageworth2, damonlpsg2025videollama3}, merely extending the context window \citep{phi3, liu2025world} or memory consolidation mechanisms \citep{balazevic2024memory,song2023moviechat, song2024moviechat+, santos2025inftyvideotrainingfreeapproachlong}. A separate line of work first densely captions videos and then answers questions based on text only \citep{zhang2024simple, wang2024videoagent, wang2024videotree}. We focus instead on benchmarking VLMs in movie narrative understanding without the use of costly captioning pipelines.

\paragraph{Long video understanding benchmarks.}

Understanding long videos presents substantial challenges, requiring models to track complex temporal dependencies and retain narrative context over extended durations. While existing benchmarks have driven progress in temporal reasoning over short clips \citep{xiao2021next, wu2021star} and in domain-specific settings such as instructional or egocentric videos \citep{yang2021justask, mangalam2023egoschema, qiu2024egoplanbench2}, most focus on content under three minutes or can be solved with a few keyframes \citep{yu2019activityqa, zhang2023movqabenchmarkversatilequestionanswering}. More recent benchmarks targeting longer videos~\citep{mangalam2023egoschema, parmar2024causalchaos,wu2024longvideobench, hu2025videommmu,vimul_2025},  often rely on general-purpose video content making them suitable for assessing perception and reasoning capabilities over extended sequences, but less effective for assessing narrative understanding. Other approaches that also focus on long videos primarily target localized “needle-in-a-haystack” retrieval rather than core narrative elements~\citep{tan2024-synopground, timescope2025blog}. 
\paragraph{(Movie) Narrative Understanding.}
Narratives have emerged as an important aspect of long-context evaluation in textual domains, due to their natural temporal and causal structure and rich information-dense content~\citep{xu2024detectiveqa,bonomo-etal-2025-literaryqa, hamilton-etal-2026-narrabench, karpinska-etal-2024-one}. They provide a natural setting for evaluating whether models can integrate information across extended contexts, while maintaining coherence over long event sequences.
Similarly, in the video domain, narrative understanding has been studied using movies and TV shows, which naturally contain structured and temporally extended storylines. \citet{papalampidi-etal-2019-turning-points,papalampidi-etal-2020-screenplay} introduce the task of turning point identification in movies. Other works build benchmarks by sourcing movies~\citep{huang2020movienet,song2023moviechat,song2024moviechat+,rawal2024cinepile}, however they are typically limited in temporal coverage. More recent benchmarks~\citep{ataallah2024infinibenchcomprehensivebenchmarklarge} scale to longer content but rely on automatic question generation and treat all movie content uniformly, often producing questions unrelated to the central storyline (e.g., “needle-in-a-haystack” queries). In contrast, our benchmark is designed with a focus on \textit{memorability}, targeting core narrative events that humans can recall without rewatching the movie. This is intended to better reflect what is salient in human memory and narrative recall. Beyond that, all movies are open-licensed, ensuring a fully reproducible setting and supporting future research efforts, unlike prior datasets that rely on copyright-restricted content or recycle the same sources.


\section{Conclusions}
In this work, we evaluated state-of-the-art models on movie narrative understanding as a long-context multimodal setting. To support this analysis, we introduced \acro, a fully manually curated benchmark built from open-licensed full-length movies, with emphasis on memorable and storyline-central information that existing long-context evaluations often overlook. Our experiments reveal a substantial gap between human and model performance. While subtitles provide a strong signal for narrative understanding and visual cues offer complementary information, current VLMs still struggle to effectively leverage and integrate such information in long multimodal contexts. More broadly, our findings suggest that brute-force encoding of visual content remains insufficient for narrative understanding, as models fail to identify, prioritize, and retain the most relevant information across extended videos, unlike humans. We hope \acro will encourage future work on models that can selectively consolidate and reason over salient narrative information while discarding irrelevant details.

\section*{Limitations}
Despite careful design and validation, our dataset is not free from imperfections. Minor issues such as typos may remain, and annotators—although shown one claim at a time—may have recalled earlier claims from the same pair, potentially influencing subsequent judgments. Models do not share this limitation, as they process claims independently. As future work, claims from each pair could be split into disjoint sets and rated by different annotators to better isolate such effects. Furthermore, although our collection spans a broad variety of genres, most films primarily reflect Western cultural contexts, resulting in limited cultural diversity. While capturing diverse cultural perspectives was not the primary goal of this benchmark and lies outside the scope of this work---given that it is likely orthogonal to aspects of narrative understanding---future work could extend the dataset to include films from a wider range of cultural contexts. Finally, the human and model evaluations differ in their available input modalities, as humans have access to audio while models are evaluated using visual inputs and subtitles. Although some VLMs support audio input, current models are not yet capable of efficiently processing long-form video with all three modalities jointly, limiting the feasibility of a fully matched comparison.

\section*{Ethical considerations}
We adhered to established scientific and ethical standards in constructing and releasing \acro. All source movies are released under the permissive Public Domain 1.0 license. Claims and annotations were created and validated exclusively by the authors; no external crowdworkers were employed. To encourage a plurality of perspectives in the annotation process, the annotation team consists of individuals from diverse demographic, institutional, and geographic backgrounds.
Since \acro is derived entirely from fictional movies, it contains no personally identifiable information (PII) of real individuals. Nonetheless, some fictional content may reflect cultural stereotypes or outdated social norms. We caution researchers that models evaluated on \acro may inherit such biases, and we recommend appropriate safeguards when interpreting or deploying results.
We advise users to employ \acro strictly within the scope of this work, namely as a benchmark for evaluating vision–language models on movie narrative understanding, and discourage its use beyond this intended scope.

\section*{Acknowledgements}
We gratefully acknowledge Miguel Graça, Evan Paces-Wiles, Maya Nachesa, Daniil Larionov, Bryan Sukidi, and José Pombal for their participation in the human evaluation process.
This work was supported by the project DECOLLAGE (ERC-2022-CoG 101088763), by a research grant (VIL53122) from VILLUM FONDEN, by the Pioneer Center for AI DNRF grant number P1, by the NNF project 0094281, by FCT/MECI through national funds and when applicable co-funded EU funds under UID/50008: Instituto de Telecomunicações, by DARPA ECOLE Program No. HR00112390060, NSF-AI Engage Institute DRL-2112635, ARO Award W911NF2110220, ONR Grant N00014-23-1-2356, the Microsoft Accelerate Foundation Models Research (AFMR) grant program, by the project PID2023-151351NB-I00 funded by MCIN/ AEI /10.13039/501100011033 and by ERDF, UE. This collaboration resulted from  an ELLIS workshop at MFO, the Oberwolfach Research Institute for Mathematics in the German Black Forest. The event was funded by the state of Baden-W\"urttemberg (Germany) and organised in collaboration with the ELLIS Institute T\"ubingen and the Max Planck Institute for Intelligent Systems.

\bibliography{acl_latex}

@inproceedings{hudson-al-moubayed-2022-muld,
    title = "{M}u{LD}: The Multitask Long Document Benchmark",
    author = "Hudson, George  and
      Al Moubayed, Noura",
    editor = "Calzolari, Nicoletta  and
      B{\'e}chet, Fr{\'e}d{\'e}ric  and
      Blache, Philippe  and
      Choukri, Khalid  and
      Cieri, Christopher  and
      Declerck, Thierry  and
      Goggi, Sara  and
      Isahara, Hitoshi  and
      Maegaard, Bente  and
      Mariani, Joseph  and
      Mazo, H{\'e}l{\`e}ne  and
      Odijk, Jan  and
      Piperidis, Stelios",
    booktitle = "Proceedings of the Thirteenth Language Resources and Evaluation Conference",
    month = jun,
    year = "2022",
    address = "Marseille, France",
    publisher = "European Language Resources Association",
    url = "https://aclanthology.org/2022.lrec-1.392/",
    pages = "3675--3685"
}

@inproceedings{an-etal-2024-l,
    title = "{L}-Eval: Instituting Standardized Evaluation for Long Context Language Models",
    author = "An, Chenxin  and
      Gong, Shansan  and
      Zhong, Ming  and
      Zhao, Xingjian  and
      Li, Mukai  and
      Zhang, Jun  and
      Kong, Lingpeng  and
      Qiu, Xipeng",
    editor = "Ku, Lun-Wei  and
      Martins, Andre  and
      Srikumar, Vivek",
    booktitle = "Proceedings of the 62nd Annual Meeting of the Association for Computational Linguistics (Volume 1: Long Papers)",
    month = aug,
    year = "2024",
    address = "Bangkok, Thailand",
    publisher = "Association for Computational Linguistics",
    url = "https://aclanthology.org/2024.acl-long.776/",
    doi = "10.18653/v1/2024.acl-long.776",
    pages = "14388--14411"
}

@inproceedings{piper-bagga-2024-using,
    title = "Using Large Language Models for Understanding Narrative Discourse",
    author = "Piper, Andrew  and
      Bagga, Sunyam",
    editor = "Lal, Yash Kumar  and
      Clark, Elizabeth  and
      Iyyer, Mohit  and
      Chaturvedi, Snigdha  and
      Brei, Anneliese  and
      Brahman, Faeze  and
      Chandu, Khyathi Raghavi",
    booktitle = "Proceedings of the 6th Workshop on Narrative Understanding",
    month = nov,
    year = "2024",
    address = "Miami, Florida, USA",
    publisher = "Association for Computational Linguistics",
    url = "https://aclanthology.org/2024.wnu-1.4/",
    doi = "10.18653/v1/2024.wnu-1.4",
    pages = "37--46"
}

@inproceedings{karpinska-etal-2024-one,
    title = "One Thousand and One Pairs: A {\textquotedblleft}novel{\textquotedblright} challenge for long-context language models",
    author = "Karpinska, Marzena  and
      Thai, Katherine  and
      Lo, Kyle  and
      Goyal, Tanya  and
      Iyyer, Mohit",
    editor = "Al-Onaizan, Yaser  and
      Bansal, Mohit  and
      Chen, Yun-Nung",
    booktitle = "Proceedings of the 2024 Conference on Empirical Methods in Natural Language Processing",
    month = nov,
    year = "2024",
    address = "Miami, Florida, USA",
    publisher = "Association for Computational Linguistics",
    url = "https://aclanthology.org/2024.emnlp-main.948/",
    doi = "10.18653/v1/2024.emnlp-main.948",
    pages = "17048--17085"
}

@inproceedings{hamilton-etal-2026-narrabench,
    title = "{N}arra{B}ench: A Comprehensive Framework for Narrative Benchmarking",
    author = "Hamilton, Sil  and
      Wilkens, Matthew  and
      Piper, Andrew",
    editor = "Demberg, Vera  and
      Inui, Kentaro  and
      Marquez, Llu{\'i}s",
    booktitle = "Proceedings of the 19th Conference of the {E}uropean Chapter of the {A}ssociation for {C}omputational {L}inguistics (Volume 1: Long Papers)",
    month = mar,
    year = "2026",
    address = "Rabat, Morocco",
    publisher = "Association for Computational Linguistics",
    url = "https://aclanthology.org/2026.eacl-long.176/",
    doi = "10.18653/v1/2026.eacl-long.176",
    pages = "3786--3801",
    ISBN = "979-8-89176-380-7"
}

@inproceedings{chandrasegaran2024hourvideo,
    title={HourVideo: 1-Hour Video-Language Understanding},
    author={Chandrasegaran, Keshigeyan and Gupta, Agrim and Hadzic, Lea M. and Kota, Taran and 
    He, Jimming and Eyzaguirre, Cristobal and Durante, Zane and Li, Manling and Wu, Jiajun and Li, Fei-Fei},
    booktitle = {Advances in Neural Information Processing Systems},
    year={2024},
    volume = {37},
}

@inproceedings{ataallah2024infinibenchcomprehensivebenchmarklarge,
    title = "{I}nfini{B}ench: A Benchmark for Large Multi-Modal Models in Long-Form Movies and {TV} Shows",
    author = "Ataallah, Kirolos  and
      Bakr, Eslam Mohamed  and
      Ahmed, Mahmoud  and
      Gou, Chenhui  and
      Pahwa, Khushbu  and
      Ding, Jian  and
      Elhoseiny, Mohamed",
    editor = "Christodoulopoulos, Christos  and
      Chakraborty, Tanmoy  and
      Rose, Carolyn  and
      Peng, Violet",
    booktitle = "Proceedings of the 2025 Conference on Empirical Methods in Natural Language Processing",
    month = nov,
    year = "2025",
    address = "Suzhou, China",
    publisher = "Association for Computational Linguistics",
    url = "https://aclanthology.org/2025.emnlp-main.984/",
    doi = "10.18653/v1/2025.emnlp-main.984",
    pages = "19485--19512",
    ISBN = "979-8-89176-332-6"
}

@inproceedings{wang2025lvbench,
  title={Lvbench: An extreme long video understanding benchmark},
  author={Wang, Weihan and He, Zehai and Hong, Wenyi and Cheng, Yean and Zhang, Xiaohan and Qi, Ji and Ding, Ming and Gu, Xiaotao and Huang, Shiyu and Xu, Bin and others},
  booktitle={Proceedings of the IEEE/CVF International Conference on Computer Vision},
  pages={22958--22967},
  year={2025}
}

@inproceedings{fu2025videomme,
  title={Video-mme: The first-ever comprehensive evaluation benchmark of multi-modal llms in video analysis},
  author={Fu, Chaoyou and Dai, Yuhan and Luo, Yongdong and Li, Lei and Ren, Shuhuai and Zhang, Renrui and Wang, Zihan and Zhou, Chenyu and Shen, Yunhang and Zhang, Mengdan and others},
  booktitle={CVPR},
  year={2025}
}

@inproceedings{wu2024longvideobench,
 author = {Wu, Haoning and Li, Dongxu and Chen, Bei and Li, Junnan},
 booktitle = {Advances in Neural Information Processing Systems},
 doi = {10.52202/079017-0907},
 editor = {A. Globerson and L. Mackey and D. Belgrave and A. Fan and U. Paquet and J. Tomczak and C. Zhang},
 pages = {28828--28857},
 publisher = {Curran Associates, Inc.},
 title = {LongVideoBench: A Benchmark for Long-context Interleaved Video-Language Understanding},
 url = {https://proceedings.neurips.cc/paper_files/paper/2024/file/329ad516cf7a6ac306f29882e9c77558-Paper-Datasets_and_Benchmarks_Track.pdf},
 volume = {37},
 year = {2024}
}

@inproceedings{huang2020movienet,
  title={Movienet: A holistic dataset for movie understanding},
  author={Huang, Qingqiu and Xiong, Yu and Rao, Anyi and Wang, Jiaze and Lin, Dahua},
  booktitle={Computer Vision--ECCV 2020: 16th European Conference, Glasgow, UK, August 23--28, 2020, Proceedings, Part IV 16},
  pages={709--727},
  year={2020},
  organization={Springer}
}

@article{song2023moviechat,
  title={MovieChat: From Dense Token to Sparse Memory for Long Video Understanding},
  author={Song, Enxin and Chai, Wenhao and Wang, Guanhong and Zhang, Yucheng and Zhou, Haoyang and Wu, Feiyang and Guo, Xun and Ye, Tian and Lu, Yan and Hwang, Jenq-Neng and others},
  journal={arXiv preprint arXiv:2307.16449},
  year={2023}
}

@article{song2024moviechat+,
  title={MovieChat+: Question-aware Sparse Memory for Long Video Question Answering},
  author={Song, Enxin and Chai, Wenhao and Ye, Tian and Hwang, Jenq-Neng and Li, Xi and Wang, Gaoang},
  journal={arXiv preprint arXiv:2404.17176},
  year={2024}
}

@inproceedings{papalampidi-etal-2019-turning-points,
    title = "Movie Plot Analysis via Turning Point Identification",
    author = "Papalampidi, Pinelopi  and
      Keller, Frank  and
      Lapata, Mirella",
    editor = "Inui, Kentaro  and
      Jiang, Jing  and
      Ng, Vincent  and
      Wan, Xiaojun",
    booktitle = "Proceedings of the 2019 Conference on Empirical Methods in Natural Language Processing and the 9th International Joint Conference on Natural Language Processing (EMNLP-IJCNLP)",
    month = nov,
    year = "2019",
    address = "Hong Kong, China",
    publisher = "Association for Computational Linguistics",
    url = "https://aclanthology.org/D19-1180/",
    doi = "10.18653/v1/D19-1180",
    pages = "1707--1717"
}

@inproceedings{papalampidi-etal-2020-screenplay,
    title = "Screenplay Summarization Using Latent Narrative Structure",
    author = "Papalampidi, Pinelopi  and
      Keller, Frank  and
      Frermann, Lea  and
      Lapata, Mirella",
    editor = "Jurafsky, Dan  and
      Chai, Joyce  and
      Schluter, Natalie  and
      Tetreault, Joel",
    booktitle = "Proceedings of the 58th Annual Meeting of the Association for Computational Linguistics",
    month = jul,
    year = "2020",
    address = "Online",
    publisher = "Association for Computational Linguistics",
    url = "https://aclanthology.org/2020.acl-main.174/",
    doi = "10.18653/v1/2020.acl-main.174",
    pages = "1920--1933"
}

@inproceedings{xiao2021next,
  title={Next-qa: Next phase of question-answering to explaining temporal actions},
  author={Xiao, Junbin and Shang, Xindi and Yao, Angela and Chua, Tat-Seng},
  booktitle={Proceedings of the IEEE/CVF conference on computer vision and pattern recognition},
  pages={9777--9786},
  year={2021}
}

@misc{zhang2023movqabenchmarkversatilequestionanswering,
      title={MoVQA: A Benchmark of Versatile Question-Answering for Long-Form Movie Understanding}, 
      author={Hongjie Zhang and Yi Liu and Lu Dong and Yifei Huang and Zhen-Hua Ling and Yali Wang and Limin Wang and Yu Qiao},
      year={2023},
      eprint={2312.04817},
      archivePrefix={arXiv},
      primaryClass={cs.CV},
      url={https://arxiv.org/abs/2312.04817}, 
}

@inproceedings{radford2023robust_whisper,
  title={Robust speech recognition via large-scale weak supervision},
  author={Radford, Alec and Kim, Jong Wook and Xu, Tao and Brockman, Greg and McLeavey, Christine and Sutskever, Ilya},
  booktitle={International conference on machine learning},
  pages={28492--28518},
  year={2023},
  organization={PMLR}
}

@inproceedings{
fu2025hidden,
title={Hidden in plain sight: {VLM}s overlook their visual representations},
author={Stephanie Fu and tyler bonnen and Devin Guillory and Trevor Darrell},
booktitle={Second Conference on Language Modeling},
year={2025},
url={https://openreview.net/forum?id=qQb1JLrwol}
}

@misc{zheng2025mllmsdeeplyaffectedmodality,
      title={MLLMs are Deeply Affected by Modality Bias}, 
      author={Xu Zheng and Chenfei Liao and Yuqian Fu and Kaiyu Lei and Yuanhuiyi Lyu and Lutao Jiang and Bin Ren and Jialei Chen and Jiawen Wang and Chengxin Li and Linfeng Zhang and Danda Pani Paudel and Xuanjing Huang and Yu-Gang Jiang and Nicu Sebe and Dacheng Tao and Luc Van Gool and Xuming Hu},
      year={2025},
      eprint={2505.18657},
      archivePrefix={arXiv},
      primaryClass={cs.AI},
      url={https://arxiv.org/abs/2505.18657}, 
}

@misc{openai2024gpt4ocard,
      title={GPT-4o System Card}, 
      author={OpenAI and : and Aaron Hurst and Adam Lerer and Adam P. Goucher and Adam Perelman and Aditya Ramesh and Aidan Clark and AJ Ostrow and Akila Welihinda and Alan Hayes and Alec Radford and Aleksander Mądry and Alex Baker-Whitcomb and Alex Beutel and Alex Borzunov and Alex Carney and Alex Chow and Alex Kirillov and Alex Nichol and Alex Paino and Alex Renzin and Alex Tachard Passos and Alexander Kirillov and Alexi Christakis and Alexis Conneau and Ali Kamali and Allan Jabri and Allison Moyer and Allison Tam and Amadou Crookes and Amin Tootoochian and Amin Tootoonchian and Ananya Kumar and Andrea Vallone and Andrej Karpathy and Andrew Braunstein and Andrew Cann and Andrew Codispoti and Andrew Galu and Andrew Kondrich and Andrew Tulloch and Andrey Mishchenko and Angela Baek and Angela Jiang and Antoine Pelisse and Antonia Woodford and Anuj Gosalia and Arka Dhar and Ashley Pantuliano and Avi Nayak and Avital Oliver and Barret Zoph and Behrooz Ghorbani and Ben Leimberger and Ben Rossen and Ben Sokolowsky and Ben Wang and Benjamin Zweig and Beth Hoover and Blake Samic and Bob McGrew and Bobby Spero and Bogo Giertler and Bowen Cheng and Brad Lightcap and Brandon Walkin and Brendan Quinn and Brian Guarraci and Brian Hsu and Bright Kellogg and Brydon Eastman and Camillo Lugaresi and Carroll Wainwright and Cary Bassin and Cary Hudson and Casey Chu and Chad Nelson and Chak Li and Chan Jun Shern and Channing Conger and Charlotte Barette and Chelsea Voss and Chen Ding and Cheng Lu and Chong Zhang and Chris Beaumont and Chris Hallacy and Chris Koch and Christian Gibson and Christina Kim and Christine Choi and Christine McLeavey and Christopher Hesse and Claudia Fischer and Clemens Winter and Coley Czarnecki and Colin Jarvis and Colin Wei and Constantin Koumouzelis and Dane Sherburn and Daniel Kappler and Daniel Levin and Daniel Levy and David Carr and David Farhi and David Mely and David Robinson and David Sasaki and Denny Jin and Dev Valladares and Dimitris Tsipras and Doug Li and Duc Phong Nguyen and Duncan Findlay and Edede Oiwoh and Edmund Wong and Ehsan Asdar and Elizabeth Proehl and Elizabeth Yang and Eric Antonow and Eric Kramer and Eric Peterson and Eric Sigler and Eric Wallace and Eugene Brevdo and Evan Mays and Farzad Khorasani and Felipe Petroski Such and Filippo Raso and Francis Zhang and Fred von Lohmann and Freddie Sulit and Gabriel Goh and Gene Oden and Geoff Salmon and Giulio Starace and Greg Brockman and Hadi Salman and Haiming Bao and Haitang Hu and Hannah Wong and Haoyu Wang and Heather Schmidt and Heather Whitney and Heewoo Jun and Hendrik Kirchner and Henrique Ponde de Oliveira Pinto and Hongyu Ren and Huiwen Chang and Hyung Won Chung and Ian Kivlichan and Ian O'Connell and Ian O'Connell and Ian Osband and Ian Silber and Ian Sohl and Ibrahim Okuyucu and Ikai Lan and Ilya Kostrikov and Ilya Sutskever and Ingmar Kanitscheider and Ishaan Gulrajani and Jacob Coxon and Jacob Menick and Jakub Pachocki and James Aung and James Betker and James Crooks and James Lennon and Jamie Kiros and Jan Leike and Jane Park and Jason Kwon and Jason Phang and Jason Teplitz and Jason Wei and Jason Wolfe and Jay Chen and Jeff Harris and Jenia Varavva and Jessica Gan Lee and Jessica Shieh and Ji Lin and Jiahui Yu and Jiayi Weng and Jie Tang and Jieqi Yu and Joanne Jang and Joaquin Quinonero Candela and Joe Beutler and Joe Landers and Joel Parish and Johannes Heidecke and John Schulman and Jonathan Lachman and Jonathan McKay and Jonathan Uesato and Jonathan Ward and Jong Wook Kim and Joost Huizinga and Jordan Sitkin and Jos Kraaijeveld and Josh Gross and Josh Kaplan and Josh Snyder and Joshua Achiam and Joy Jiao and Joyce Lee and Juntang Zhuang and Justyn Harriman and Kai Fricke and Kai Hayashi and Karan Singhal and Katy Shi and Kavin Karthik and Kayla Wood and Kendra Rimbach and Kenny Hsu and Kenny Nguyen and Keren Gu-Lemberg and Kevin Button and Kevin Liu and Kiel Howe and Krithika Muthukumar and Kyle Luther and Lama Ahmad and Larry Kai and Lauren Itow and Lauren Workman and Leher Pathak and Leo Chen and Li Jing and Lia Guy and Liam Fedus and Liang Zhou and Lien Mamitsuka and Lilian Weng and Lindsay McCallum and Lindsey Held and Long Ouyang and Louis Feuvrier and Lu Zhang and Lukas Kondraciuk and Lukasz Kaiser and Luke Hewitt and Luke Metz and Lyric Doshi and Mada Aflak and Maddie Simens and Madelaine Boyd and Madeleine Thompson and Marat Dukhan and Mark Chen and Mark Gray and Mark Hudnall and Marvin Zhang and Marwan Aljubeh and Mateusz Litwin and Matthew Zeng and Max Johnson and Maya Shetty and Mayank Gupta and Meghan Shah and Mehmet Yatbaz and Meng Jia Yang and Mengchao Zhong and Mia Glaese and Mianna Chen and Michael Janner and Michael Lampe and Michael Petrov and Michael Wu and Michele Wang and Michelle Fradin and Michelle Pokrass and Miguel Castro and Miguel Oom Temudo de Castro and Mikhail Pavlov and Miles Brundage and Miles Wang and Minal Khan and Mira Murati and Mo Bavarian and Molly Lin and Murat Yesildal and Nacho Soto and Natalia Gimelshein and Natalie Cone and Natalie Staudacher and Natalie Summers and Natan LaFontaine and Neil Chowdhury and Nick Ryder and Nick Stathas and Nick Turley and Nik Tezak and Niko Felix and Nithanth Kudige and Nitish Keskar and Noah Deutsch and Noel Bundick and Nora Puckett and Ofir Nachum and Ola Okelola and Oleg Boiko and Oleg Murk and Oliver Jaffe and Olivia Watkins and Olivier Godement and Owen Campbell-Moore and Patrick Chao and Paul McMillan and Pavel Belov and Peng Su and Peter Bak and Peter Bakkum and Peter Deng and Peter Dolan and Peter Hoeschele and Peter Welinder and Phil Tillet and Philip Pronin and Philippe Tillet and Prafulla Dhariwal and Qiming Yuan and Rachel Dias and Rachel Lim and Rahul Arora and Rajan Troll and Randall Lin and Rapha Gontijo Lopes and Raul Puri and Reah Miyara and Reimar Leike and Renaud Gaubert and Reza Zamani and Ricky Wang and Rob Donnelly and Rob Honsby and Rocky Smith and Rohan Sahai and Rohit Ramchandani and Romain Huet and Rory Carmichael and Rowan Zellers and Roy Chen and Ruby Chen and Ruslan Nigmatullin and Ryan Cheu and Saachi Jain and Sam Altman and Sam Schoenholz and Sam Toizer and Samuel Miserendino and Sandhini Agarwal and Sara Culver and Scott Ethersmith and Scott Gray and Sean Grove and Sean Metzger and Shamez Hermani and Shantanu Jain and Shengjia Zhao and Sherwin Wu and Shino Jomoto and Shirong Wu and Shuaiqi and Xia and Sonia Phene and Spencer Papay and Srinivas Narayanan and Steve Coffey and Steve Lee and Stewart Hall and Suchir Balaji and Tal Broda and Tal Stramer and Tao Xu and Tarun Gogineni and Taya Christianson and Ted Sanders and Tejal Patwardhan and Thomas Cunninghman and Thomas Degry and Thomas Dimson and Thomas Raoux and Thomas Shadwell and Tianhao Zheng and Todd Underwood and Todor Markov and Toki Sherbakov and Tom Rubin and Tom Stasi and Tomer Kaftan and Tristan Heywood and Troy Peterson and Tyce Walters and Tyna Eloundou and Valerie Qi and Veit Moeller and Vinnie Monaco and Vishal Kuo and Vlad Fomenko and Wayne Chang and Weiyi Zheng and Wenda Zhou and Wesam Manassra and Will Sheu and Wojciech Zaremba and Yash Patil and Yilei Qian and Yongjik Kim and Youlong Cheng and Yu Zhang and Yuchen He and Yuchen Zhang and Yujia Jin and Yunxing Dai and Yury Malkov},
      year={2024},
      eprint={2410.21276},
      archivePrefix={arXiv},
      primaryClass={cs.CL},
      url={https://arxiv.org/abs/2410.21276}, 
}

@article{team2023gemini,
  title={Gemini: a family of highly capable multimodal models},
  author={Team, Gemini and Anil, Rohan and Borgeaud, Sebastian and Alayrac, Jean-Baptiste and Yu, Jiahui and Soricut, Radu and Schalkwyk, Johan and Dai, Andrew M and Hauth, Anja and Millican, Katie and others},
  journal={arXiv preprint arXiv:2312.11805},
  year={2023}
}

@article{Qwen2.5-VL,
  title={Qwen2.5-VL Technical Report},
  author={Bai, Shuai and Chen, Keqin and Liu, Xuejing and Wang, Jialin and Ge, Wenbin and Song, Sibo and Dang, Kai and Wang, Peng and Wang, Shijie and Tang, Jun and Zhong, Humen and Zhu, Yuanzhi and Yang, Mingkun and Li, Zhaohai and Wan, Jianqiang and Wang, Pengfei and Ding, Wei and Fu, Zheren and Xu, Yiheng and Ye, Jiabo and Zhang, Xi and Xie, Tianbao and Cheng, Zesen and Zhang, Hang and Yang, Zhibo and Xu, Haiyang and Lin, Junyang},
  journal={arXiv preprint arXiv:2502.13923},
  year={2025}
}

@article{molmo2024,
  title={Molmo and PixMo: Open Weights and Open Data for State-of-the-Art Multimodal Models},
  author={Matt Deitke and Christopher Clark and Sangho Lee and Rohun Tripathi and Yue Yang and Jae Sung Park and Mohammadreza Salehi and Niklas Muennighoff and Kyle Lo and Luca Soldaini and Jiasen Lu and Taira Anderson and Erin Bransom and Kiana Ehsani and Huong Ngo and YenSung Chen and Ajay Patel and Mark Yatskar and Chris Callison-Burch and Andrew Head and Rose Hendrix and Favyen Bastani and Eli VanderBilt and Nathan Lambert and Yvonne Chou and Arnavi Chheda and Jenna Sparks and Sam Skjonsberg and Michael Schmitz and Aaron Sarnat and Byron Bischoff and Pete Walsh and Chris Newell and Piper Wolters and Tanmay Gupta and Kuo-Hao Zeng and Jon Borchardt and Dirk Groeneveld and Jen Dumas and Crystal Nam and Sophie Lebrecht and Caitlin Wittlif and Carissa Schoenick and Oscar Michel and Ranjay Krishna and Luca Weihs and Noah A. Smith and Hannaneh Hajishirzi and Ross Girshick and Ali Farhadi and Aniruddha Kembhavi},
  journal={arXiv preprint arXiv:2409.17146},
  year={2024}
}

@article{Qwen3-VL,
      title={Qwen3-VL Technical Report}, 
      author={Shuai Bai and Yuxuan Cai and Ruizhe Chen and Keqin Chen and Xionghui Chen and Zesen Cheng and Lianghao Deng and Wei Ding and Chang Gao and Chunjiang Ge and Wenbin Ge and Zhifang Guo and Qidong Huang and Jie Huang and Fei Huang and Binyuan Hui and Shutong Jiang and Zhaohai Li and Mingsheng Li and Mei Li and Kaixin Li and Zicheng Lin and Junyang Lin and Xuejing Liu and Jiawei Liu and Chenglong Liu and Yang Liu and Dayiheng Liu and Shixuan Liu and Dunjie Lu and Ruilin Luo and Chenxu Lv and Rui Men and Lingchen Meng and Xuancheng Ren and Xingzhang Ren and Sibo Song and Yuchong Sun and Jun Tang and Jianhong Tu and Jianqiang Wan and Peng Wang and Pengfei Wang and Qiuyue Wang and Yuxuan Wang and Tianbao Xie and Yiheng Xu and Haiyang Xu and Jin Xu and Zhibo Yang and Mingkun Yang and Jianxin Yang and An Yang and Bowen Yu and Fei Zhang and Hang Zhang and Xi Zhang and Bo Zheng and Humen Zhong and Jingren Zhou and Fan Zhou and Jing Zhou and Yuanzhi Zhu and Ke Zhu},
	  journal={arXiv preprint arXiv:2511.21631},
      year={2025}
}

@ARTICLE{internvl3,
       author = {{Zhu}, Jinguo and {Wang}, Weiyun and {Chen}, Zhe and {Liu}, Zhaoyang and {Ye}, Shenglong and {Gu}, Lixin and {Tian}, Hao and {Duan}, Yuchen and {Su}, Weijie and {Shao}, Jie and {Gao}, Zhangwei and {Cui}, Erfei and {Wang}, Xuehui and {Cao}, Yue and {Liu}, Yangzhou and {Wei}, Xingguang and {Zhang}, Hongjie and {Wang}, Haomin and {Xu}, Weiye and {Li}, Hao and {Wang}, Jiahao and {Deng}, Nianchen and {Li}, Songze and {He}, Yinan and {Jiang}, Tan and {Luo}, Jiapeng and {Wang}, Yi and {He}, Conghui and {Shi}, Botian and {Zhang}, Xingcheng and {Shao}, Wenqi and {He}, Junjun and {Xiong}, Yingtong and {Qu}, Wenwen and {Sun}, Peng and {Jiao}, Penglong and {Lv}, Han and {Wu}, Lijun and {Zhang}, Kaipeng and {Deng}, Huipeng and {Ge}, Jiaye and {Chen}, Kai and {Wang}, Limin and {Dou}, Min and {Lu}, Lewei and {Zhu}, Xizhou and {Lu}, Tong and {Lin}, Dahua and {Qiao}, Yu and {Dai}, Jifeng and {Wang}, Wenhai},
        title = "{InternVL3: Exploring Advanced Training and Test-Time Recipes for Open-Source Multimodal Models}",
      journal = {arXiv e-prints},
         year = 2025,
        month = apr,
          eid = {arXiv:2504.10479},
        pages = {arXiv:2504.10479},
          doi = {10.48550/arXiv.2504.10479},
archivePrefix = {arXiv},
       eprint = {2504.10479},
 primaryClass = {cs.CV},
       adsurl = {https://ui.adsabs.harvard.edu/abs/2025arXiv250410479Z}
}

@misc{wang2025internvl35advancingopensourcemultimodal,
      title={InternVL3.5: Advancing Open-Source Multimodal Models in Versatility, Reasoning, and Efficiency}, 
      author={Weiyun Wang and Zhangwei Gao and Lixin Gu and Hengjun Pu and Long Cui and Xingguang Wei and Zhaoyang Liu and Linglin Jing and Shenglong Ye and Jie Shao and Zhaokai Wang and Zhe Chen and Hongjie Zhang and Ganlin Yang and Haomin Wang and Qi Wei and Jinhui Yin and Wenhao Li and Erfei Cui and Guanzhou Chen and Zichen Ding and Changyao Tian and Zhenyu Wu and Jingjing Xie and Zehao Li and Bowen Yang and Yuchen Duan and Xuehui Wang and Zhi Hou and Haoran Hao and Tianyi Zhang and Songze Li and Xiangyu Zhao and Haodong Duan and Nianchen Deng and Bin Fu and Yinan He and Yi Wang and Conghui He and Botian Shi and Junjun He and Yingtong Xiong and Han Lv and Lijun Wu and Wenqi Shao and Kaipeng Zhang and Huipeng Deng and Biqing Qi and Jiaye Ge and Qipeng Guo and Wenwei Zhang and Songyang Zhang and Maosong Cao and Junyao Lin and Kexian Tang and Jianfei Gao and Haian Huang and Yuzhe Gu and Chengqi Lyu and Huanze Tang and Rui Wang and Haijun Lv and Wanli Ouyang and Limin Wang and Min Dou and Xizhou Zhu and Tong Lu and Dahua Lin and Jifeng Dai and Weijie Su and Bowen Zhou and Kai Chen and Yu Qiao and Wenhai Wang and Gen Luo},
      year={2025},
      eprint={2508.18265},
      archivePrefix={arXiv},
      primaryClass={cs.CV},
      url={https://arxiv.org/abs/2508.18265}, 
}

@article{kangaroogroup,
	title={Kangaroo: A Powerful Video-Language Model Supporting Long-context Video Input},
	author={Liu, Jiajun and Wang, Yibing and Ma, Hanghang and Wu, Xiaoping and Ma, Xiaoqi and Wei, xiaoming and Jiao, Jianbin and Wu, Enhua and Hu, Jie},
	journal={arXiv preprint arXiv:2408.15542},
	year={2024}
}

@ARTICLE{Ovis,
       author = {{Lu}, Shiyin and {Li}, Yang and {Chen}, Qing-Guo and {Xu}, Zhao and {Luo}, Weihua and {Zhang}, Kaifu and {Ye}, Han-Jia},
        title = "{Ovis: Structural Embedding Alignment for Multimodal Large Language Model}",
      journal = {arXiv e-prints},
         year = 2024,
        month = may,
          eid = {arXiv:2405.20797},
        pages = {arXiv:2405.20797},
          doi = {10.48550/arXiv.2405.20797},
archivePrefix = {arXiv},
       eprint = {2405.20797},
 primaryClass = {cs.CV},
       adsurl = {https://ui.adsabs.harvard.edu/abs/2024arXiv240520797L}
}

@misc{gemmateam2025gemma3technicalreport,
      title={Gemma 3 Technical Report}, 
      author={Gemma Team and Aishwarya Kamath and Johan Ferret and Shreya Pathak and Nino Vieillard and Ramona Merhej and Sarah Perrin and Tatiana Matejovicova and Alexandre Ramé and Morgane Rivière and Louis Rouillard and Thomas Mesnard and Geoffrey Cideron and Jean-bastien Grill and Sabela Ramos and Edouard Yvinec and Michelle Casbon and Etienne Pot and Ivo Penchev and Gaël Liu and Francesco Visin and Kathleen Kenealy and Lucas Beyer and Xiaohai Zhai and Anton Tsitsulin and Robert Busa-Fekete and Alex Feng and Noveen Sachdeva and Benjamin Coleman and Yi Gao and Basil Mustafa and Iain Barr and Emilio Parisotto and David Tian and Matan Eyal and Colin Cherry and Jan-Thorsten Peter and Danila Sinopalnikov and Surya Bhupatiraju and Rishabh Agarwal and Mehran Kazemi and Dan Malkin and Ravin Kumar and David Vilar and Idan Brusilovsky and Jiaming Luo and Andreas Steiner and Abe Friesen and Abhanshu Sharma and Abheesht Sharma and Adi Mayrav Gilady and Adrian Goedeckemeyer and Alaa Saade and Alex Feng and Alexander Kolesnikov and Alexei Bendebury and Alvin Abdagic and Amit Vadi and András György and André Susano Pinto and Anil Das and Ankur Bapna and Antoine Miech and Antoine Yang and Antonia Paterson and Ashish Shenoy and Ayan Chakrabarti and Bilal Piot and Bo Wu and Bobak Shahriari and Bryce Petrini and Charlie Chen and Charline Le Lan and Christopher A. Choquette-Choo and CJ Carey and Cormac Brick and Daniel Deutsch and Danielle Eisenbud and Dee Cattle and Derek Cheng and Dimitris Paparas and Divyashree Shivakumar Sreepathihalli and Doug Reid and Dustin Tran and Dustin Zelle and Eric Noland and Erwin Huizenga and Eugene Kharitonov and Frederick Liu and Gagik Amirkhanyan and Glenn Cameron and Hadi Hashemi and Hanna Klimczak-Plucińska and Harman Singh and Harsh Mehta and Harshal Tushar Lehri and Hussein Hazimeh and Ian Ballantyne and Idan Szpektor and Ivan Nardini and Jean Pouget-Abadie and Jetha Chan and Joe Stanton and John Wieting and Jonathan Lai and Jordi Orbay and Joseph Fernandez and Josh Newlan and Ju-yeong Ji and Jyotinder Singh and Kat Black and Kathy Yu and Kevin Hui and Kiran Vodrahalli and Klaus Greff and Linhai Qiu and Marcella Valentine and Marina Coelho and Marvin Ritter and Matt Hoffman and Matthew Watson and Mayank Chaturvedi and Michael Moynihan and Min Ma and Nabila Babar and Natasha Noy and Nathan Byrd and Nick Roy and Nikola Momchev and Nilay Chauhan and Noveen Sachdeva and Oskar Bunyan and Pankil Botarda and Paul Caron and Paul Kishan Rubenstein and Phil Culliton and Philipp Schmid and Pier Giuseppe Sessa and Pingmei Xu and Piotr Stanczyk and Pouya Tafti and Rakesh Shivanna and Renjie Wu and Renke Pan and Reza Rokni and Rob Willoughby and Rohith Vallu and Ryan Mullins and Sammy Jerome and Sara Smoot and Sertan Girgin and Shariq Iqbal and Shashir Reddy and Shruti Sheth and Siim Põder and Sijal Bhatnagar and Sindhu Raghuram Panyam and Sivan Eiger and Susan Zhang and Tianqi Liu and Trevor Yacovone and Tyler Liechty and Uday Kalra and Utku Evci and Vedant Misra and Vincent Roseberry and Vlad Feinberg and Vlad Kolesnikov and Woohyun Han and Woosuk Kwon and Xi Chen and Yinlam Chow and Yuvein Zhu and Zichuan Wei and Zoltan Egyed and Victor Cotruta and Minh Giang and Phoebe Kirk and Anand Rao and Kat Black and Nabila Babar and Jessica Lo and Erica Moreira and Luiz Gustavo Martins and Omar Sanseviero and Lucas Gonzalez and Zach Gleicher and Tris Warkentin and Vahab Mirrokni and Evan Senter and Eli Collins and Joelle Barral and Zoubin Ghahramani and Raia Hadsell and Yossi Matias and D. Sculley and Slav Petrov and Noah Fiedel and Noam Shazeer and Oriol Vinyals and Jeff Dean and Demis Hassabis and Koray Kavukcuoglu and Clement Farabet and Elena Buchatskaya and Jean-Baptiste Alayrac and Rohan Anil and Dmitry and Lepikhin and Sebastian Borgeaud and Olivier Bachem and Armand Joulin and Alek Andreev and Cassidy Hardin and Robert Dadashi and Léonard Hussenot},
      year={2025},
      eprint={2503.19786},
      archivePrefix={arXiv},
      primaryClass={cs.CL},
      url={https://arxiv.org/abs/2503.19786}, 
}

@misc{vteam2026glm45vglm41vthinkingversatilemultimodal,
      title={GLM-4.5V and GLM-4.1V-Thinking: Towards Versatile Multimodal Reasoning with Scalable Reinforcement Learning}, 
      author={V Team and Wenyi Hong and Wenmeng Yu and Xiaotao Gu and Guo Wang and Guobing Gan and Haomiao Tang and Jiale Cheng and Ji Qi and Junhui Ji and Lihang Pan and Shuaiqi Duan and Weihan Wang and Yan Wang and Yean Cheng and Zehai He and Zhe Su and Zhen Yang and Ziyang Pan and Aohan Zeng and Baoxu Wang and Bin Chen and Boyan Shi and Changyu Pang and Chenhui Zhang and Da Yin and Fan Yang and Guoqing Chen and Haochen Li and Jiale Zhu and Jiali Chen and Jiaxing Xu and Jiazheng Xu and Jing Chen and Jinghao Lin and Jinhao Chen and Jinjiang Wang and Junjie Chen and Leqi Lei and Letian Gong and Leyi Pan and Mingdao Liu and Mingde Xu and Mingzhi Zhang and Qinkai Zheng and Ruiliang Lyu and Shangqin Tu and Sheng Yang and Shengbiao Meng and Shi Zhong and Shiyu Huang and Shuyuan Zhao and Siyan Xue and Tianshu Zhang and Tianwei Luo and Tianxiang Hao and Tianyu Tong and Wei Jia and Wenkai Li and Xiao Liu and Xiaohan Zhang and Xin Lyu and Xinyu Zhang and Xinyue Fan and Xuancheng Huang and Yadong Xue and Yanfeng Wang and Yanling Wang and Yanzi Wang and Yifan An and Yifan Du and Yiheng Huang and Yilin Niu and Yiming Shi and Yu Wang and Yuan Wang and Yuanchang Yue and Yuchen Li and Yusen Liu and Yutao Zhang and Yuting Wang and Yuxuan Zhang and Zhao Xue and Zhengxiao Du and Zhenyu Hou and Zihan Wang and Peng Zhang and Debing Liu and Bin Xu and Juanzi Li and Minlie Huang and Yuxiao Dong and Jie Tang},
      year={2026},
      eprint={2507.01006},
      archivePrefix={arXiv},
      primaryClass={cs.CV},
      url={https://arxiv.org/abs/2507.01006}, 
}

@inproceedings{chen2025longvila-r1,
      title={Scaling RL to Long Videos},
      author={Yukang Chen and Wei Huang and Baifeng Shi and Qinghao Hu and Hanrong Ye and Ligeng Zhu and Zhijian Liu and Pavlo Molchanov and Jan Kautz and Xiaojuan Qi and Sifei Liu and Hongxu Yin and Yao Lu and Song Han},
      booktitle={Advances in Neural Information Processing Systems (NeurIPS)},
      year={2025},
}

@misc{coreteam2025mimovltechnicalreport,
      title={MiMo-VL Technical Report}, 
      author={LLM-Core-Team Xiaomi},
      year={2025},
      eprint={2506.03569},
      archivePrefix={arXiv},
      primaryClass={cs.CL},
      url={https://arxiv.org/abs/2506.03569}, 
}

@article{li2024anchored_gpt2_mcq_positional_bias,
  title={Anchored Answers: Unravelling Positional Bias in GPT-2's Multiple-Choice Questions},
  author={Li, Ruizhe and Gao, Yanjun},
  journal={arXiv preprint arXiv:2405.03205},
  year={2024}
}

@article{loginova2024_addressing_blind_guessing_selection_bias,
  title={Addressing Blind Guessing: Calibration of Selection Bias in Multiple-Choice Question Answering by Video Language Models},
  author={Loginova, Olga and Bezrukov, Oleksandr and Kravets, Alexey},
  journal={arXiv preprint arXiv:2410.14248},
  year={2024}
}

@article{singh2025_too_many_options_pitfalls_mcq,
  title={It is Too Many Options: Pitfalls of Multiple-Choice Questions in Generative AI and Medical Education},
  author={Singh, Shrutika and Alyakin, Anton and Alber, Daniel Alexander and Stryker, Jaden and Tong, Ai Phuong S and Sangwon, Karl and Goff, Nicolas and de la Paz, Mathew and Hernandez-Rovira, Miguel and Park, Ki Yun and others},
  journal={arXiv preprint arXiv:2503.13508},
  year={2025}
}

@inproceedings{molfese2025_right_answer_wrong_score_multiple_choice,
    title = "Right Answer, Wrong Score: Uncovering the Inconsistencies of {LLM} Evaluation in Multiple-Choice Question Answering",
    author = "Molfese, Francesco Maria  and
      Moroni, Luca  and
      Gioffr{\'e}, Luca  and
      Scir{\`e}, Alessandro  and
      Conia, Simone  and
      Navigli, Roberto",
    editor = "Che, Wanxiang  and
      Nabende, Joyce  and
      Shutova, Ekaterina  and
      Pilehvar, Mohammad Taher",
    booktitle = "Findings of the Association for Computational Linguistics: ACL 2025",
    month = jul,
    year = "2025",
    address = "Vienna, Austria",
    publisher = "Association for Computational Linguistics",
    url = "https://aclanthology.org/2025.findings-acl.950/",
    doi = "10.18653/v1/2025.findings-acl.950",
    pages = "18477--18494",
    ISBN = "979-8-89176-256-5"
}

@inproceedings{bavaresco2024llms,
    title = "{LLM}s instead of Human Judges? A Large Scale Empirical Study across 20 {NLP} Evaluation Tasks",
    author = "Bavaresco, Anna  and
      Bernardi, Raffaella  and
      Bertolazzi, Leonardo  and
      Elliott, Desmond  and
      Fern{\'a}ndez, Raquel  and
      Gatt, Albert  and
      Ghaleb, Esam  and
      Giulianelli, Mario  and
      Hanna, Michael  and
      Koller, Alexander  and
      Martins, Andre  and
      Mondorf, Philipp  and
      Neplenbroek, Vera  and
      Pezzelle, Sandro  and
      Plank, Barbara  and
      Schlangen, David  and
      Suglia, Alessandro  and
      Surikuchi, Aditya K  and
      Takmaz, Ece  and
      Testoni, Alberto",
    editor = "Che, Wanxiang  and
      Nabende, Joyce  and
      Shutova, Ekaterina  and
      Pilehvar, Mohammad Taher",
    booktitle = "Proceedings of the 63rd Annual Meeting of the Association for Computational Linguistics (Volume 2: Short Papers)",
    month = jul,
    year = "2025",
    address = "Vienna, Austria",
    publisher = "Association for Computational Linguistics",
    url = "https://aclanthology.org/2025.acl-short.20/",
    doi = "10.18653/v1/2025.acl-short.20",
    pages = "238--255",
    ISBN = "979-8-89176-252-7"
}

@inproceedings{
liu2025_is_your_vlm_reliable_judge,
title={Is Your Video Language Model a Reliable Judge?},
author={Ming Liu and Wensheng Zhang},
booktitle={The Thirteenth International Conference on Learning Representations},
year={2025},
url={https://openreview.net/forum?id=m8yby1JfbU}
}

@inproceedings{
ye2025justice,
title={Justice or Prejudice? Quantifying Biases in {LLM}-as-a-Judge},
author={Jiayi Ye and Yanbo Wang and Yue Huang and Dongping Chen and Qihui Zhang and Nuno Moniz and Tian Gao and Werner Geyer and Chao Huang and Pin-Yu Chen and Nitesh V Chawla and Xiangliang Zhang},
booktitle={The Thirteenth International Conference on Learning Representations},
year={2025},
url={https://openreview.net/forum?id=3GTtZFiajM}
}

@article{damonlpsg2023videollama,
  title = {Video-LLaMA: An Instruction-tuned Audio-Visual Language Model for Video Understanding},
  author = {Zhang, Hang and Li, Xin and Bing, Lidong},
  journal = {arXiv preprint arXiv:2306.02858},
  year = {2023},
  url = {https://arxiv.org/abs/2306.02858}
}

@article{li2023videochat,
  title={VideoChat: Chat-Centric Video Understanding},
  author={Li, KunChang and He, Yinan and Wang, Yi and Li, Yizhuo and Wang, Wenhai and Luo, Ping and Wang, Yali and Wang, Limin and Qiao, Yu},
  journal={arXiv preprint arXiv:2305.06355},
  year={2023}
}

@inproceedings{Maaz2023VideoChatGPT,
    title={Video-ChatGPT: Towards Detailed Video Understanding via Large Vision and Language Models},
    author={Maaz, Muhammad and Rasheed, Hanoona and Khan, Salman and Khan, Fahad Shahbaz},
    booktitle={Proceedings of the 62nd Annual Meeting of the Association for Computational Linguistics (ACL 2024)},
    year={2024}
}

@misc{luo2023valley,
      title={Valley: Video Assistant with Large Language model Enhanced abilitY},
      author={Ruipu Luo and Ziwang Zhao and Min Yang and Junwei Dong and Minghui Qiu and Pengcheng Lu and Tao Wang and Zhongyu Wei},
      year={2023},
      eprint={2306.07207},
      archivePrefix={arXiv},
      primaryClass={cs.CV}
}

@misc{xu2024pllava,
      title={PLLaVA : Parameter-free LLaVA Extension from Images to Videos for Video Dense Captioning}, 
      author={Lin Xu and Yilin Zhao and Daquan Zhou and Zhijie Lin and See Kiong Ng and Jiashi Feng},
      year={2024},
      eprint={2404.16994},
      archivePrefix={arXiv},
      primaryClass={cs.CV}
}

@misc{li2023llamavidimageworth2,
      title={LLaMA-VID: An Image is Worth 2 Tokens in Large Language Models}, 
      author={Yanwei Li and Chengyao Wang and Jiaya Jia},
      year={2023},
      eprint={2311.17043},
      archivePrefix={arXiv},
}

@misc{liu2023llava,
      title={Visual Instruction Tuning}, 
      author={Liu, Haotian and Li, Chunyuan and Wu, Qingyang and Lee, Yong Jae},
      publisher={NeurIPS},
      year={2023},
}

@misc{liu2024nvilaefficientfrontiervisual,
      title={NVILA: Efficient Frontier Visual Language Models}, 
      author={Zhijian Liu and Ligeng Zhu and Baifeng Shi and Zhuoyang Zhang and Yuming Lou and Shang Yang and Haocheng Xi and Shiyi Cao and Yuxian Gu and Dacheng Li and Xiuyu Li and Yunhao Fang and Yukang Chen and Cheng-Yu Hsieh and De-An Huang and An-Chieh Cheng and Vishwesh Nath and Jinyi Hu and Sifei Liu and Ranjay Krishna and Daguang Xu and Xiaolong Wang and Pavlo Molchanov and Jan Kautz and Hongxu Yin and Song Han and Yao Lu},
      year={2024},
      eprint={2412.04468},
      archivePrefix={arXiv},
}

@article{tschannen2025siglip,
  title={SigLIP 2: Multilingual Vision-Language Encoders with Improved Semantic Understanding, Localization, and Dense Features},
  author={Tschannen, Michael and Gritsenko, Alexey and Wang, Xiao and Naeem, Muhammad Ferjad and Alabdulmohsin, Ibrahim and Parthasarathy, Nikhil and Evans, Talfan and Beyer, Lucas and Xia, Ye and Mustafa, Basil and H\'enaff, Olivier and Harmsen, Jeremiah and Steiner, Andreas and Zhai, Xiaohua},
  year={2025},
  journal={arXiv preprint arXiv:2502.14786}
}

@InProceedings{clip,
  title = 	 {Learning Transferable Visual Models From Natural Language Supervision},
  author =       {Radford, Alec and Kim, Jong Wook and Hallacy, Chris and Ramesh, Aditya and Goh, Gabriel and Agarwal, Sandhini and Sastry, Girish and Askell, Amanda and Mishkin, Pamela and Clark, Jack and Krueger, Gretchen and Sutskever, Ilya},
  booktitle = 	 {Proceedings of the 38th International Conference on Machine Learning},
  pages = 	 {8748--8763},
  year = 	 {2021},
  editor = 	 {Meila, Marina and Zhang, Tong},
  volume = 	 {139},
  series = 	 {Proceedings of Machine Learning Research},
  month = 	 {18--24 Jul},
  publisher =    {PMLR},
  url = 	 {https://proceedings.mlr.press/v139/radford21a.html}
}

@inproceedings{
liu2025world,
title={World Model on Million-Length Video And Language With Blockwise RingAttention},
author={Hao Liu and Wilson Yan and Matei Zaharia and Pieter Abbeel},
booktitle={The Thirteenth International Conference on Learning Representations},
year={2025},
url={https://openreview.net/forum?id=HN8V0flwJF}
}

@article{phi3,
  publtype={informal},
  author={Marah I Abdin and Sam Ade Jacobs and Ammar Ahmad Awan and Jyoti Aneja and Ahmed Awadallah and Hany Awadalla and Nguyen Bach and Amit Bahree and Arash Bakhtiari and Harkirat S. Behl and Alon Benhaim and Misha Bilenko and Johan Bjorck and Sébastien Bubeck and Martin Cai and Caio César Teodoro Mendes and Weizhu Chen and Vishrav Chaudhary and Parul Chopra and Allie Del Giorno and Gustavo de Rosa and Matthew Dixon and Ronen Eldan and Dan Iter and Amit Garg and Abhishek Goswami and Suriya Gunasekar and Emman Haider and Junheng Hao and Russell J. Hewett and Jamie Huynh and Mojan Javaheripi and Xin Jin and Piero Kauffmann and Nikos Karampatziakis and Dongwoo Kim and Mahoud Khademi and Lev Kurilenko and James R. Lee and Yin Tat Lee and Yuanzhi Li and Chen Liang and Weishung Liu and Eric Lin and Zeqi Lin and Piyush Madan and Arindam Mitra and Hardik Modi and Anh Nguyen and Brandon Norick and Barun Patra and Daniel Perez-Becker and Thomas Portet and Reid Pryzant and Heyang Qin and Marko Radmilac and Corby Rosset and Sambudha Roy and Olatunji Ruwase and Olli Saarikivi and Amin Saied and Adil Salim and Michael Santacroce and Shital Shah and Ning Shang and Hiteshi Sharma and Xia Song and Masahiro Tanaka and Xin Wang and Rachel Ward and Guanhua Wang and Philipp Witte and Michael Wyatt and Can Xu and Jiahang Xu and Sonali Yadav and Fan Yang and Ziyi Yang and Donghan Yu and Chengruidong Zhang and Cyril Zhang and Jianwen Zhang and Li Lyna Zhang and Yi Zhang and Yue Zhang and Yunan Zhang and Xiren Zhou},
  title={Phi-3 Technical Report: A Highly Capable Language Model Locally on Your Phone},
  year={2024},
  cdate={1704067200000},
  journal={CoRR},
  volume={abs/2404.14219},
  url={https://doi.org/10.48550/arXiv.2404.14219}
}

@article{damonlpsg2025videollama3,
  title={VideoLLaMA 3: Frontier Multimodal Foundation Models for Image and Video Understanding},
  author={Boqiang Zhang and Kehan Li and Zesen Cheng and Zhiqiang Hu and Yuqian Yuan and Guanzheng Chen and Sicong Leng and Yuming Jiang and Hang Zhang and Xin Li and Peng Jin and Wenqi Zhang and Fan Wang and Lidong Bing and Deli Zhao},
  journal={arXiv preprint arXiv:2501.13106},
  year={2025},
  url={https://arxiv.org/abs/2501.13106}
}

@inproceedings{balazevic2024memory,
  title={Memory Consolidation Enables Long-Context Video Understanding},
  author={Balazevic, Ivana and Shi, Yuge and Papalampidi, Pinelopi and Chaabouni, Rahma and Koppula, Skanda and Henaff, Olivier J},
  booktitle={Forty-first International Conference on Machine Learning},
  year={2024}
}

@inproceedings{
santos2025inftyvideotrainingfreeapproachlong,
title={$\infty$-Video: A Training-Free Approach to Long Video Understanding via Continuous-Time Memory Consolidation},
author={Saul Santos and Ant{\'o}nio Farinhas and Daniel C McNamee and Andre Martins},
booktitle={Forty-second International Conference on Machine Learning},
year={2025},
url={https://openreview.net/forum?id=afDHwQ1ZDO}
}

@inproceedings{zhang2024simple,
  title={A Simple LLM Framework for Long-Range Video Question-Answering},
  author={Zhang, Ce and Lu, Taixi and Islam, Md Mohaiminul and Wang, Ziyang and Yu, Shoubin and Bansal, Mohit and Bertasius, Gedas},
  booktitle={Proceedings of the 2024 Conference on Empirical Methods in Natural Language Processing},
  pages={21715--21737},
  year={2024}
}

@inproceedings{wang2024videoagent,
  title={Videoagent: Long-form video understanding with large language model as agent},
  author={Wang, Xiaohan and Zhang, Yuhui and Zohar, Orr and Yeung-Levy, Serena},
  booktitle={European Conference on Computer Vision},
  pages={58--76},
  year={2024},
  organization={Springer}
}

@article{wang2024videotree,
  title={Videotree: Adaptive tree-based video representation for llm reasoning on long videos},
  author={Wang, Ziyang and Yu, Shoubin and Stengel-Eskin, Elias and Yoon, Jaehong and Cheng, Feng and Bertasius, Gedas and Bansal, Mohit},
  journal={arXiv preprint arXiv:2405.19209},
  year={2024}
}

@inproceedings{
wu2021star,
title={{STAR}: A Benchmark for Situated Reasoning in Real-World Videos},
author={Bo Wu and Shoubin Yu and Zhenfang Chen and Joshua B. Tenenbaum and Chuang Gan},
booktitle={Thirty-fifth Conference on Neural Information Processing Systems Datasets and Benchmarks Track (Round 2)},
year={2021},
url={https://openreview.net/forum?id=EfgNF5-ZAjM}
}

@inproceedings{yang2021justask,
title={Just Ask: Learning To Answer Questions From Millions of Narrated Videos},
author={Antoine Yang and Antoine Miech and Josef Sivic and Ivan Laptev and Cordelia Schmid},
booktitle={ICCV},
year={2021}}

@article{mangalam2023egoschema,
  title={Egoschema: A diagnostic benchmark for very long-form video language understanding},
  author={Karttikeya Mangalam and Raiymbek Akshulakov and Jitendra Malik},
  journal={arXiv preprint arXiv:2308.09126},
  year={2023}
}

@article{qiu2024egoplanbench2,
    title = {EgoPlan-Bench2: A Benchmark for Multimodal Large Language Model Planning in Real-World Scenarios},
    author = {Qiu, Lu and Chen, Yi and Ge, Yuying and Ge, Yixiao and Shan, Ying and Liu, Xihui},
    year = {2024},
    journal = {arXiv preprint arXiv:2412.04447}
}

@inproceedings{
parmar2024causalchaos,
title={CausalChaos! Dataset for Comprehensive Causal Action Question Answering Over Longer Causal Chains Grounded in Dynamic Visual Scenes},
author={Paritosh Parmar and Eric Peh and Ruirui Chen and Ting En Lam and Yuhan Chen and Elston Tan and Basura Fernando},
booktitle={The Thirty-eight Conference on Neural Information Processing Systems Datasets and Benchmarks Track},
year={2024},
url={https://openreview.net/forum?id=gP4aAi7q8S}
}

@article{hu2025videommmu,
  title={Video-mmmu: Evaluating knowledge acquisition from multi-discipline professional videos},
  author={Hu, Kairui and Wu, Penghao and Pu, Fanyi and Xiao, Wang and Zhang, Yuanhan and Yue, Xiang and Li, Bo and Liu, Ziwei},
  journal={arXiv preprint arXiv:2501.13826},
  year={2025}
}

@misc{vimul_2025,
      title={A Culturally-diverse Multilingual Multimodal Video Benchmark \& Model}, 
      author={Bhuiyan Sanjid Shafique and Ashmal Vayani and Muhammad Maaz and Hanoona Abdul Rasheed and Dinura Dissanayake and Mohammed Irfan Kurpath and Yahya Hmaiti and Go Inoue and Jean Lahoud and Md. Safirur Rashid and Shadid Intisar Quasem and Maheen Fatima and Franco Vidal and Mykola Maslych and Ketan Pravin More and Sanoojan Baliah and Hasindri Watawana and Yuhao Li and Fabian Farestam and Leon Schaller and Roman Tymtsiv and Simon Weber and Hisham Cholakkal and Ivan Laptev and Shin'ichi Satoh and Michael Felsberg and Mubarak Shah and Salman Khan and Fahad Shahbaz Khan},
      year={2025},
      eprint={2506.07032},
      archivePrefix={arXiv},
      primaryClass={cs.CL},
      url={https://arxiv.org/abs/2506.07032}, 
}

@misc{tan2024-synopground,
      title={SynopGround: A Large-Scale Dataset for Multi-Paragraph Video Grounding from TV Dramas and Synopses}, 
      author={Chaolei Tan and Zihang Lin and Junfu Pu and Zhongang Qi and Wei-Yi Pei and Zhi Qu and Yexin Wang and Ying Shan and Wei-Shi Zheng and Jian-Fang Hu},
      year={2024},
      eprint={2408.01669},
      archivePrefix={arXiv},
      primaryClass={cs.CV},
      url={https://arxiv.org/abs/2408.01669}, 
}

@misc{timescope2025blog,
  title        = {TimeScope: How Long Can Your Video Large Multimodal Model Go?},
  author       = {Orr Zohar and Rui Li and Andres Marafioti and Xiaohan Wang and Stanford AI Team and Hugging Face},
  year         = {2025},
  month        = {July},
  day          = {23},
  howpublished = {\url{https://huggingface.co/blog/timescope-video-lmm-benchmark}},
}

@inproceedings{yu2019activityqa,
    author = {Yu, Zhou and Xu, Dejing and Yu, Jun and Yu, Ting and Zhao, Zhou and Zhuang, Yueting and Tao, Dacheng},
    title = {ActivityNet-QA: A Dataset for Understanding Complex Web Videos via Question Answering},
    booktitle = {AAAI},
    pages = {9127--9134},
    year = {2019}
}

@inproceedings{bonomo-etal-2025-literaryqa,
    title = "{L}iterary{QA}: Towards Effective Evaluation of Long-document Narrative {QA}",
    author = "Bonomo, Tommaso  and
      Gioffr{\'e}, Luca  and
      Navigli, Roberto",
    editor = "Christodoulopoulos, Christos  and
      Chakraborty, Tanmoy  and
      Rose, Carolyn  and
      Peng, Violet",
    booktitle = "Proceedings of the 2025 Conference on Empirical Methods in Natural Language Processing",
    month = nov,
    year = "2025",
    address = "Suzhou, China",
    publisher = "Association for Computational Linguistics",
    url = "https://aclanthology.org/2025.emnlp-main.1729/",
    doi = "10.18653/v1/2025.emnlp-main.1729",
    pages = "34086--34107",
    ISBN = "979-8-89176-332-6"
}

@article{xu2024detectiveqa,
  title={Detectiveqa: Evaluating long-context reasoning on detective novels},
  author={Xu, Zhe and Ye, Jiasheng and Liu, Xiaoran and Liu, Xiangyang and Sun, Tianxiang and Liu, Zhigeng and Guo, Qipeng and Li, Linlin and Liu, Qun and Huang, Xuanjing and others},
  journal={arXiv preprint arXiv:2409.02465},
  year={2024}
}

@article{rawal2024cinepile,
  title={Cinepile: A long video question answering dataset and benchmark},
  author={Rawal, Ruchit and Saifullah, Khalid and Farr{\'e}, Miquel and Basri, Ronen and Jacobs, David and Somepalli, Gowthami and Goldstein, Tom},
  journal={arXiv preprint arXiv:2405.08813},
  year={2024}
}

@inproceedings{
ye2025mplugowl3,
title={m{PLUG}-Owl3: Towards Long Image-Sequence Understanding in Multi-Modal Large Language Models},
author={Jiabo Ye and Haiyang Xu and Haowei Liu and Anwen Hu and Ming Yan and Qi Qian and Ji Zhang and Fei Huang and Jingren Zhou},
booktitle={The Thirteenth International Conference on Learning Representations},
year={2025},
url={https://openreview.net/forum?id=pr37sbuhVa}
}

@inproceedings{LLaVAOneVision1.5,
  title={LLaVA-OneVision-1.5: Fully Open Framework for Democratized Multimodal Training},
  author={An, Xiang and Xie, Yin and Yang, Kaicheng and Zhang, Wenkang and Zhao, Xiuwei and Cheng, Zheng and Wang, Yirui and Xu, Songcen and Chen, Changrui and Wu, Chunsheng and Tan, Huajie and Li, Chunyuan and Yang, Jing and Yu, Jie and Wang, Xiyao and Qin, Bin and Wang, Yumeng and Yan, Zizhen and Feng, Ziyong and Liu, Ziwei and Li, Bo and Deng, Jiankang},
  booktitle={arXiv},  
  year={2025}
 }

@ARTICLE{llava_video,
       author = {{Zhang}, Yuanhan and {Wu}, Jinming and {Li}, Wei and {Li}, Bo and {Ma}, Zejun and {Liu}, Ziwei and {Li}, Chunyuan},
        title = "{Video Instruction Tuning With Synthetic Data}",
      journal = {arXiv e-prints},
         year = 2024,
        month = oct,
          eid = {arXiv:2410.02713},
        pages = {arXiv:2410.02713},
          doi = {10.48550/arXiv.2410.02713},
archivePrefix = {arXiv},
       eprint = {2410.02713},
 primaryClass = {cs.CV},
       adsurl = {https://ui.adsabs.harvard.edu/abs/2024arXiv241002713Z}
}

@misc{clark2026molmo2openweightsdata,
      title={Molmo2: Open Weights and Data for Vision-Language Models with Video Understanding and Grounding}, 
      author={Christopher Clark and Jieyu Zhang and Zixian Ma and Jae Sung Park and Mohammadreza Salehi and Rohun Tripathi and Sangho Lee and Zhongzheng Ren and Chris Dongjoo Kim and Yinuo Yang and Vincent Shao and Yue Yang and Weikai Huang and Ziqi Gao and Taira Anderson and Jianrui Zhang and Jitesh Jain and George Stoica and Winson Han and Ali Farhadi and Ranjay Krishna},
      year={2026},
      eprint={2601.10611},
      archivePrefix={arXiv},
      primaryClass={cs.CV},
      url={https://arxiv.org/abs/2601.10611}, 
}

@misc{microsoft2025phi4minitechnicalreportcompact,
      title={Phi-4-Mini Technical Report: Compact yet Powerful Multimodal Language Models via Mixture-of-LoRAs}, 
      author={Microsoft and : and Abdelrahman Abouelenin and Atabak Ashfaq and Adam Atkinson and Hany Awadalla and Nguyen Bach and Jianmin Bao and Alon Benhaim and Martin Cai and Vishrav Chaudhary and Congcong Chen and Dong Chen and Dongdong Chen and Junkun Chen and Weizhu Chen and Yen-Chun Chen and Yi-ling Chen and Qi Dai and Xiyang Dai and Ruchao Fan and Mei Gao and Min Gao and Amit Garg and Abhishek Goswami and Junheng Hao and Amr Hendy and Yuxuan Hu and Xin Jin and Mahmoud Khademi and Dongwoo Kim and Young Jin Kim and Gina Lee and Jinyu Li and Yunsheng Li and Chen Liang and Xihui Lin and Zeqi Lin and Mengchen Liu and Yang Liu and Gilsinia Lopez and Chong Luo and Piyush Madan and Vadim Mazalov and Arindam Mitra and Ali Mousavi and Anh Nguyen and Jing Pan and Daniel Perez-Becker and Jacob Platin and Thomas Portet and Kai Qiu and Bo Ren and Liliang Ren and Sambuddha Roy and Ning Shang and Yelong Shen and Saksham Singhal and Subhojit Som and Xia Song and Tetyana Sych and Praneetha Vaddamanu and Shuohang Wang and Yiming Wang and Zhenghao Wang and Haibin Wu and Haoran Xu and Weijian Xu and Yifan Yang and Ziyi Yang and Donghan Yu and Ishmam Zabir and Jianwen Zhang and Li Lyna Zhang and Yunan Zhang and Xiren Zhou},
      year={2025},
      eprint={2503.01743},
      archivePrefix={arXiv},
      primaryClass={cs.CL},
      url={https://arxiv.org/abs/2503.01743}, 
}

@misc{li2025ariaopenmultimodalnative,
      title={Aria: An Open Multimodal Native Mixture-of-Experts Model}, 
      author={Dongxu Li and Yudong Liu and Haoning Wu and Yue Wang and Zhiqi Shen and Bowen Qu and Xinyao Niu and Fan Zhou and Chengen Huang and Yanpeng Li and Chongyan Zhu and Xiaoyi Ren and Chao Li and Yifan Ye and Peng Liu and Lihuan Zhang and Hanshu Yan and Guoyin Wang and Bei Chen and Junnan Li},
      year={2025},
      eprint={2410.05993},
      archivePrefix={arXiv},
      primaryClass={cs.CV},
      url={https://arxiv.org/abs/2410.05993}, 
}

@inproceedings{faure-etal-2025-moviecore,
    title = "{M}ovie{CORE}: {CO}gnitive {RE}asoning in Movies",
    author = "Faure, Gueter Josmy  and
      Chen, Min-Hung  and
      Yeh, Jia-Fong  and
      Cheng, Ying  and
      Su, Hung-Ting  and
      Tang, Yung-Hao  and
      Lai, Shang-Hong  and
      Hsu, Winston H.",
    editor = "Christodoulopoulos, Christos  and
      Chakraborty, Tanmoy  and
      Rose, Carolyn  and
      Peng, Violet",
    booktitle = "Proceedings of the 2025 Conference on Empirical Methods in Natural Language Processing",
    month = nov,
    year = "2025",
    address = "Suzhou, China",
    publisher = "Association for Computational Linguistics",
    url = "https://aclanthology.org/2025.emnlp-main.66/",
    doi = "10.18653/v1/2025.emnlp-main.66",
    pages = "1253--1272",
    ISBN = "979-8-89176-332-6"
}

\clearpage
\appendix


\section{Data Annotation Details}\label{app:details_dataset}

\paragraph{Movie Collection Metadata}
We collected 53 full-length movies from the Internet Archive\footnote{\url{https://archive.org}}, focusing on titles released under the Public Domain 1.0 license to ensure legal reusability and support open-access research. Original-language subtitles, mostly in English, were obtained from OpenSubtitles.org\footnote{\url{https://www.opensubtitles.org}}. For one movie without available subtitles, we used whisper-1 \citep{radford2023robust_whisper}\footnote{\url{https://platform.openai.com/docs/models/whisper-1}}
to generate a transcript and manually post-edited to ensure high quality. Each movie includes audio and aligned subtitles, with an average duration of 88 minutes. In~\cref{tab:movie_details_1,tab:movie_details_2}, we provide detailed information on the 53 released movies, including their genre, original language, and duration. Collectively, the movies span a wide temporal range of approximately five decades (1920–1970) and cover a diverse set of genres. 

\paragraph{Annotator Demographics}
All annotations were performed manually by 26 co-authors from 12 institutions across 7 countries, with a roughly balanced distribution of career stages (PhD students, postdocs, and faculty members) and gender. We did not rely on crowdsourcing or semi-automatic approaches, as constructing claims about central narrative events—often requiring reasoning across multiple scenes—demands careful human judgment. Having co-authors perform the annotations also provided strong intrinsic motivation and accountability, ensuring high-quality, consistent work that might not be achievable with anonymous or paid crowdworkers. Detailed guidelines, multiple rounds of discussion, and collective revisions were used to ensure consistency and high quality, while annotators were not informed of any specific hypotheses or expected outcomes, reducing the risk of bias.

\paragraph{Annotated Claim Pairs} Figure~\ref{app:fig:claim_pairs_distribution} shows the number of  annotated claim pairs per movie.

\paragraph{Comprehension Dimensions}
Each claim pair was labeled according to one or more comprehension dimensions, which capture specific aspects of narrative understanding. These dimensions are informed by prior work \citep{xiao2021next,zhang2023movqabenchmarkversatilequestionanswering,wang2025lvbench} and are defined in~\cref{tab:comprehension-dimensions}.

\definecolor{gray}{RGB}{214, 204, 194}  

\begin{figure*}[h]
\centering
\begin{tikzpicture}
\begin{axis}[
    ybar,
    bar width=0.08cm,           
    width=\textwidth,           
    height=4cm,
    ymin=0,
    axis y line=none,
    axis x line*=bottom,
    xtick={1,3,...,53},         
    xticklabel style={
        rotate=0,
        anchor=center,          
        font=\scriptsize
    },
    nodes near coords,           
    every node near coord/.style={
        font=\scriptsize,     
        rotate=0,
        anchor=south,           
        yshift=2pt
    },
    enlarge x limits=0.02        
]

\addplot[fill=gray] coordinates {
    (1,12) (2,20) (3,11) (4,6) (5,15) (6,12) (7,22) (8,15) (9,8) (10,18)
    (11,25) (12,22) (13,20) (14,16) (15,26) (16,19) (17,19) (18,19) (19,15) (20,18)
    (21,12) (22,15) (23,17) (24,20) (25,15) (26,20) (27,22) (28,20) (29,19) (30,20)
    (31,8) (32,7) (33,14) (34,15) (35,15) (36,13) (37,16) (38,17) (39,22) (40,17)
    (41,11) (42,13) (43,8) (44,15) (45,17) (46,21) (47,14) (48,19) (49,26) (50,19)
    (51,8) (52,20) (53,15)
};
\end{axis}
\end{tikzpicture}
\caption{Number of claim pairs per movie.}
\label{app:fig:claim_pairs_distribution}
\end{figure*}

\begin{table*}[t]
\centering
\begin{tabularx}{\textwidth}{l X}  
\toprule
\textbf{Comprehension Dimension} & \textbf{Definition} \\
\midrule
\textit{Event/Entity Understanding} & Involves identifying key entities (e.g., people, places, or objects) and understanding the events they participate in. This includes tracking entities across scenes, interpreting their roles, and recognizing their interactions and relationships throughout the narrative.\\
\addlinespace
\textit{Temporal Perception} & Requires reasoning about the timeline of events---determining whether actions occur before, after, or simultaneously---and may also include counting or sequencing events. 
The focus is on broader temporal relationships within the narrative.\\
\addlinespace
\textit{Emotion Understanding} & Involves recognizing the emotional states of characters and interpreting how these emotions evolve throughout the story.\\
\addlinespace
\textit{Causal Reasoning} & Focuses on identifying cause-and-effect relationships between events or actions, including both explicit and implicit dependencies that may span multiple scenes.\\
\bottomrule
\end{tabularx}
\caption{Definitions of comprehension dimensions.}
\label{tab:comprehension-dimensions}
\end{table*}

\section{Pairwise Accuracy vs Simple Accuracy}\label{app:paiwise_acc_vs_acc}
Our evaluation protocol is motivated as follows: multiple-choice settings introduce distractor-driven biases, where models exploit superficial cues or distractor order~\citep{molfese2025_right_answer_wrong_score_multiple_choice}. Following prior work on narrative understanding~\citep{karpinska-etal-2024-one}, we evaluate each minimally differing claim independently, removing distractors and reducing such shortcut behaviors. Under this setup, pairwise accuracy is stricter and more informative; a model is counted as correct only if it predicts both labels accurately, minimizing the chance that it appears “correct for the wrong reason”~\citep{karpinska-etal-2024-one}. In contrast, in simple accuracy each prediction contributes independently to the score making it less suitable for our case and more suitable for multiple-choice formats, which we intentionally avoid for the aforementioned reasons.
 In our evaluation, the relationship between the two predictions is essential. \textbf{The model must implicitly distinguish what truly happened from a minimally edited alternative that is also plausible, even though it never sees the two statements jointly during inference.} That said, the evaluation depends on the relationship between the two predictions, not the predictions alone.

As a further illustration of why interpreting results based only on simple accuracy can be misleading, we report in Table~\ref{tab:simple_vs_acc} several metrics (pairwise accuracy, simple accuracy, accuracy on facts and accuracy on fibs) for some of the open-weight models and for Gemini 2.5 Pro, across both the video-only and video with subtitles settings. \textit{Simple accuracy can appear high simply because the model performs well on the fibs, while failing on the facts}, highlighting a bias towards predicting False. Taken together \textbf{these considerations motivate our choice of pairwise accuracy as the primary evaluation metric}. Nevertheless, we also report accuracy for completeness.

\begin{table*}[t]
\centering
\label{tab:model_performance}
\begin{tabular}{lcccc}
\toprule
\textbf{Model} & \textbf{Pairwise Acc} & \textbf{Average Acc} & \textbf{Acc on Facts} & \textbf{Acc on Fibs} \\
\midrule
\multicolumn{5}{c}{\textit{Video}} \\
\midrule
Gemini 2.5 Pro     & 37.9  & 64.4  & 47.4  & 81.5  \\
Gemma3-27B      & 31.5  & 61.2  & 57.4  & 65.1  \\
Qwen3VL-IT-30B     & 19.2  & 56.9  & 24.3  & 89.5  \\
InternVL3.5-38B  & 26.6  & 60.0  & 38.8  & 81.1  \\
Qwen2.5VL-72B   & 29.7 & 58.8 & 45.4 & 72.1  \\
InternVL3-78B   & 22.1 & 58.0 & 37.7 & 78.3  \\
\midrule
\multicolumn{5}{c}{\textit{Video+Subtitles}} \\
\midrule
Gemini 2.5 Pro      & 63.8  & 79.8  & 72.2  & 87.3  \\
Gemma3-27B      & 39.7  & 65.3  & 57.6  & 73.0  \\
Qwen3VL-IT-30B     & 43.8  & 69.5  & 53.0  & 86.0  \\
InternVL3.5-38B  & 46.2  & 70.3  & 61.6  & 79.0  \\
Qwen2.5VL-72B   & 47.5 & 69.0  & 59.3  & 78.7  \\
InternVL3-78B   & 51.3  & 72.7 & 67.5  & 77.9  \\
\bottomrule
\end{tabular}
\caption{Reporting pairwise accuracy, average accuracy, accuracy on facts and accuracy on fibs for some open-weight models and Gemini2.5. Results highlight that simple accuracy on its own can be misleading.}
\label{tab:simple_vs_acc} 
\end{table*}

\section{Mid and Large scale Model Analysis}\label{appendix:ablation_studies}

\paragraph{Beyond vision and subtitles: the role of summarized content (synopsis) and pretrained (world) knowledge.}
Table~\ref{app:tab:big_model_synopsis_analysis} reports an ablation study examining the extent to which models rely on textual information beyond subtitles. When conditioned only on the movie title and release year, performance remains at or only slightly above chance level, indicating that parametric knowledge acquired during pretraining contributes only marginally to the task, unlike the stronger effect observed for Gemini 2.5 Pro as reported in the main text. Likewise, providing only the movie synopsis results to near-random performance for most models, suggesting that high-level textual summaries alone are insufficient for reliable prediction. In both settings, performance remains far below human level, indicating that \textit{neither memorized knowledge nor condensed narrative summaries are sufficient to successfully solve the task}.

\begin{table*}[t]
\centering
\begin{tabular}{lrrrrr}
\toprule
\textbf{Model}  & \textbf{Movie Title} & \textbf{Synopsis} & \textbf{Subs} & \textbf{Video} & \textbf{Video+Subs} \\
\midrule
Gemini 2.5 Pro   & 45.4 & 27.9 & 58.9 & 37.9 & 63.8   \\
Gemma3-27B        &  28.1 & 29.6 & 42.3 & 31.5 &  39.7  \\
Qwen3VL-IT-30B       &  12.9 & 26.6 & 33.5 & 19.2 &  43.8  \\
InternVL3.5-38B    &  27.0 & 31.8 & 47.5 & 26.6 &  46.2  \\
Qwen2.5-VL-72B    &  15.2 & 25.7 & 37.6 & 29.7 &  47.5  \\
LLaVA-Video-72B & 17.1 & 38.7 & 41.7 & 15.6 &  41.8  \\
InternVL3-78B   &  30.4 & 30.4 & 49.0 & 22.1 & 51.3  \\
\bottomrule
\end{tabular}
\caption{Extended version of Table~\ref{tab:gemini_ablation} reporting pairwise performance for Gemini 2.5 Pro and mid- to large-scale open-weight models across different input modalities. \textit{Video} uses only the visual stream; \textit{Subs} includes only subtitle information; \textit{Synopsis} relies only on the synopsis of the movie obtained from Wikipedia; \textit{Video+Subs} combines both visual and subtitles inputs; and \textit{Movie Title} uses only the claim, along with the movie title and release year, without access to movie content.}
\label{app:tab:big_model_synopsis_analysis}
\end{table*}

\begin{figure*}[h]
\centering
\resizebox{\textwidth}{!}{ 
\begin{tikzpicture}

\begin{groupplot}[
    group style={
        group name=barplots,
        group size=7 by 1,
        horizontal sep=0.27cm,
        xlabels at=edge bottom
    },
    width=3.2cm,
    height=4cm,
    legend style={
        at={(3.8,-0.1)},
        anchor=north,
        font=\small,
        legend columns=6,
        /tikz/every even column/.append style={column sep=1cm}
    },
    ybar,
    ymin=0, ymax=95, 
    enlarge x limits=15,
    axis lines=none,
    xtick=data,
    symbolic x coords={ModA, ModB, ModC, ModD, ModE, ModF},
    xticklabels={}, 
    title style={yshift=2mm, font=\small, text depth=0.5ex}, 
    nodes near coords,
    nodes near coords align={vertical},
]

\nextgroupplot[title={Event/Entity}]
\addplot+[
    fill=c1,
    draw=c1b,
    bar width=5.5pt,
    nodes near coords,
    point meta=explicit symbolic,
    font=\scriptsize,
    text=black
] coordinates {
    (ModA, 43.3) [43]
};
\addlegendentry{LLaVA-Video}

\addplot+[
    fill=c2,
    draw=c2b,
    bar width=5.5pt,
    nodes near coords,
    point meta=explicit symbolic,
    font=\scriptsize,
    text=black
] coordinates {
    (ModB,52.9) [53]
};
\addlegendentry{InternVL3}

\addplot+[
    fill=c3,
    draw=c3b,
    bar width=5.5pt,
    nodes near coords,
    point meta=explicit symbolic,
    text=black,
    font=\scriptsize,
] coordinates {
    (ModC,46.9) [47]
};
\addlegendentry{Qwen2.5-VL}

\addplot+[
    fill=c4,
    draw=c4b,
    bar width=5.5pt,
    nodes near coords,
    point meta=explicit symbolic,
    text=black,
    font=\scriptsize,
] coordinates {
    (ModD,50.5) [51]
};
\addlegendentry{GPT-4o}

\addplot+[
    fill=c5,
    draw=c5b,
    bar width=5.5pt,
    nodes near coords,
    point meta=explicit symbolic,
    text=black,
    font=\scriptsize,
] coordinates {
    (ModE,63) [63]
};
\addlegendentry{Gemini 2.5 Pro}

\addplot+[
    fill=c6,,
    draw=c6b,
    bar width=5.5pt,
    nodes near coords,
    point meta=explicit symbolic,
    text=black,
    font=\scriptsize,
] coordinates {
    (ModF,83) [83]
};
\addlegendentry{Human}

\nextgroupplot[title={Temporal}]
\addplot+[
    fill=c1,
    draw=c1b,
    bar width=5.5pt,
    nodes near coords,
    point meta=explicit symbolic,
    font=\scriptsize,
    text=black
] coordinates {
    (ModA, 33) [33]
};

\addplot+[
    fill=c2,
    draw=c2b,
    bar width=5.5pt,
    nodes near coords,
    point meta=explicit symbolic,
    font=\scriptsize,
    text=black
] coordinates {
    (ModB,39.5) [40]
};

\addplot+[
    fill=c3,
    draw=c3b,
    bar width=5.5pt,
    nodes near coords,
    point meta=explicit symbolic,
    text=black,
    font=\scriptsize,
] coordinates {
    (ModC,50.8) [51]
};

\addplot+[
    fill=c4,
    draw=c4b,
    bar width=5.5pt,
    nodes near coords,
    point meta=explicit symbolic,
    text=black,
    font=\scriptsize,
] coordinates {
    (ModD,44.7) [45]
};

\addplot+[
    fill=c5,
    draw=c5b,
    bar width=5.5pt,
    nodes near coords,
    point meta=explicit symbolic,
    text=black,
    font=\scriptsize,
] coordinates {
    (ModE,64) [64]
};

\addplot+[
    fill=c6,,
    draw=c6b,
    bar width=5.5pt,
    nodes near coords,
    point meta=explicit symbolic,
    text=black,
    font=\scriptsize,
] coordinates {
    (ModF,80) [80]
};
\nextgroupplot[title={Causal}]
\addplot+[
    fill=c1,
    draw=c1b,
    bar width=5.5pt,
    nodes near coords,
    point meta=explicit symbolic,
    font=\scriptsize,
    text=black
] coordinates {
    (ModA, 40.6) [41]
};

\addplot+[
    fill=c2,
    draw=c2b,
    bar width=5.5pt,
    nodes near coords,
    point meta=explicit symbolic,
    font=\scriptsize,
    text=black
] coordinates {
    (ModB,48.3) [48]
};

\addplot+[
    fill=c3,
    draw=c3b,
    bar width=5.5pt,
    nodes near coords,
    point meta=explicit symbolic,
    text=black,
    font=\scriptsize,
] coordinates {
    (ModC,51.6) [52]
};

\addplot+[
    fill=c4,
    draw=c4b,
    bar width=5.5pt,
    nodes near coords,
    point meta=explicit symbolic,
    text=black,
    font=\scriptsize,
] coordinates {
    (ModD,49.7) [50]
};

\addplot+[
    fill=c5,
    draw=c5b,
    bar width=5.5pt,
    nodes near coords,
    point meta=explicit symbolic,
    text=black,
    font=\scriptsize,
] coordinates {
    (ModE,63) [63]
};

\addplot+[
    fill=c6,,
    draw=c6b,
    bar width=5.5pt,
    nodes near coords,
    point meta=explicit symbolic,
    text=black,
    font=\scriptsize,
] coordinates {
    (ModF,84) [84]
};
\nextgroupplot[title={Emotion}]
\addplot+[
    fill=c1,
    draw=c1b,
    bar width=5.5pt,
    nodes near coords,
    point meta=explicit symbolic,
    font=\scriptsize,
    text=black
] coordinates {
    (ModA, 33.6) [34]
};

\addplot+[
    fill=c2,
    draw=c2b,
    bar width=5.5pt,
    nodes near coords,
    point meta=explicit symbolic,
    font=\scriptsize,
    text=black
] coordinates {
    (ModB,48.1) [48]
};

\addplot+[
    fill=c3,
    draw=c3b,
    bar width=5.5pt,
    nodes near coords,
    point meta=explicit symbolic,
    text=black,
    font=\scriptsize,
] coordinates {
    (ModC,43.6) [44]
};

\addplot+[
    fill=c4,
    draw=c4b,
    bar width=5.5pt,
    nodes near coords,
    point meta=explicit symbolic,
    text=black,
    font=\scriptsize,
] coordinates {
    (ModD,53.6) [54]
};

\addplot+[
    fill=c5,
    draw=c5b,
    bar width=5.5pt,
    nodes near coords,
    point meta=explicit symbolic,
    text=black,
    font=\scriptsize,
] coordinates {
    (ModE,71) [71]
};
\addplot+[
    fill=c6,,
    draw=c6b,
    bar width=5.5pt,
    nodes near coords,
    point meta=explicit symbolic,
    text=black,
    font=\scriptsize,
] coordinates {
    (ModF,90) [90]
};
\nextgroupplot[]
\addplot+[
    fill=white,
    draw=white,
    bar width=5.5pt,
    nodes near coords,
    point meta=explicit symbolic,
    font=\scriptsize,
    text=black
] coordinates {
    (ModA, 15) []
};

\nextgroupplot[title={Single}]
\addplot+[
    fill=c1,
    draw=c1b,
    bar width=5.5pt,
    nodes near coords,
    point meta=explicit symbolic,
    font=\scriptsize,
    text=black
] coordinates {
    (ModA, 42.9) [43]
};

\addplot+[
    fill=c2,
    draw=c2b,
    bar width=5.5pt,
    nodes near coords,
    point meta=explicit symbolic,
    font=\scriptsize,
    text=black
] coordinates {
    (ModB,53.1) [53]
};

\addplot+[
    fill=c3,
    draw=c3b,
    bar width=5.5pt,
    nodes near coords,
    point meta=explicit symbolic,
    text=black,
    font=\scriptsize,
] coordinates {
    (ModC,47.0) [47]
};

\addplot+[
    fill=c4,
    draw=c4b,
    bar width=5.5pt,
    nodes near coords,
    point meta=explicit symbolic,
    text=black,
    font=\scriptsize,
] coordinates {
    (ModD,49.6) [49]
};

\addplot+[
    fill=c5,
    draw=c5b,
    bar width=5.5pt,
    nodes near coords,
    point meta=explicit symbolic,
    text=black,
    font=\scriptsize,
] coordinates {
    (ModE,63) [63]
};
\addplot+[
    fill=c6,,
    draw=c6b,
    bar width=5.5pt,
    nodes near coords,
    point meta=explicit symbolic,
    text=black,
    font=\scriptsize,
] coordinates {
    (ModF,85) [85]
};
\nextgroupplot[title={Multi/Global}]
\addplot+[
    fill=c1,
    draw=c1b,
    bar width=5.5pt,
    nodes near coords,
    point meta=explicit symbolic,
    font=\scriptsize,
    text=black
] coordinates {
    (ModA, 40.7) [41]
};

\addplot+[
    fill=c2,
    draw=c2b,
    bar width=5.5pt,
    nodes near coords,
    point meta=explicit symbolic,
    font=\scriptsize,
    text=black
] coordinates {
    (ModB,49.4) [49]
};

\addplot+[
    fill=c3,
    draw=c3b,
    bar width=5.5pt,
    nodes near coords,
    point meta=explicit symbolic,
    text=black,
    font=\scriptsize,
] coordinates {
    (ModC,47.8) [48]
};

\addplot+[
    fill=c4,
    draw=c4b,
    bar width=5.5pt,
    nodes near coords,
    point meta=explicit symbolic,
    text=black,
    font=\scriptsize,
] coordinates {
    (ModD,49.9) [50]
};

\addplot+[
    fill=c5,
    draw=c6b,
    bar width=5.5pt,
    nodes near coords,
    point meta=explicit symbolic,
    text=black,
    font=\scriptsize,
] coordinates {
    (ModE,64.6) [65]
};

\addplot+[
    fill=c6,,
    draw=c6b,
    bar width=5.5pt,
    nodes near coords,
    point meta=explicit symbolic,
    text=black,
    font=\scriptsize,
] coordinates {
    (ModF,84) [84]
};
\end{groupplot}

\node[font=\small, above=0.8cm] 
at ($(barplots c1r1.north west)!0.5!(barplots c4r1.north east)$) 
{Comprehension Dimension};

\node[font=\small, above=0.8cm,xshift=8mm] 
at ($(barplots c5r1.north west)!0.5!(barplots c7r1.north east)$) 
{Reasoning Granularity};
\end{tikzpicture}
}
\captionsetup{labelfont=bf} 
\caption{Pairwise accuracy for large-scale models with video and subtitles, and human baseline per comprehension dimension and reasoning granularity.}
\label{fig:big_models_ablation}
\end{figure*}

\paragraph{A fine-grained view of large-scale model performance across comprehension dimensions and reasoning granularities.}
Fig.~\ref{fig:big_models_ablation} shows that, among the large-scale models, Gemini 2.5 Pro still demonstrates inferior performance, ranking second to humans in various categories. Other models like LLaVA-Video and InternVL3 generally show lower scores, suggesting areas for improvement. The results also highlight varying degrees of difficulty across the tasks, with emotion comprehension appearing to be a strong point for humans and models, while temporal perception is the weakest for both. Interestingly, the analysis on reasoning granularity reveals an interesting pattern between humans and models: as reasoning shifts from single-scene to multi-scene/global, model performance tends to decrease for most models, while human performance remains stable.

\section{Detailed Guidelines for Data Annotation and Human Evaluation}\label{appendix:guidelines_human}

\subsection{Data Annotation Guidelines}
Figures~\ref{fig:guidelines_data_annotation_part1} and~\ref{fig:guidelines_data_annotation_part2} present the detailed guidelines provided to annotators during the data annotation process. These include instructions for constructing contrastive claim pairs, and labeling each pair with the appropriate reasoning granularity and  comprehension dimensions. Furthermore, in~\cref{fig:guidelines_data_annotation_examples_reasoning,fig:guidelines_data_annotation_examples_comprehension_dims}, we include a subset of illustrative examples shown to annotators to guide their annotations of reasoning granularity and comprehension dimensions, respectively.

We note that among the comprehension dimensions annotators could assign to each claim pair, an ``Other'' category was included to account for cases that did not clearly align with any of the predefined dimensions. As this label was selected rarely (0.49\% of the data), it is excluded from the figures presented in the main text. Additionally, although annotators were instructed to distinguish between single-scene, multi-scene, and global reasoning granularities, annotator feedback indicated that the distinction between multi-scene and global reasoning was not always clear-cut in practice, with substantial overlap between the two categories. Based on this feedback, we therefore merge them into a single category (multi-scene/global) in the analysis reported in the main text.

\paragraph{Consent and Data Use.} We confirm that all contributors whose data were used in this project provided informed consent prior to participation. The data consist solely of contributions created for benchmark construction; no personal data were collected. Contributors were informed that their contributions would be released under a CC-BY 4.0 license.

\subsection{Human Evaluation Guidelines}
\begin{figure*}
    \centering
    \begin{tcolorbox}[title={Guidelines for Human Evaluation (Part 2)}, colback=gray!2, colframe=black!30, arc=1mm,
boxrule=0.4pt, width=0.95\textwidth, fonttitle=\bfseries, fontupper=\small]

\paragraph{Stage 2:}
\noindent
\\
Once you complete Stage 1, you will see a message asking you if you want to proceed to Stage 2 (Stage 2 is optional).\\

During Stage 2, you will be shown again with the choices you selected during Stage 1, but now \textbf{you can revise your answers by looking back to the movie} (you can reuse the movie link we provided you). You will be shown for each claim with the choices you did in Stage 1. You are free to change them and proceed to the next claims. Don’t worry your answers will not be overwritten. Once you finish with Stage 2, you will be shown with a confirmation message.
\\
\\
If you have any questions or encounter any technical issues, please report them to our team! Thank you for your time and effort!
\noindent
\\

\begin{center}
    \includegraphics[width=\textwidth]{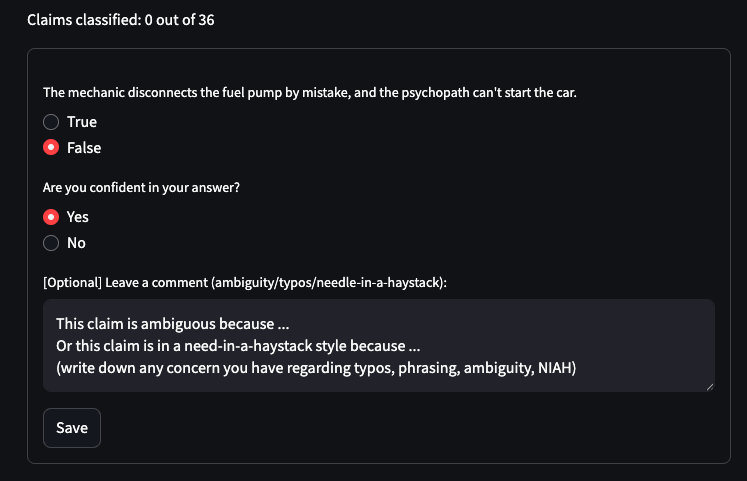}\\
    \small{Illustration of the human evaluation interface.}
\end{center}

\end{tcolorbox}
    \caption{Guidelines provided for human evaluation (Part 2).}
    \label{fig:guidelines_human_eval_part2}
\end{figure*}

In~\cref{fig:guidelines_human_eval_part1,fig:guidelines_human_eval_part2}, we provide the full set of guidelines shared to participants during the human evaluation process, which consists of two stages: an initial stage in which evaluators respond without revisiting the movie, and an optional second stage that allows revisiting. While we only analyze the results from Stage 1---as our goal is to assess movie understanding based on memorable events without allowing participants to rewatch parts of the film---we include the complete instructions for both stages to offer full context. Additionally, we provide an illustration of the evaluation interface to clarify the evaluation setup.

\section{Details on Experimental Setup}
\label{app:experimental_settings}

\subsection{Frame selection}
For all models except GPT-4o, we first downsample videos to 1 frame per second. From these frames, we then uniformly sample a subset, adjusting the number of frames based on each model’s maximum context length and original training settings. For GPT-4o, frames are uniformly sampled directly from the original videos without prior downsampling. The exact number of frames sampled for each model is reported in Table ~\ref{tab:frames_overview}.

\begin{table*}[t]
\centering
\small
\renewcommand{\arraystretch}{1.05}

\begin{tabular*}{\textwidth}{@{\extracolsep{\fill}}l c c c}
\toprule

\textbf{Model} & \textbf{\#Params} & \textbf{\#Frames} & \textbf{HF Checkpoint} \\

\midrule
\multicolumn{4}{c}{\textit{\textbf{Closed Models}}} \\
\midrule

GPT-4o \citep{openai2024gpt4ocard} & - & 50 & -- \\
Gemini 2.5 Pro \citep{team2023gemini} & - & 120 & -- \\

\midrule
\multicolumn{4}{c}{\textit{\textbf{Open-weight Models}}} \\
\midrule

mPLUG-Owl3* \citep{ye2025mplugowl3} & 7B & -- & mPLUG/mPLUG-Owl3-7B-240728 \\
VideoLLaMA3 \citep{damonlpsg2025videollama3} & 7B & 180 & DAMO-NLP-SG/VideoLLaMA3-7B \\
Qwen2.5-VL \citep{Qwen2.5-VL} & 7B & 1024 & Qwen/Qwen2.5-VL-7B-Instruct \\
LLaVA-Video \citep{llava_video} & 7B & 64 & lmms-lab/LLaVA-Video-7B-Qwen2 \\
LongVILA-R1 \citep{chen2025longvila-r1} & 7B & 180 & Efficient-Large-Model/LongVILA-R1-7B \\
MiMo-VL-7B-RL \citep{coreteam2025mimovltechnicalreport} & 7B & 256 & XiaomiMiMo/MiMo-VL-7B-RL \\
Kangaroo* \citep{kangaroogroup} & 8B & 24 & KangarooGroup/kangaroo \\
InternVL3.5 \citep{internvl3} & 8B & 64 & OpenGVLab/InternVL3\_5-8B \\
LLaVAOneVision1.5 \citep{LLaVAOneVision1.5} & 8B & 32 & lmms-lab/LLaVA-One-Vision-1.5-8B-Instruct \\
Molmo2 \citep{clark2026molmo2openweightsdata} & 8B & 256 & allenai/Molmo2-8B \\
Phi4* \citep{microsoft2025phi4minitechnicalreportcompact} & 14B & 256 & microsoft/Phi-4-multimodal-instruct \\
Aria \citep{li2025ariaopenmultimodalnative} & 25B(3.9B) & 64 & rhymes-ai/Aria \\
Gemma3 \citep{gemmateam2025gemma3technicalreport} & 27B & 180 & google/gemma-3-27b-it \\
Qwen3VL-IT \citep{Qwen3-VL} & 30B & 1024 & Qwen/Qwen3-VL-30B-A3B-Instruct \\
Ovis2 \citep{Ovis} & 34B & 10 & AIDC-AI/Ovis2-34B \\
InternVL3.5 \citep{wang2025internvl35advancingopensourcemultimodal} & 38B & 64 & OpenGVLab/InternVL3\_5-38B-Instruct \\
Qwen2.5-VL \citep{Qwen2.5-VL} & 72B & 1024 & Qwen/Qwen2.5-VL-72B-Instruct \\
LLaVA-Video\citep{llava_video} & 72B & 64 & lmms-lab/LLaVA-Video-72B-Qwen2 \\
InternVL3 \citep{internvl3} & 78B & 64 & OpenGVLab/InternVL3-78B \\
GLM-4.5V-FP8 \citep{vteam2026glm45vglm41vthinkingversatilemultimodal} & 108B & 512 & zai-org/GLM-4.5V-FP8 \\

\bottomrule
\end{tabular*}
\caption{Overview of the evaluated vision-language models, including parameter scale, number of input frames, and hf checkpoint.}
\label{tab:frames_overview}
\end{table*}

\subsection{Prompt Templates \& Answer Extraction}
\label{appendix:prompts}
In~\cref{fig:direct_prompt_template,fig:explanation_prompt_template} we present the direct and explanation prompt templates used for open-weight and closed models, respectively. The former  requests only a True/False response, while the latter additionally asks for a brief justification before the final answer. We found that the direct prompt yielded better performance for open-weight models, while the explanation prompt proved more effective for closed models. When experimenting with different input modalities---such as adding the synopsis, subtitles, or movie title---we adapt the prompts accordingly. To extract predictions, we use regular expressions to identify True/False answers in the model outputs, selecting the last valid match.

\subsection{Resources}
\label{appendix:resources}
Our infrastructure consists of a single machine equipped with 2 NVIDIA H200 GPUs (140GB each) and 12 Intel Xeon Gold 6348 CPUs (2.60GHz, 1TB RAM). Most experiments were conducted on a single GPU, except for evaluations involving mid and large scale open-weight models, where all 2 GPUs were used to accelerate inference.

\section{Illustrative Examples}\label{app:examples}
In this part, we include a small set of examples (Tables~\ref{tab:suddenly-example1-part1},~\ref{tab:suddenly-example1-part2},~\ref{tab:sabotage-example2-part1},~\ref{tab:sabotage-example2-part2},~\ref{tab:scandal-example3-part1},~\ref{tab:scandal-example3-part2}, ~\ref{tab:tightspot-example4-part1},~\ref{tab:tightspot-example4-part2},~\ref{tab:tightspot-example5-part1},~\ref{tab:tightspot-example5-part2}) showing when visual or text cues are needed to resolve the claims and how Gemini 2.5 Pro (best performing model) succeeds or fails on them, along with its corresponding explanations. 

\section*{AI Assistance}
We would like to note that large language models (ChatGPT) were used to assist in drafting and polishing the writing of this work.

\begin{figure*}
    \centering
    \begin{tcolorbox}[title={Guidelines for Data Annotation (Part 1)}, colback=gray!2, colframe=black!30, arc=1mm,
boxrule=0.4pt, width=0.95\textwidth, fonttitle=\bfseries, fontupper=\small]

We are conducting a research study on long movie understanding as part of a broader effort to explore how well viewers comprehend and recall complex narratives. Your task is to create claims that test a viewer's comprehension of a movie after watching it. These claims will be used in a human evaluation study to assess how well participants understand and recall key events from the movie. We appreciate your participation in this data collection process.

\paragraph{General Task Instructions}Select a movie from the current ``Pool'' of movies (the ``Pool'' can be found in \textsc{<link>}). Make sure this movie is not selected by another annotator.
\begin{itemize}
    \item Watch the entire movie carefully. 
    \item We highly recommend reading the example claims provided to gain a better understanding of the task you need to fulfil.
    \item Start writing down your claims following the template available in \textsc{<link>} (you will find two tabs available: the ``Examples'' tab contains claim examples, and the ``Annotations Template'' tab is the template you should follow). Please create another sheet with your claims–do not directly use the current template–and send it to us once it is completed.
\end{itemize}

\paragraph{Annotation Process}

\paragraph{1. Writing Claims}
You are asked to create pairs of contrastive claims, where one claim is true (fact) and the counterfactual version is false (fib). The two claims should differ by minimal edits, meaning they should be as similar as possible while maintaining contrast. Each claim should differ in a subtle but meaningful way, challenging comprehension without being overly obvious.\\
\textbf{Example:}\\
Fact: The first bomb exploded in the bus.\\
Fib: The first bomb exploded in the aquarium.\\
Why this works: The counterfactual claim is created with minimal edits, maintaining contrast while testing the understanding of a key event.

\paragraph{2. Select Claim Granularity}
For each pair of claims you constructed, indicate whether answering them correctly requires reasoning based on a single scene, multiple scenes, or globally within the movie.\\
\textbf{Definition of scene:}\\
A scene in film refers to a complete unit of storytelling, usually consisting of a sequence of events and dialogue taking place in a specific location and time. It often involves one or more characters and is usually shot in one continuous take or consisting of a sequence of shots.\\
\textbf{Reasoning Granularity Labels:}
\begin{itemize}
    \item \textbf{Single-scene:} Claims that are answerable using information from a single scene.
    \item \textbf{Multi-scene:} Claims falling into this granularity require information/evidence from multiple distinct scenes, but not from the whole film. In this case, details are usually spread out between the multiple scenes. The supporting information/evidence is distributed, but explicit and locatable (timestamps/scenes can be clearly identified and referenced)
    \item \textbf{Global:} Claims falling into this granularity require a holistic understanding of the movie narrative. They cannot be easily tied to specific scenes or timestamps, and need to infer or accumulate information/evidence that emerges across the entire narrative (timestamps/scenes can not be clearly identified and referenced).
\end{itemize}

\textit{Note:} Reasoning granularity labels should be selected based on the fact (true claim). Check the examples provided in the ``Examples for Reasoning Granularity'' part.
\end{tcolorbox}
\caption{Guidelines provided for the data annotation procedure (Part 1).}
    \label{fig:guidelines_data_annotation_part1}
\end{figure*}

\begin{figure*}
    \centering
    \begin{tcolorbox}[title={Guidelines for Data Annotation (Part 2)}, colback=gray!2, colframe=black!30, arc=1mm,
boxrule=0.4pt, width=0.95\textwidth, fonttitle=\bfseries, fontupper=\small]
\paragraph{3. Claim Categorization}
Identify the comprehension dimensions the constructed pair of claims examines. Sometimes more than one dimension is examined, so we allow for multiple labels. \\
\textbf{Comprehension Dimension Labels:}
\begin{itemize}
    \item \textbf{Event/Entity Understanding:} it refers to claims that require the identification of key entities (such as people, places, or objects) and understanding of actions or events involving those entities throughout the narrative.  Understanding these claims involves tracking the presence and role of entities across scenes, extracting relationships among them, observing and interpreting their actions, and linking them to relevant events in the narrative.
    \item \textbf{Temporal Perception:} temporal perception refers to claims that require understanding of the timeline of events. It involves reasoning about the order in which events or actions occur---e.g., determining whether an event/action takes place before, after or at the same time as another---and may also require counting the number of specific actions or events. Unlike tasks focused on localizing a specific action in time, temporal perception emphasizes comprehension of broader temporal relationships within the evolving storyline.
    \item \textbf{Emotion Understanding:} emotional understanding refers to claims that involve recognizing and interpreting the emotional development of characters throughout the narrative.
    \item \textbf{Causal Reasoning:} causal reasoning refers to claims that require identifying cause-and-effect relationships between events or actions, where the relationship may be either direct or implicit.
    \item \textbf{Other:} If none of the above fit, select "Other" and suggest a new category.
\end{itemize}

\textit{Note:} The categorization is based on both claims (fact and fib). Check the examples provided in the “Examples for Comprehension Dimensions” part.

\paragraph{Important Points To Consider}
\begin{itemize}
    \item \textbf{Ensure claims assess the viewer’s understanding of the movie}. To put it simply, claims should refer to \textbf{significant moments} in the movie, \textbf{avoiding trivial details or Needle in a Haystack (NIAH)-style claims}, such as: ``The detective wears a red T-shirt'' (if this detail is not important in the movie). 
    \item \textbf{Claims must be clear and unambiguous in isolation}, meaning they should be understandable without requiring additional context but should still require reasoning based on the movie. \textbf{Each claim should be self-contained and make sense independently}, without referencing its counterfactual version. Also, \textbf{avoid highly subjective or interpretive claims}. Each claim should still have a definitive answer based on the movie’s content. 
    \item \textbf{Avoid providing unnecessary contextual details}. For example, do not use phrases like ``in the beginning of the movie, …'', ``in the final scene, …'' unless such information is essential to understanding the claim.
    \item Ensure that claims focus on memorable and salient narrative content that can be recalled without rewatching the movie.
    \item Once you finish the annotation process, please \textbf{go through your claims and confirm that they are in line with the points raised above} (these points are important to be covered to ensure good  quality of annotations).
\end{itemize}
\end{tcolorbox}
\caption{Guidelines provided for the data annotation procedure (Part 2).}
    \label{fig:guidelines_data_annotation_part2}
\end{figure*}

\begin{figure*}
    \centering
    \begin{tcolorbox}[title={Guidelines for Data Annotation (Part 3)}, colback=gray!2, colframe=black!30, arc=1mm,
boxrule=0.4pt, width=0.95\textwidth, fonttitle=\bfseries, fontupper=\small]

\paragraph{Examples for Reasoning Granularity} In this part, we provide examples to illustrate how to assign reasoning granularity labels.\\

\textbf{Example 1:}\\
\textit{Fact:} According to the Hattley, the individual shown in the photograph (Marakelli) worked with Constain.\\
\textit{Fib:} According to Hattley, the individual shown in the photograph (Marakelli) had no connection or working relationship with Constain.\\
\textit{Reasoning Granularity:} Single-scene.\\
\textit{Justification:} This event is categorized as single-scene because it takes place within one specific scene: Hattley shows the photograph to Conley, they are having a discussion and it is implied that Marakelli worked with Constain in the mafia.\\

\textbf{Example 2:}\\
\textit{Fact:} Hattley appeared visibly bothered with the discussion he had in his office with Constain’s attorney.\\
\textit{Fib:} Hattley appeared pleased with the discussion he had in his office with Constain’s attorney.\\
\textit{Reasoning Granularity:} Single-scene.\\
\textit{Justification:}  That is again a single scene event. Constain’s attorney enters the office and they are having a discussion. After a while, Hattley kicks him out.\\

\textbf{Example 3:}\\
\textit{Fact:} Miss Conley received a dress as a personal gift from the policeman.\\
\textit{Fib:} Miss Conley received a dress as a gift from the government, delivered by the policeman.\\
\textit{Reasoning Granularity:} Multi-scene.\\
\textit{Justification:} That is a multi-scene event, that we need to ground on 2 independent scenes to answer the question correctly. In the first scene Miss Conley receives a gift from the policeman, who says that the gift is from the government. After a while (some scenes are interleaved), she understands that the policeman bought the gift for her and not the government. So to answer correctly, we need to ground on these 2 specific scenes.\\

\textbf{Example 4:}\\
\textit{Fact:} Conley’s statement about her occupation, describing herself as a “gang buster,” implicitly refers to Constain.\\
\textit{Fib:} Conley’s statement about her occupation, describing herself as a “gang buster,” implicitly refers to Pete Tinelli.\\
\textit{Reasoning Granularity:} Global\\
\textit{Justification:} There is a single scene in the end of a movie during which Conley characterises herself as a “gang buster”. Although it is a single scene, it is impossible to understand solely by this scene why she said it and to whom she is referring to. We need to watch a big part of the movie (if not all of it) to understand that refers to Constain. 

\end{tcolorbox}
\caption{Guidelines provided for the data annotation procedure (Part 3). This part of the guidelines provides examples given to annotators to illustrate how to assign reasoning granularity labels. While more examples were shared during the annotation process, we include a selection here for illustrative purposes.}
    \label{fig:guidelines_data_annotation_examples_reasoning}
\end{figure*}

\begin{figure*}
    \centering
    \begin{tcolorbox}[title={Guidelines for Data Annotation (Part 4)}, colback=gray!2, colframe=black!30, arc=1mm,
boxrule=0.4pt, width=0.95\textwidth, fonttitle=\bfseries, fontupper=\small]

\paragraph{Examples for Comprehension Dimensions} In this part we provide examples to illustrate how to assign comprehension dimension labels.\\

\textbf{Example 1:}\\
\textit{Fact:} At Jim's bar, the Connel keeps drinking as he talks to the fake John Doe, expressing his frustration and concern.\\
\textit{Fib:} At Jim's bar, the Connel keeps drinking as he talks to the fake John Doe, expressing hope and happiness.\\
\textit{Comprehension Dimension:} emotion understanding\\
\textit{Justification:} We need to understand what emotion Connel expressed, to answer the pair of claims correctly.\\

\textbf{Example 2:}\\
\textit{Fact:} Conley’s statement about her occupation describing herself as a “gang buster”, implicitly refers to Constain.\\
\textit{Fib:} Conley’s statement about her occupation describing herself as a “gang buster”, implicitly refers to Pete Tinelli.\\
\textit{Comprehension Dimension:} entity/event understanding\\
\textit{Justification:} We need to understand to whom the expression “gang buster” refers to. So, the comprehension dimension is entity understanding. \\

\textbf{Example 3:}\\
\textit{Fact:} Hallet brought Conley’s sister to the hotel with the intent to make Conley testify in the trial.\\
\textit{Fib:} Hallet brought Conley’s sister to the hotel with the intent to make her feel safe.\\
\textit{Comprehension Dimension:} causal reasoning\\
\textit{Justification:} Here we need to understand why Hallet brought Conley’s sister to the hotel. So it examines a causal-and-effect relationship.\\

\textbf{Example 4:}\\
\textit{Fact:} Conley decided to testify only after Wiloughby’s death.\\
\textit{Fib:} Conley had already decided to testify before Wiloughby’s death.\\
\textit{Comprehension Dimension:} temporal perception\\
\textit{Justification:} that pair examines the temporal dimension (if the decision was taken before or after Wiloughby’s death).
\end{tcolorbox}
\caption{Guidelines provided for the data annotation procedure (Part 4). This part of the guidelines provides examples given to annotators to illustrate how to assign comprehension dimension labels. While more examples were shared during the annotation process, we include a selection here for illustrative purposes.}
\label{fig:guidelines_data_annotation_examples_comprehension_dims}
\end{figure*}

\begin{figure*}
    \centering
    \begin{tcolorbox}[title={Guidelines for Human Evaluation (Part 1)}, colback=gray!2, colframe=black!30, arc=1mm,
boxrule=0.4pt, width=0.95\textwidth, fonttitle=\bfseries, fontupper=\small]
This evaluation study aims to assess how well people comprehend and recall key events from a movie. You will watch a movie and then evaluate a series of claims about its content. Your goal is to determine whether each claim is True or False, based solely on what was shown in the movie.  We appreciate your participation in this study.
\paragraph{Task Instructions}
\begin{itemize}
    \item Assign to yourself the movies you want to watch and do the test (we expect 2 movies per person). Please add your name to the Human-Eval column, on this \textsc{link}. 
    \item Visit the platform for evaluation \textsc{link}. 
    \item Provide your email to receive access to the movie (it will be used as your unique identifier).
    \item Once you submit your email, you should carefully select from the drop-down list the corresponding movie you assigned yourself  and proceed with the evaluation. You will be shown with the movie link. Please open it in a new tab.
\end{itemize}

The test is divided in \textbf{2 stages}:
The \textbf{first stage} is \textbf{mandatory} and should be completed by everyone (\textit{during this stage you are not allowed to go back to the movie while answering the questions}). The \textbf{second stage} is \textbf{optional} (\textit{during this stage you are allowed to go back to the movie while answering the questions}).
\paragraph{Stage 1:}
\begin{enumerate}
    \item \textbf{Watch the entire movie carefully before proceeding to the evaluation}. Pay attention to details and context in the movie, as some claims may be subtle or require careful reasoning. 
\item After watching, it’s time to proceed to Stage 1. \textbf{Please do not go back to the movie until Stage 1 of the test is completed.} Press the ``Start Classifying Claims'' button, and you will be shown with \textbf{one claim at a time}. For each claim shown, you need to do the following:
\begin{itemize}
    \item \textbf{Classify the claim as True/False} (you should always answer truthfully, without aiming to maximise you score).
    \item Mark your \textbf{confidence} about your answer. This is helpful for stage 2, where you will have the opportunity to revise your claims (by looking back at the movie).
    \item Leave a comment if any of the following applies: If a claim is \textbf{ambiguous, unclear, open to interpretation, has a bad phrasing or typos,  you may leave an optional comment explaining your concerns}. You can also comment on the claim in case it is \textbf{needle-in-a-haystack style} and you think it is too detailed and doesn't test the understanding of the movie.
    \item Once you answered, click "Save" to submit your response and move on to the next claim.
\end{itemize} 
\end{enumerate}
\paragraph{Important details:}
Once you submit an answer, you cannot go back and change it.
At this stage, you are \textbf{strictly prohibited from searching back in the movie, rewinding, or rewatching scenes while answering the claims}. Your responses \textbf{should be based on your memory} and \textbf{understanding}.
You must \textbf{NOT use any AI tools or external sources to verify or generate answers}. The goal of this study is to assess human understanding of long movies, not automated retrieval or AI-assisted responses. Also you are not allowed to take any paper notes, while watching the movie.
\end{tcolorbox}
\caption{Guidelines provided for human evaluation (Part 1).}
    \label{fig:guidelines_human_eval_part1}
\end{figure*}

\begin{figure*}[t]
\centering
\begin{tcolorbox}[colback=gray!2, colframe=black!30, arc=1mm, boxrule=0.4pt, width=0.95\textwidth,
  title={\small Direct Prompt Template}, fontupper=\footnotesize, fonttitle=\small, before skip=5pt, after skip=5pt, left=3pt, right=3pt, top=3pt, bottom=3pt]

\textbf{System:} You are a helpful AI assistant. Your task is to carefully analyze the provided content and determine whether statements made about it are true or false based on the available information.

\smallskip
\textbf{User:} You are provided with a movie and a statement. 
    Your task is to carefully watch the movie and then determine whether the statement is true or false.\\
    Answer TRUE if the statement is true in its entirety based on the movie.\\
    Answer FALSE if any part of the statement is false based on the movie. \\
    \\
    \textbf{Statement: \{claim\}}\\
    Based on the movie, is the above statement TRUE or FALSE?\\
    Provide only your final answer.\\  
\end{tcolorbox}
\caption{Direct prompt template used for open-weight models.}\label{fig:direct_prompt_template}
\end{figure*}

\begin{figure*}[t]
\centering
\begin{tcolorbox}[colback=gray!2, colframe=black!30, arc=1mm, boxrule=0.4pt, width=0.95\textwidth,
  title={\small Explanation Prompt Template}, fontupper=\footnotesize, fonttitle=\small, before skip=5pt, after skip=5pt, left=3pt, right=3pt, top=3pt, bottom=3pt]

\textbf{System:} You are a helpful AI assistant. Your task is to carefully analyze the provided content and determine whether statements made about it are true or false based on the available information.

\smallskip
\textbf{User:} You are provided with a movie and a statement. 
    Your task is to carefully watch the movie and then determine whether the statement is true or false.\\
    Answer TRUE if the statement is true in its entirety based on the movie.\\
    Answer FALSE if any part of the statement is false based on the movie. \\
    \\
    \textbf{Statement: \{claim\}}\\
    Based on the movie, is the above statement TRUE or FALSE?\\
    First provide an explanation of your decision-making process in at most one paragraph, and then provide your final answer.\\  
\end{tcolorbox}
\caption{Explanation prompt template used for closed models.}\label{fig:explanation_prompt_template}
\end{figure*}
\clearpage

{
\renewcommand{\arraystretch}{1.1}
\rowcolors{1}{white}{lightgrayrow}

\begin{table*}[t]
\centering
\small
\begin{tabularx}{\textwidth}{
    p{4.9cm}
    X
    >{\centering\arraybackslash}p{0.8cm}
    >{\centering\arraybackslash}p{1.1cm}
}
\toprule
\textbf{Movie (Year)} &
\textbf{Genre (IMDB)} &
\textbf{Lang.} &
\textbf{Duration} \\
\midrule

The Last Chance (1945) & Drama, War & en, it & 93.84 \\
They Made Me a Criminal (1939) & Boxing, Film Noir, Crime, Drama, Sport & en & 91.21 \\
Tokyo After Dark (1959) & Drama & en & 81.23 \\
The Sadist (1963) & Horror, Thriller & en & 91.63 \\
Suddenly (1954) & Film Noir, Psychological Thriller, Crime, Drama, Thriller & en & 76.71 \\
Sabotage (Hitchcock) (1936) & Psychological Thriller, Spy, Crime, Thriller & en & 75.92 \\
Murder By Contract (1958) & Film Noir, Crime, Drama, Thriller & en & 80.45 \\
Pushover (1954) & Film Noir, Crime, Drama, Thriller & en & 87.77 \\
Go for Broke (1951) & Drama, History, War & en & 90.85 \\
Meet John Doe (1941) & Political Drama, Satire, Comedy, Drama, Romance & en & 122.87 \\
Scarlet Street (1945) & Film Noir, Tragedy, Crime, Drama, Thriller & en & 102.39 \\
Little Lord Fauntleroy (1936) & Period Drama, Drama, Family & en & 100.72 \\
Deadline - U.S.A. (1952) & Film Noir, Crime, Drama & en & 87.06 \\
My Favorite Brunette (1947) & Hard-boiled Detective, Comedy, Crime, Mystery, Romance, Thriller & en & 87.34 \\
Woman in the Moon (1929) & Adventure, Comedy, Drama, Romance, Sci-Fi & de & 168.73 \\
Lonely Wives (1931) & Comedy, Romance & en & 85.35 \\
Nothing Sacred (1937) & Satire, Screwball Comedy, Comedy, Drama, Fantasy, Romance & en & 73.57 \\
Fingerman (1955) & Film Noir, Crime, Drama, Thriller & en & 82.06 \\
Borderline (1950) & Film Noir, Crime, Drama, Thriller & en & 88.16 \\
Babes in Toyland (1934) & Screwball Comedy, Slapstick, Comedy, Family, Fantasy, Musical & en & 77.26 \\
The Man From Utah (1934) & Drama, Western & en & 51.49 \\
The Man With The Golden Arm (1955) & Drug Crime, Psychological Drama, Crime, Drama, Romance & en & 119.07 \\
A Star Is Born (1937) & Tragic Romance, Drama, Romance & en & 110.98 \\
Africa Screams (1949) & Farce, Action, Adventure, Comedy & en & 79.13 \\
Dementia 13 (1963) & Slasher Horror, Horror, Thriller & en & 74.94 \\
Fear and Desire (1952) & Drama, Thriller, War & en & 70.19 \\
The Little Princess (1939) & Costume Drama, Comedy, Drama, Family, Musical & en & 92.77 \\
Father's Little Dividend (1951) & Comedy, Drama, Romance & en & 81.74 \\
Kansas City Confidential (1952) & Conspiracy Thriller, Film Noir, Heist, Crime, Drama, Thriller & en & 99.27 \\
Of Human Bondage (1934) & Dark Romance, Film Noir, Medical Drama, Tragedy, Tragic Romance, Drama, Romance & en & 82.77 \\
Half Shot at Sunrise (1930) & Comedy, Musical & en, fr & 78.04 \\
Bowery at Midnight (1942) & B-Horror, Crime, Horror, Thriller & en & 62.05 \\

\bottomrule
\end{tabularx}

\caption{Metadata of collected movies (Part 1).}
\label{tab:movie_details_1}
\end{table*}
}

{
\renewcommand{\arraystretch}{1.1}
\rowcolors{1}{white}{lightgrayrow}

\begin{table*}[t!]
\centering
\small
\begin{tabularx}{\textwidth}{
    p{4.9cm}
    X
    >{\centering\arraybackslash}p{0.8cm}
    >{\centering\arraybackslash}p{1.1cm}
}
\toprule
\textbf{Movie (Year)} &
\textbf{Genre (IMDB)} &
\textbf{Lang.} &
\textbf{Duration} \\
\midrule

The Emperor Jones (1933) & Drama, Music & en & 76.29 \\
The Deadly Companions (1961) & Adventure, Drama, Western & en & 93.62 \\
The Red House (1947) & Film Noir, Drama, Mystery, Thriller & en & 100.39 \\
Trapped (1949) & Film Noir, Crime, Drama, Thriller & en & 79.4 \\
City of Fear (1959) & Crime, Drama, Thriller & en & 75.18 \\
Kid Monk Baroni (1952) & Action, Drama, Sport & en & 79.56 \\
Tight Spot (1955) & Film Noir, Crime, Drama, Thriller & en & 95.99 \\
Captain Kidd (1945) & Costume Drama, Swashbuckler, Adventure, Biography, Drama, History & en & 87.53 \\
The Front Page (1931) & Dark Comedy, Satire, Screwball Comedy, Comedy, Crime, Drama, Mystery, Romance & en & 101.14 \\
The Hitch-Hiker (1953) & Film Noir, Crime, Drama, Thriller & en & 70.8 \\
Obsession (1949) & Film Noir, Psychological Thriller, Crime, Thriller & en & 92.39 \\
Thunderbolt (1929) & Film Noir, Crime, Drama, Music, Romance & en & 91.27 \\
Cyrano de Bergerac (1950) & Swashbuckler, Adventure, Drama, Romance & en & 112.87 \\
Scandal Sheet (1952) & Film Noir, Crime, Drama, Romance, Thriller & en & 81.75 \\
Ladies in Retirement (1941) & Film Noir, Crime, Drama & en & 92.31 \\
Detour (1945) & Film Noir, Crime, Drama & en & 69.09 \\
The Crooked Way (1949) & Film Noir, Crime, Drama, Thriller & en & 85.95 \\
A Bucket of Blood (1959) & Comedy, Crime, Horror & en & 65.84 \\
Love Affair (1939) & Holiday Romance, Comedy, Drama, Romance & en & 89.62 \\
The Jackie Robinson Story (1950) & Biography, Drama, Sport & en & 76.82 \\
The Last Time I Saw Paris (1954) & Tragedy, Tragic Romance, Drama, Romance & en & 116.02 \\
\bottomrule
\end{tabularx}

\caption{Metadata of collected movies (Part 2).}
\label{tab:movie_details_2}
\end{table*}
}
\clearpage
\definecolor{darkgreen}{rgb}{0,0.45,0}
\definecolor{darkred}{rgb}{0.7,0,0}
\definecolor{lightred}{rgb}{1,0.7,0.7}

\begin{table*}[t]
\centering
\begin{tabular}{p{0.94\linewidth}}
\toprule
\textbf{Example 1 (Part 1) - Movie: \textit{"Suddenly"}} \\
\midrule

\begin{minipage}{0.23\linewidth}
    \centering
    \includegraphics[width=\linewidth]{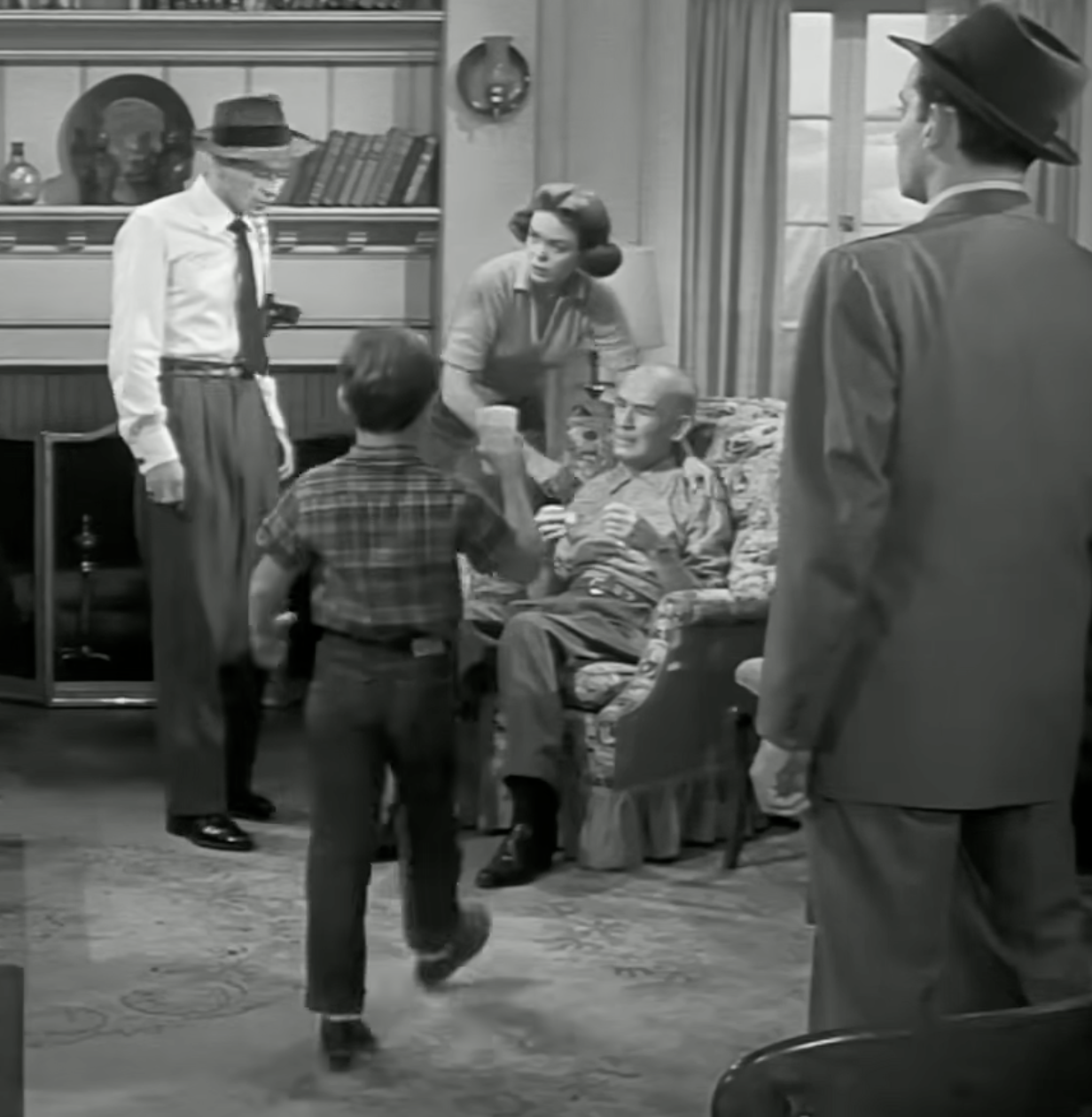}
    {\footnotesize Pidge brings the water}
\end{minipage}\hfill
\begin{minipage}{0.23\linewidth}
    \centering
    \includegraphics[width=\linewidth]{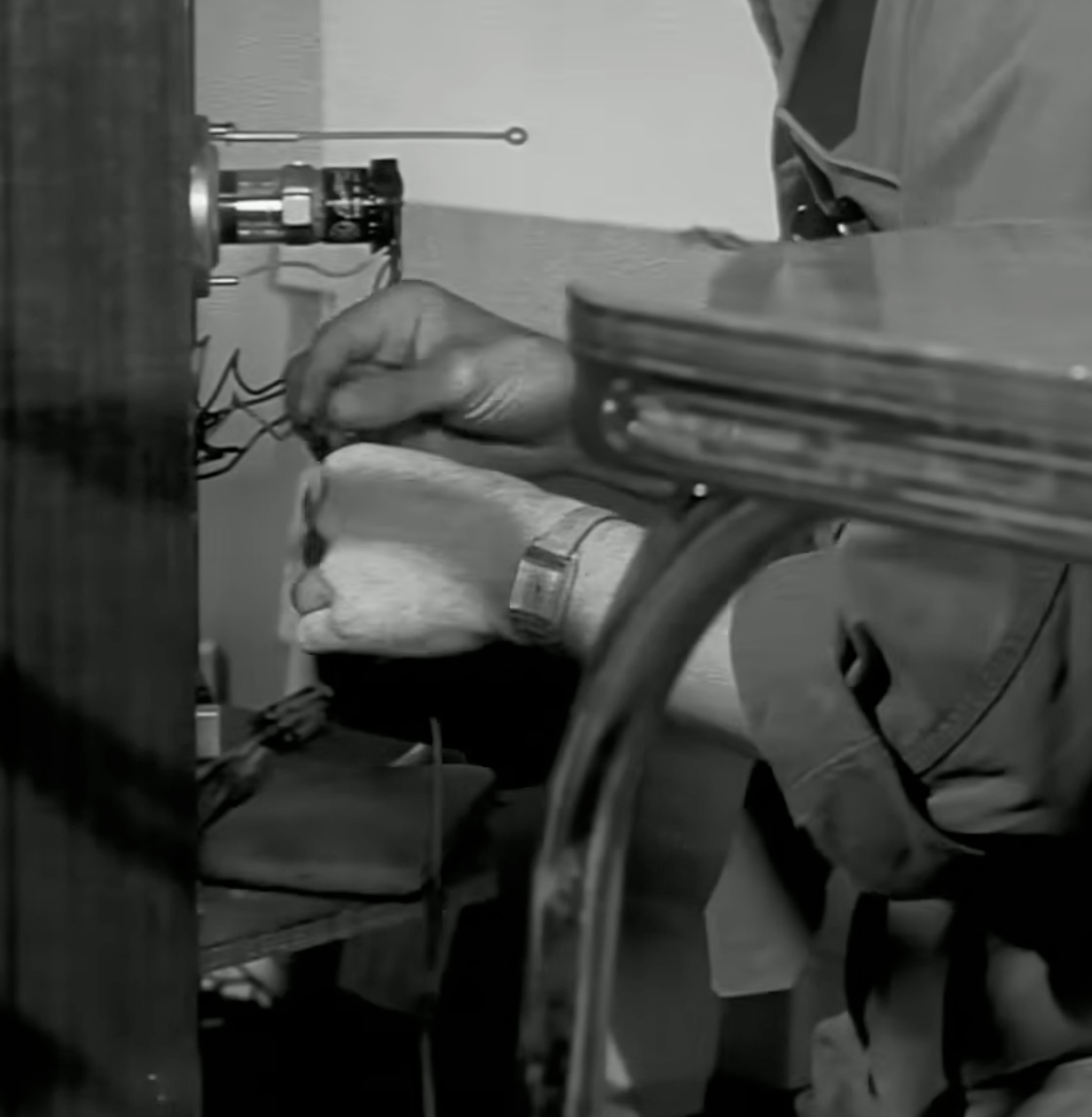}
    {\footnotesize Setting up the circuit}
\end{minipage}\hfill
\begin{minipage}{0.23\linewidth}
    \centering
    \includegraphics[width=\linewidth]{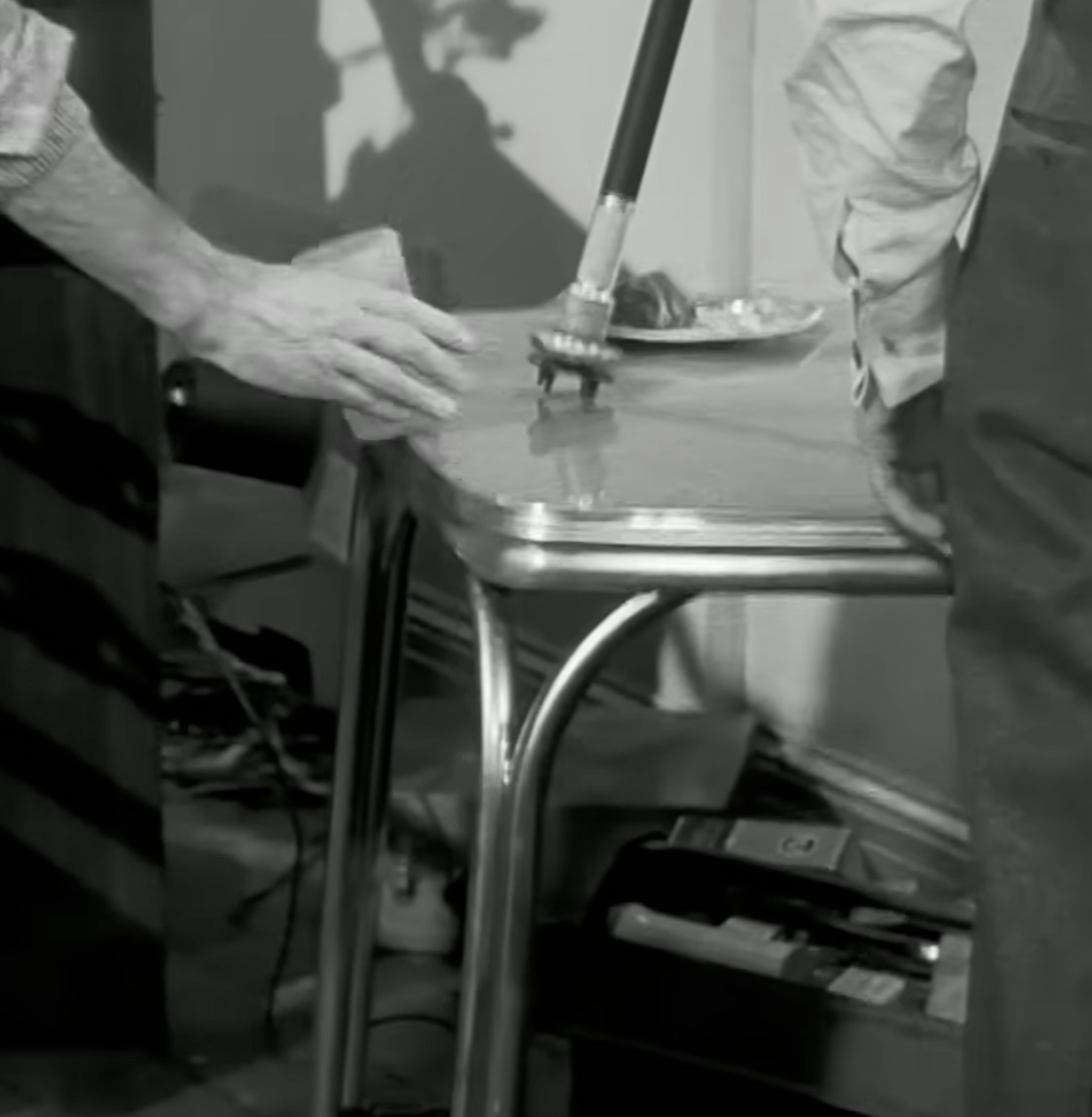}
    {\footnotesize Pop spills the water}
\end{minipage}\hfill
\begin{minipage}{0.23\linewidth}
    \centering
    \includegraphics[width=\linewidth]{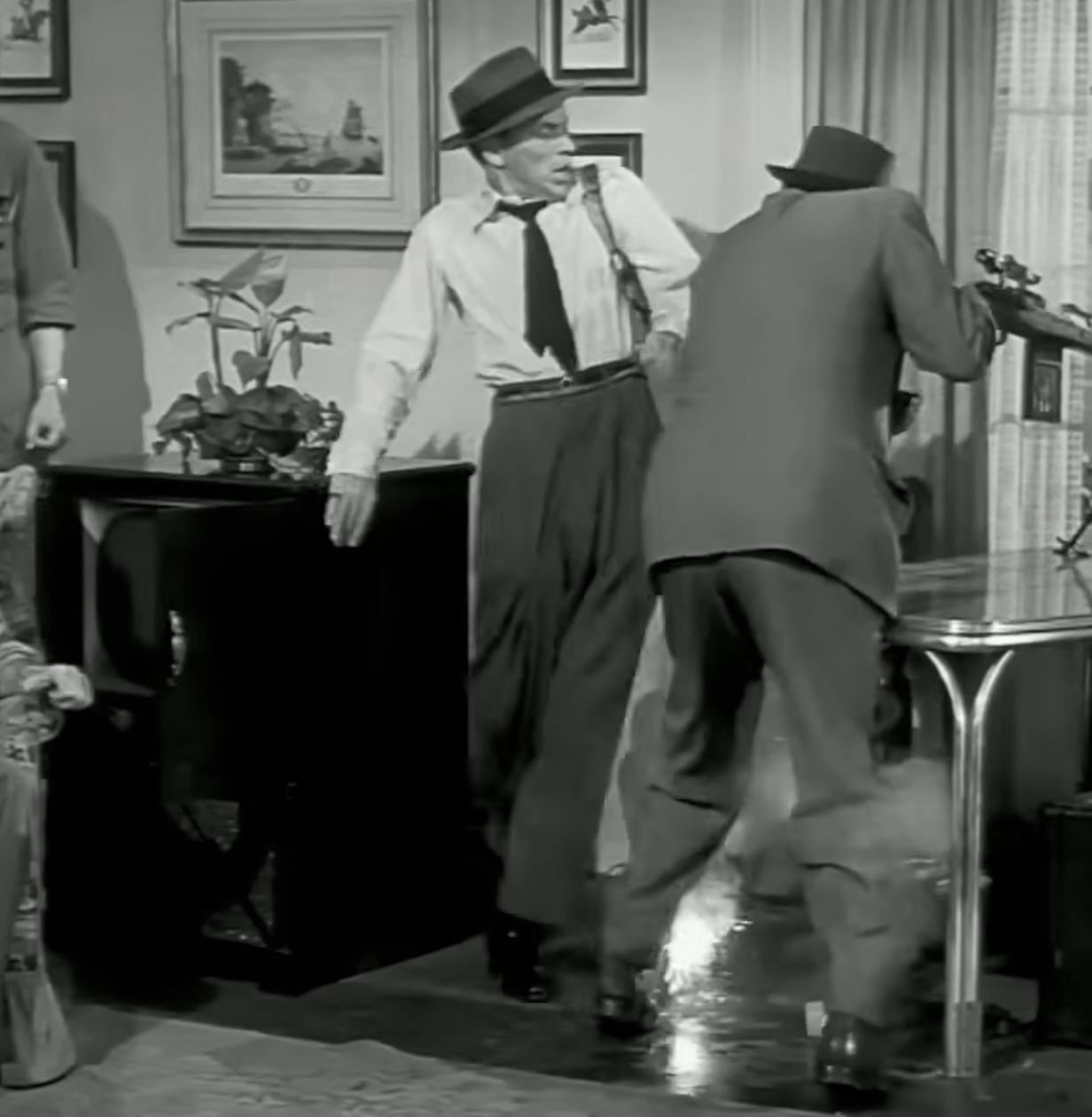}
    {\footnotesize The electrocution}
\end{minipage}

\\\\

\textbf{Fact} (\cmark): The grandfather, Peter Benson, intentionally spilled the water on the floor. \\

\textbf{Fib} (\xmark): The grandfather, Peter Benson, accidentally spilled the water on the floor. \\\\

\textbf{Granularity}: Multi-scene/Global \\\\

\textbf{Context (\textcolor{darkred}{Spoiler alert !})}: Early in the film, the TV breaks. Later, when the criminals have already broken into the house, pop (the grandfather) orchestrates the setup to save everyone (without explicitly stating it). He feigns a heart attack and asks for water, Pidge brings it. The TV technician pretends to repair the broken TV as part of a sabotage plan, while the circuit. When the moment comes, Pop \textit{intentionally} spills the water on the floor (placing it on the edge of the table). Shortly after, the electrocution occurs when one of the criminals steps on the spilled water. \textit{Both visual and textual information are needed to understand that Pop indeed spilled the water and what was his intention. Gemini fails to predict both claims correctly in every modality setting.}\\\\

\textbf{Gemini's Predictions:} \\\\
\colorbox{lightred}{\textit{Subs only setting}} \\\\
\textbf{Prediction on Fact}:  
The statement is ``The grandfather, Peter Benson, intentionally spilled the water on the floor.'' To determine if this is true, we need explicit evidence in the transcript: 1. That Peter Benson spilled water; 2. That he did so intentionally. Transcript evidence: Peter Benson (Pop) feigns a heart attack. He asks Pidge: ``Pidge, put down that gun and get some water for your grandfather.'' Pidge presumably brings the water. Later, just before the climax, the line ``Oh, butterfingers!'' appears. This implies something was dropped or spilled, likely the water. However, the transcript does not explicitly state who exclaimed ``Oh, butterfingers!'' or who spilled the water. Since the transcript does not explicitly confirm that Peter Benson spilled the water, the statement cannot be verified. \textbf{FALSE} \textcolor{darkred}{\textit{(Ground-truth: TRUE)}}\\\\

\textbf{Prediction on Fib}:  
The transcript mentions Pidge being asked to get water for his grandfather, Peter Benson (Pop), when Pop feigns a heart attack [..] However, the transcript does not state that Peter Benson subsequently spilled this water, or any water, on the floor.  [...] it is not explicitly stated who Jud is referring to or that the fumbling involved spilling water. The context is the electrocution and shooting, not an earlier incident of spilling water. There is no direct statement or clear implication in the transcript that Peter Benson accidentally spilled water on the floor. \textbf{FALSE} \textcolor{darkred}{\textit{(Ground-truth: FALSE)}}\\\\

\bottomrule
\end{tabular}
\caption{\textbf{Illustrative Example 1 from the movie ``Suddenly''}. Gemini fails to predict both claims correctly in every modality setting.}
\label{tab:suddenly-example1-part1}
\end{table*}

\begin{table*}[t]
\centering
\begin{tabular}{p{0.94\linewidth}}
\toprule
\textbf{Example 1 (Part 2) - Movie: \textit{"Suddenly"}} \\
\midrule
\\
\colorbox{lightred}{\textit{Video only setting}}\\\\
\textbf{Prediction on Fact}: The grandfather, Peter Benson, is seen at 2:17 reaching for a glass of water. He then appears to deliberately drop the glass, causing the water to spill on the floor. This creates a distraction, allowing the woman to crawl towards a gun. The context of the scene, where the family is being held hostage and looking for an opportunity to fight back, combined with the grandfather's actions and the subsequent events, strongly suggests the spill was intentional. \textbf{TRUE} \textcolor{darkred}{\textit{(Ground-truth: TRUE)}}\\\\
\textbf{Prediction on Fib}: The video introduces an older character who is later identified as Peter Benson (around 0:46-0:47). Much later in the video, during a tense hostage situation, Peter Benson is seen sitting in an armchair holding a glass of water. At timestamp 2:18, he is startled and gets up abruptly, causing the water in his glass to spill onto the floor. The context of the scene, where he reacts to the unfolding events, suggests the spill was accidental due to his sudden movement or reaction. \textbf{TRUE} \textcolor{darkred}{\textit{(Ground-truth: FALSE)}}\\\\
\colorbox{lightred}{\textit{Video w/ subs setting}}\\\\
\textbf{Prediction on Fact}: The grandfather, Peter Benson, is seen getting up with a glass of water around timestamp 55:20. As he walks, he stumbles and drops the glass, spilling the water on the floor near the table leg and electrical wiring. Immediately after, he clutches his chest and feigns a heart attack. The transcript reflects this with dialogue about his pain. There is no visual or textual evidence to suggest the spill was intentional; rather, it appears accidental, possibly due to his feigned or real physical distress, and he then uses the moment to further distract the captors by exaggerating his heart condition.\textbf{TRUE} \textcolor{darkred}{\textit{(Ground-truth: TRUE)}}\\\\
\textbf{Prediction on Fib}:
The grandfather, Peter Benson, takes a glass of water and walks towards the table where the assassin has set up the rifle. He places the glass on the edge of the metal table (around 55:23). As he moves to talk to Jud, who is working on the TV, the glass tips over and spills the water on the floor (around 55:24). The spilling appears to be an unintentional consequence of his actions and the precarious placement of the glass. Therefore, it was an accident.\textbf{TRUE} \textcolor{darkred}{\textit{(Ground-truth: FALSE)}}\\

\bottomrule
\end{tabular}
\caption{\textbf{Illustrative Example 1 from the movie ``Suddenly''}. Gemini fails to predict both claims correctly in every modality setting.}
\label{tab:suddenly-example1-part2}
\end{table*}

\definecolor{darkgreen}{rgb}{0,0.45,0}
\definecolor{darkred}{rgb}{0.7,0,0}
\definecolor{lightred}{rgb}{1,0.7,0.7}

\begin{table*}[t]
\centering
\begin{tabular}{p{0.94\linewidth}}
\toprule
\textbf{Example 2 (Part 1) - Movie: \textit{"Sabotage (Hitchcock)"}} \\
\midrule

\begin{minipage}{0.23\linewidth}
    \centering
    \includegraphics[width=\linewidth]{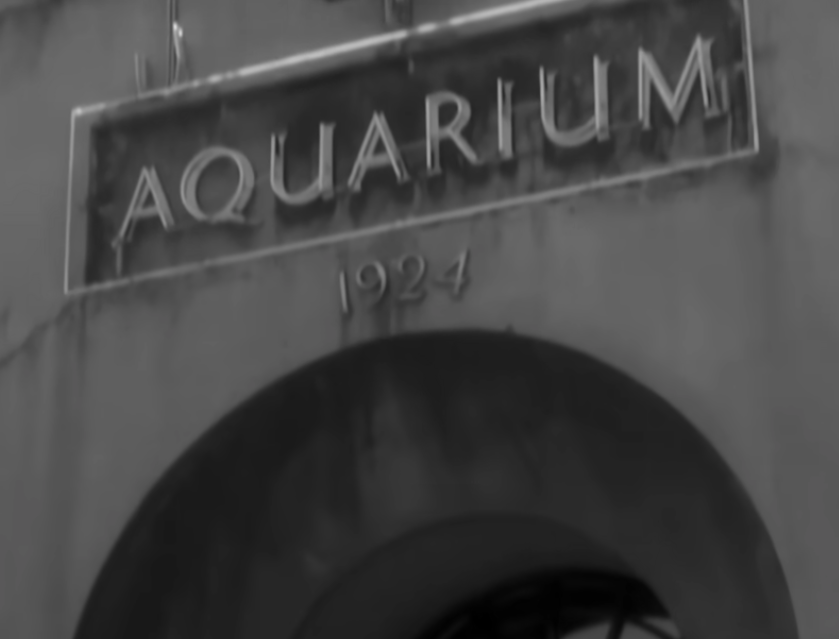}
    {\footnotesize Aquarium}
\end{minipage}\hfill
\begin{minipage}{0.23\linewidth}
    \centering
    \includegraphics[width=\linewidth]{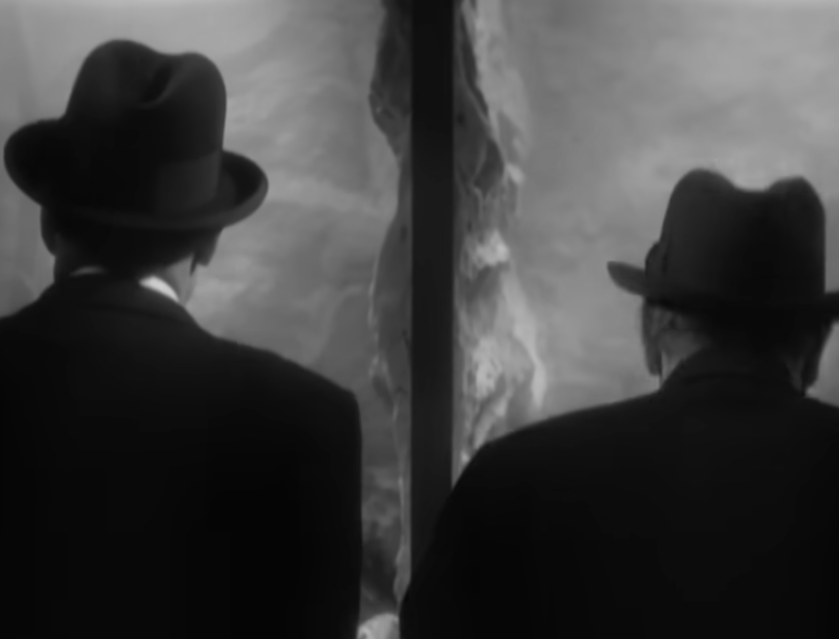}
    {\footnotesize Planning the sabotage}
\end{minipage}\hfill
\begin{minipage}{0.23\linewidth}
    \centering
    \includegraphics[width=\linewidth]{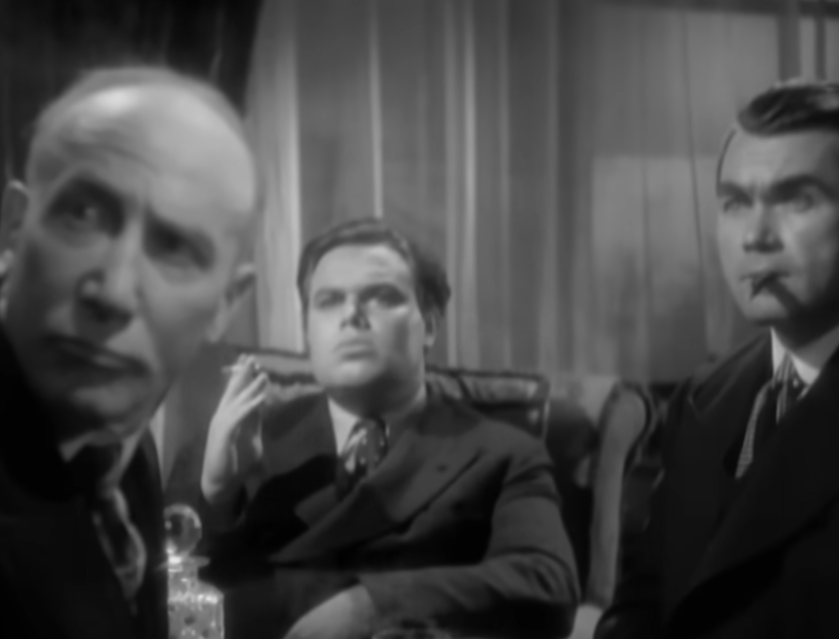}
    {\footnotesize Meet in cinema}
\end{minipage}\hfill
\begin{minipage}{0.23\linewidth}
    \centering
    \includegraphics[width=\linewidth]{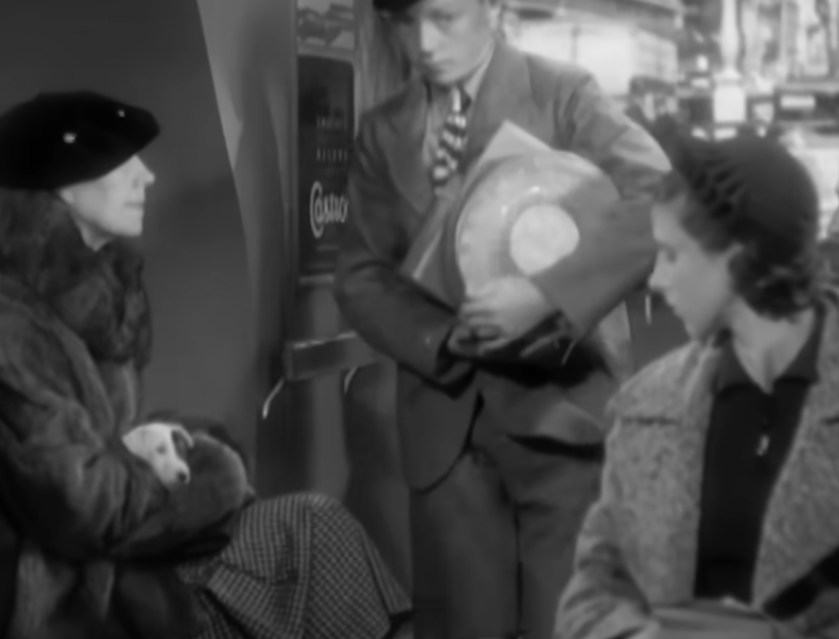}
    {\footnotesize Kid enters the bus}
\end{minipage}\hfill
\begin{minipage}{0.23\linewidth}
    \centering
    \includegraphics[width=\linewidth]{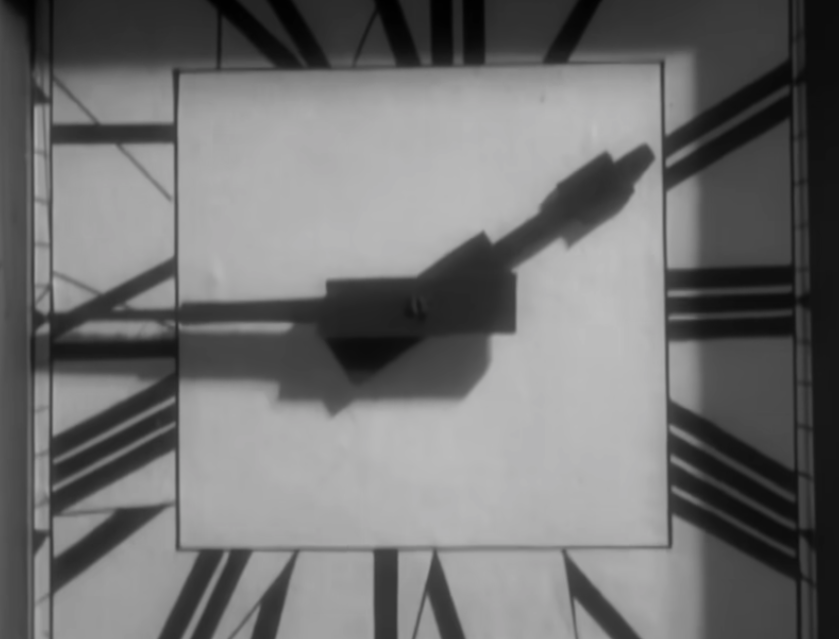}
    {\footnotesize The time arrives}
\end{minipage}\hfill
\begin{minipage}{0.23\linewidth}
    \centering
    \includegraphics[width=\linewidth]{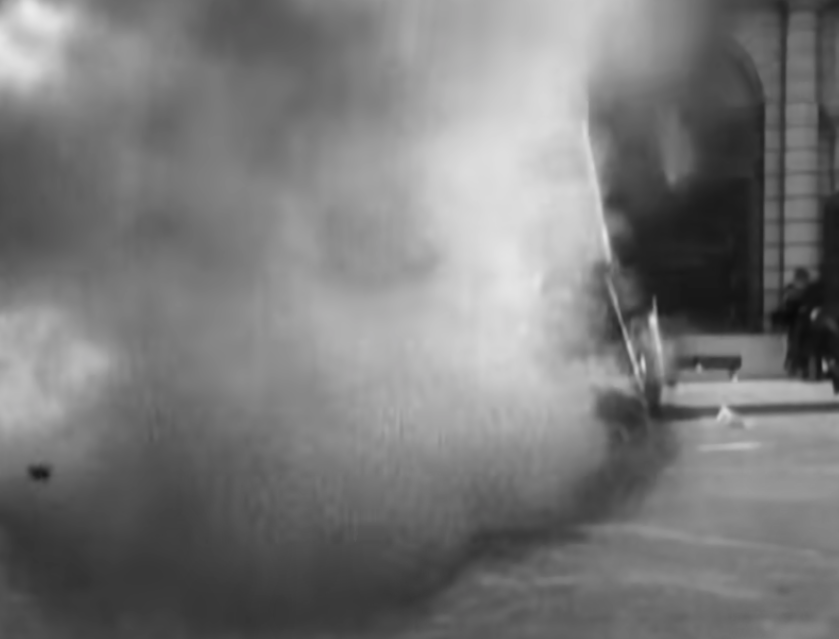}
    {\footnotesize Bus explosion}
\end{minipage}\hfill
\begin{minipage}{0.23\linewidth}
    \centering
    \includegraphics[width=\linewidth]{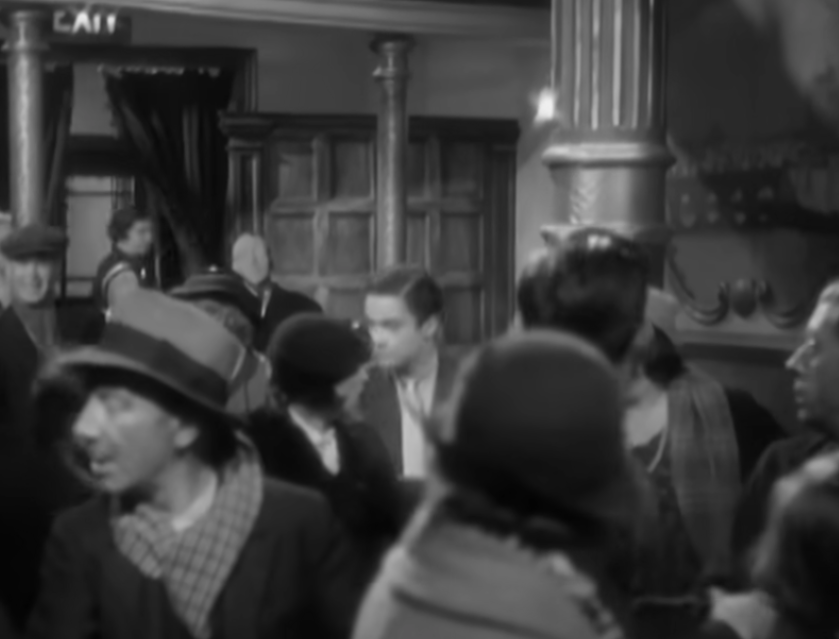}
    {\footnotesize People in the cinema}
\end{minipage}\hfill
\begin{minipage}{0.23\linewidth}
    \centering
    \includegraphics[width=\linewidth]{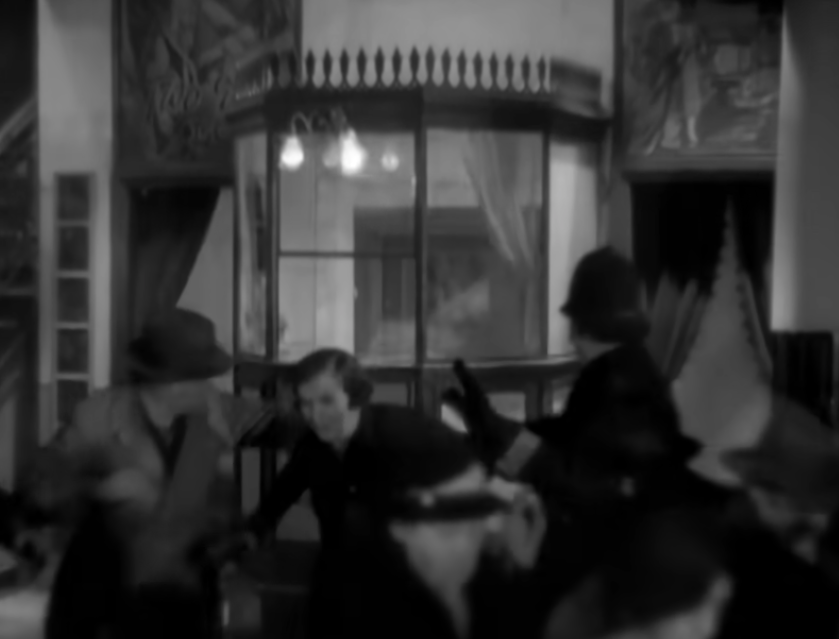}
    {\footnotesize Explosion in cinema}
\end{minipage}

\noindent \\
\textbf{Fact} (\cmark): The second bomb exploded in the cinema.\\

\textbf{Fib} (\xmark): There was a second bomb about to explode, but it didn't.\\\\

\textbf{Granularity}: Global
\\
\textbf{Context (\textcolor{darkred}{Spoiler alert !})}:  Mr. Verloc and the terrorist meet at the aquarium, where the terrorist first mentions the plan to explode a bomb (he mentions ``a parcel of fireworks'' and asks Mr. Verloc to meet his friend the bombmaker). The terrorists meet again in the cinema, to discuss about the sabotage. Later in the movie the explosion happens inside the bus. While the second explosion happens inside the cinema. \textit{Gemini fails to predict both claims correctly in every modality setting.}\\\\

\textbf{Gemini's Predictions:} \\\\
\colorbox{lightred}{\textit{Subs only setting}}\\\\
\textbf{Prediction on Fact}: To determine if the statement is true or false, we need to identify the sequence of bomb-related incidents in the transcript. 1.  **First Incident (Sabotage at the Cinema):** The transcript begins with a power outage at the cinema, described as ``Sand. Sabotage. Wrecking. Deliberate.'' This is the first act of sabotage or a device causing disruption that Verloc is connected with. [...] This incident occurred at the cinema. 2.  **Second Incident (Stevie's Bomb on the Bus):** Mr. Verloc is then tasked with a more serious act: to leave a ``small parcel... a parcel of fireworks'' (a bomb) in Piccadilly Circus. His young brother-in-law, Stevie, is sent to deliver this parcel. This bomb explodes prematurely. [...] This is the second distinct sabotage/bomb event in the plot, and this bomb exploded on a bus, not in the cinema. 3.  **Third Incident (Bomb Explosion at the Cinema):** Towards the end of the transcript, the old man who makes the bombs is cornered in the cinema. [...] This bomb then explodes: [...] This explosion occurs at the cinema. Based on this chronological sequence of events: *   The first destructive ``bomb-like'' event was the sabotage at the cinema (power outage). * The second bomb (Stevie's parcel) exploded on a bus in the West End. * The third bomb (detonated by the old man) exploded in the cinema. [...] Therefore, the statement is false.  \textbf{FALSE} \textcolor{darkred}{\textit{(Ground-truth: TRUE)}}\\
\bottomrule
\end{tabular}
\caption{\textbf{Illustrative Example 2 from ``Sabotage (Hitchcock)''}. Gemini predicts the pair of claims incorrectly in every modality setting.}
\label{tab:sabotage-example2-part1}
\end{table*}

\begin{table*}[t]
\centering
\begin{tabular}{p{0.94\linewidth}}
\toprule
\textbf{Example 2 (Part 2) - Movie: \textit{"Sabotage"}} \\
\midrule
\\
\textbf{Prediction on Fib}: The transcript describes two main bomb-related incidents. The first is the explosion that kills Stevie. The second occurs at the cinema where the old man (the bomb maker) threatens to ``blow the place sky high.'' The transcript later confirms an explosion there: ``The roof's gone. The whole of the back of the place - clean blown off.'' This indicates the second bomb *did* explode. Therefore, the statement that a second bomb was ``about to explode, but it didn't'' is false because it did explode. \textbf{FALSE} \textcolor{darkred}{\textit{(Ground-truth: FALSE)}}\\\\
\colorbox{lightred}{\textit{Video-only setting}}\\\\
\textbf{Prediction on Fact}: The video shows one clear explosion involving a bus on a street around the 1:46 mark. Following this, there are various scenes, including one where people are watching a cartoon in a cinema. However, there is no depiction or indication of a second bomb exploding, nor is there any explosion shown occurring inside the cinema. \textbf{FALSE} \textcolor{darkred}{\textit{(Ground-truth: TRUE)}}\\\\
\textbf{Prediction on Fib}: The video montage shows scenes from Alfred Hitchcock's ``Sabotage.'' The primary bomb plot shown results in an explosion on a bus around 1:46. Later in the montage, around 2:17-2:19, the Professor (the bomb maker) is seen suspiciously entering the cinema building. This implies he might be carrying a second bomb and intending to cause another explosion. However, the video montage ends shortly after this, at 2:23, without showing what happens to the Professor or if there was indeed a second bomb that was ``about to explode.'' Crucially, the video does *not* show this second bomb failing to explode or being prevented from exploding. Since the video does not provide any information about the outcome of this potential second bomb (neither showing it explode nor showing it *not* explode), the assertion ``but it didn't'' is not supported by the available video content. Therefore, the statement cannot be confirmed as true based on the provided video. \textbf{FALSE} \textcolor{darkred}{\textit{(Ground-truth: FALSE)}}\\\\

\bottomrule
\end{tabular}
\caption{\textbf{Illustrative Example 2 from ``Sabotage (Hitchcock)''}. Gemini predicts the pair of claims incorrectly in every modality setting.}
\label{tab:sabotage-example2-part2}
\end{table*}

\begin{table*}[t]
\centering
\begin{tabular}{p{0.94\linewidth}}
\toprule
\textbf{Example 2 (Part 3) - Movie: \textit{"Sabotage"}} \\
\midrule
\\
\colorbox{lightred}{\textit{Video w/ subs setting}}\\\\
\textbf{Prediction on Fact}: The first act of sabotage was the power station blackout. The second planned act of sabotage, referred to as ``a parcel of fireworks'', was to be left in the cloakroom at Piccadilly Circus underground station [...] This parcel, carried by Stevie on a bus, explodes on the bus at 1:45. A newspaper headline later confirms ``WEST END BOMB SENSATION. Fully-laden bus was blown up in Piccadilly this afternoon''. [...] An explosion then occurs at the cinema, as described by an officer: ``The roof's gone. The whole of the back of the place - clean blown off''. This is a separate, third bomb event. Therefore, the second bomb (the one Stevie was carrying) exploded on the bus, not in the cinema. The explosion in the cinema was a distinct event caused by a different bomb wielded by the bird shop owner. \textbf{FALSE} \textcolor{darkred}{\textit{(Ground-truth: TRUE)}}\\\\
\textbf{Prediction on Fib}:
The statement claims: ``There was a second bomb about to explode, but it didn't.'' Let's analyze the bombs in the film: 1.  The first act of sabotage is the power station blackout. While not explicitly shown as a bomb explosion, it is an act of ``wilful destruction.'' 2.  The second major plot device is the bomb intended for Piccadilly Circus on Lord Mayor's Show day. Verloc gives this bomb to Stevie. This bomb explodes prematurely on the bus, killing Stevie and many others [...]. So this bomb *did* explode. 3.  The third instance involving a bomb is at the cinema, where the Professor (the bomb maker) is cornered by the police. [..] This bomb *does* explode [...]. In all instances where a bomb is primed or ready for its purpose within the narrative, it either explodes (bus bomb, Professor's bomb in the cinema) or fulfills its sabotage (power station). There is no scene in the film where a bomb is ``about to explode'' and then ``didn't.'' The bombs that are set up to go off, do go off. Therefore, the statement is false. \textbf{FALSE} \textcolor{darkred}{\textit{(Ground-truth: FALSE)}}\\

\bottomrule
\end{tabular}
\caption{\textbf{Illustrative Example 2 from ``Sabotage (Hitchcock)''}. Gemini predicts the pair of claims incorrectly in every modality setting.}
\label{tab:sabotage-example2-part3}
\end{table*}

\definecolor{darkgreen}{rgb}{0,0.45,0}
\definecolor{darkred}{rgb}{0.7,0,0}
\definecolor{lightred}{rgb}{1,0.7,0.7}


\begin{table*}[t]
\centering
\begin{tabular}{p{0.94\linewidth}}
\toprule
\textbf{Example 3 (Part 1) - Movie: \textit{"Scandal Sheet"}} \\
\midrule

\begin{minipage}{0.3\linewidth}
    \centering
    \includegraphics[width=\linewidth]{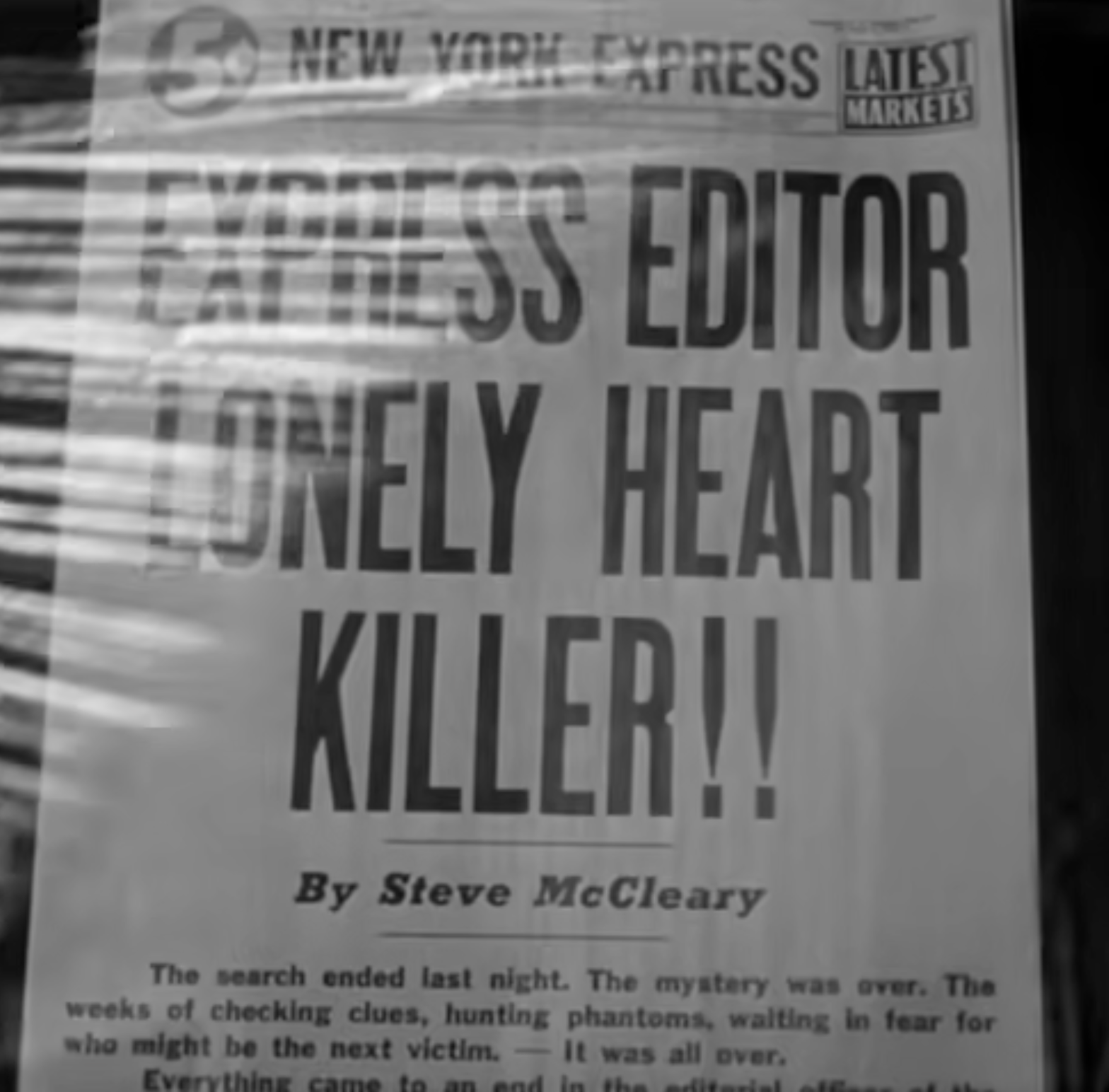}
    {\footnotesize News about the killer}
\end{minipage}\hfill
\begin{minipage}{0.3\linewidth}
    \centering
    \includegraphics[width=\linewidth]{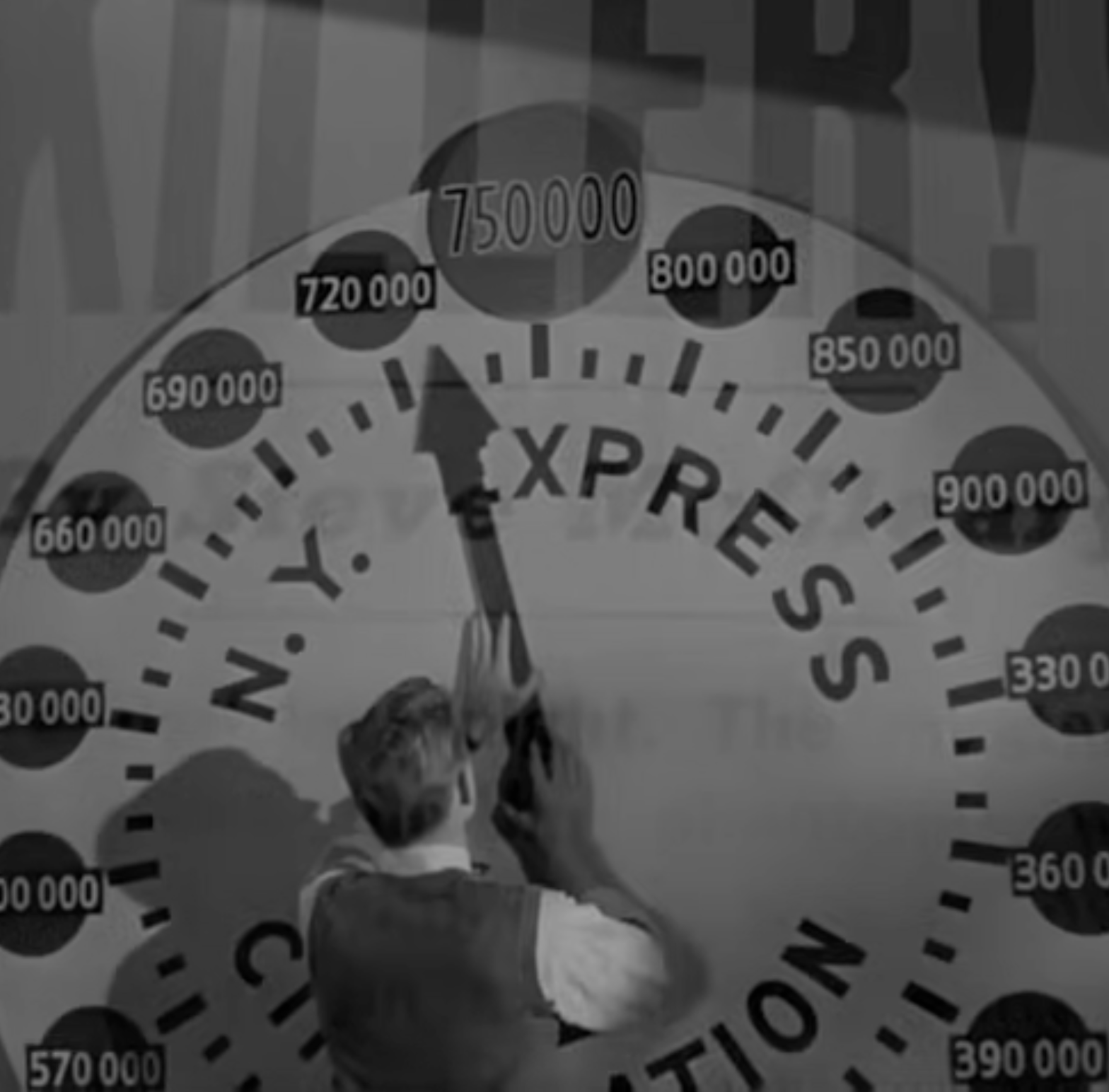}
    {\footnotesize Sales increasing..}
\end{minipage}\hfill
\begin{minipage}{0.3\linewidth}
    \centering
    \includegraphics[width=\linewidth]{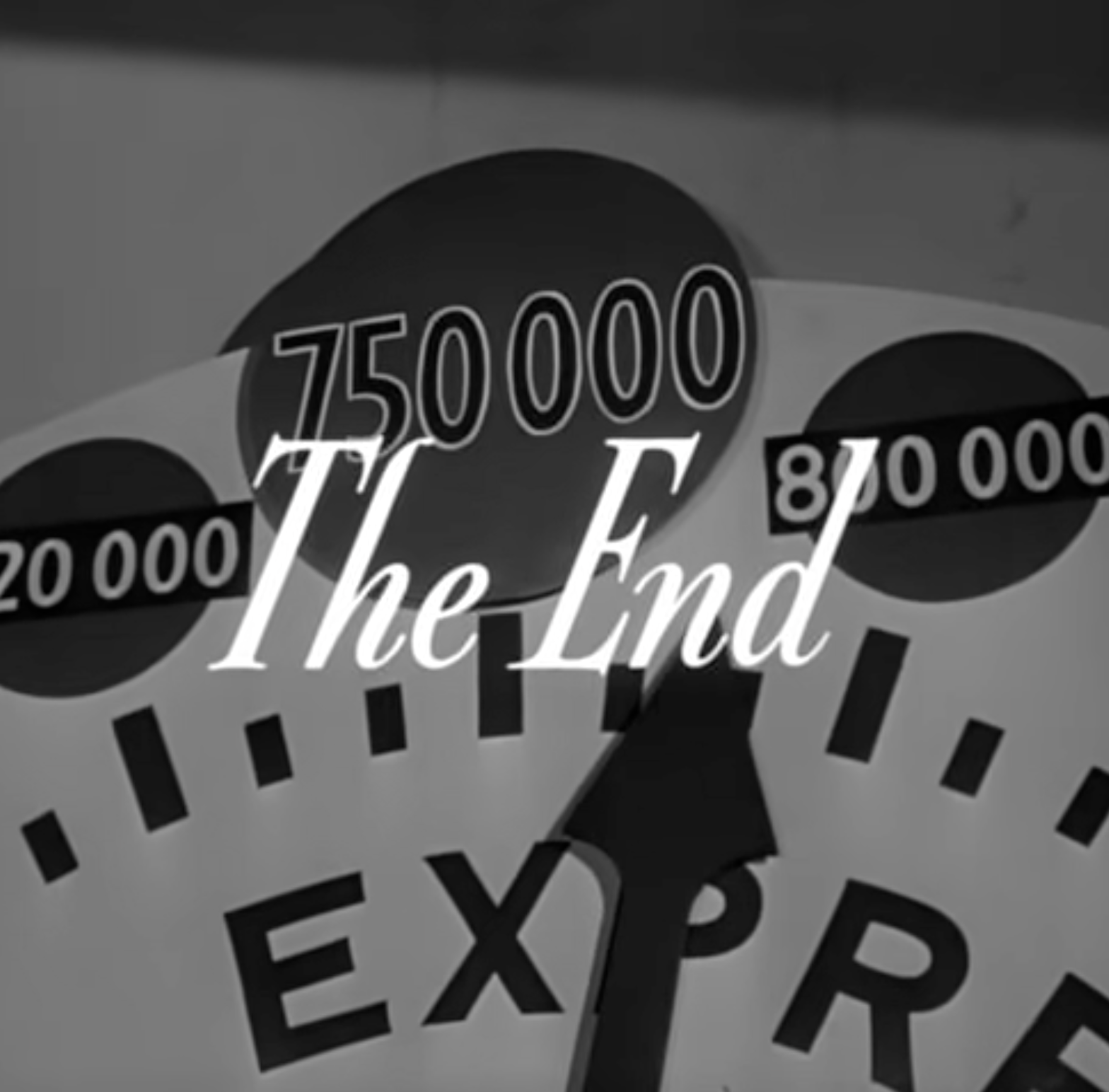}
    {\footnotesize Sales increasing..}
\end{minipage}\hfill

\\\\

\textbf{Fact} (\cmark): The sales of the newspaper increased after it was revealed that the editor was the killer. \\

\textbf{Fib} (\xmark): The sales of the newspaper decreased after it was revealed that the editor was the killer. \\\\

\textbf{Granularity}: Single-scene \\\\

\textbf{Context (\textcolor{darkred}{Spoiler alert !})}: In the film’s climax, Steve McCleary discovers that his editor-in-chief, Mark Chapman, is the “Lonely Hearts Killer.” After Chapman is exposed and confronted, McCleary—now acting as editor—publishes the full story, revealing Chapman’s true identity and crimes. Following the revelation, the newspaper’s sales surge dramatically.
\textit{This pair of claims relies primarily on visual information. Gemini predicts them incorrectly when given only a single modality, but when provided with both the transcript and the video, it correctly identifies the pair of claims.}\\\\

\textbf{Gemini's Predictions}:\\\\
\colorbox{lightred}{\textit{Subs only setting}} \\\\

\textbf{Prediction on Fact}:  
The transcript ends with Mark Chapman, the editor, being apprehended as the killer. He tells the reporter McCleary to ``Write it up big, kid. It'll sell a lot of papers.'' This indicates an expectation that the story will boost sales, but the transcript concludes before any information about the actual sales figures *after* this revelation is provided. Therefore, the statement that sales *increased* after the revelation cannot be verified from the given text. \textbf{FALSE} \textcolor{darkred}{\textit{(Ground-truth: TRUE)}}\\\\

\textbf{Prediction on Fib}:  
The transcript ends with Mark Chapman, the editor, being apprehended as the killer. His last words are, \"Write it up big, kid. It'll sell a lot of papers.\" This is a prediction about future sales, not a statement of what actually happened to the sales. The transcript provides no information about the newspaper's sales figures *after* it was revealed that the editor was the killer. Therefore, the statement cannot be confirmed as true based on the provided text. \textbf{FALSE} \textcolor{darkred}{\textit{(Ground-truth: FALSE)}}\\\\

\bottomrule
\end{tabular}
\caption{\textbf{Illustrative Example 3 from ``Scandal Sheet''}. The pair of claims relies primarily on visual information. Gemini predicts them incorrectly when given only a single modality, but when provided with both the transcript and the video, it correctly identifies the pair of claims.} 
\label{tab:scandal-example3-part1}
\end{table*}

\begin{table*}[t]
\centering
\begin{tabular}{p{0.94\linewidth}}
\toprule
\textbf{Example 3 (Part 2) - Movie: \textit{"Scandal Sheet"}} \\
\midrule
\\
\colorbox{lightred}{\textit{Video only setting}}\\\\
\textbf{Prediction on Fact}: The video shows a chart of the ``NET PAID CIRCULATION NEW YORK EXPRESS'' at various points. Initially, at 0:10, the circulation is shown with an upward trend. After the editor is revealed to be the killer and is subsequently shot (around 2:26-2:28), the final shot of the movie (2:29) shows a newspaper headline ``KILLER!!'' with the byline ``By Steve McCleary'' superimposed over the same circulation chart. At this point, the line on the chart indicating sales has reached its highest point, implying a significant increase in sales coinciding with or immediately following the revelation and capture of the killer editor. \textbf{TRUE} \textcolor{darkred}{\textit{(Ground-truth: TRUE)}}\\\\
\textbf{Prediction on Fib}: The movie shows a chart of the ``New York Express'' net paid circulation at 0:10, indicating an upward trend in sales from October to September, reaching 650,000. Later, after the editor (the man standing by the chart at 0:10) is revealed as the killer and is shot (around 2:27), the final shot at 2:29 shows a newspaper headline ``KILLER!!'' superimposed over the *same* circulation chart. The numbers and the upward trend on this chart are identical to the one shown at 0:10. The movie does not provide any information or visual evidence suggesting that the newspaper's sales decreased after the editor was revealed as the killer. The chart displayed at the end is the same one that showed increasing sales. \textbf{TRUE} \textcolor{darkred}{\textit{(Ground-truth: FALSE)}}\\\\
\colorbox{lightred}{\textit{Video w/ subs setting}}\\\\
\textbf{Prediction on Fact}: The statement's truthfulness hinges on whether newspaper sales increased after the revelation that the editor, Mark Chapman, was the killer. 1. At 3:40, Mark Chapman discusses the newspaper's circulation, showing a graph and mentioning that when circulation reaches 750,000, a big bonus will be paid. The graph shows circulation rising, with the last point around June/July showing approximately 620,000-650,000. 2. The revelation that Mark Chapman is the killer unfolds towards the end of the film. 3. At 59:39, the newspaper headline ``EXPRESS EDITOR LONELY HEART KILLER!!'' by Steve McCleary is shown. 4. Immediately following this (59:45 - 59:49), a large circulation meter for the ``N.Y. EXPRESS'' is shown. A hand physically moves the needle on the meter, stopping at 750,000. ``The End'' is then superimposed over this image. This sequence strongly implies that the story revealing the editor as the killer led to the circulation reaching 750,000. Earlier, Chapman had also told McCleary to write the story big because ``It'll sell a lot of papers'' (59:06). The visual of the circulation hitting 750,000 immediately after the headline about the editor being the killer is the final image before ``The End,'' suggesting this was the peak sales driven by this sensational story. \textbf{TRUE} \textcolor{darkred}{\textit{(Ground-truth: TRUE)}}\\\\
\textbf{Prediction on Fib}:
At the end of the movie, after the editor, Mark Chapman, is revealed as the killer and the headline ``EXPRESS EDITOR LONELY HEART KILLER!!'' is shown (59:40-59:43), the circulation meter for the New York Express is shown. A hand moves the needle to 750,000 (59:45). Earlier in the film (around 4:24), it was mentioned that reaching 750,000 in circulation would trigger a big bonus. The film ends with the circulation at this high number, indicating that sales reached this target *after* the revelation about the editor. There is no indication that sales decreased; in fact, the implication is that this sensational story would boost sales, as McCleary says, ``Write it up big, kid. It'll sell a lot of papers.'' (59:14-59:16). \textbf{FALSE} \textcolor{darkred}{\textit{(Ground-truth: FALSE)}}\\
\bottomrule
\end{tabular}
\caption{\textbf{Illustrative Example 3 from ``Scandal Sheet''}. The pair of claims relies primarily on visual information. Gemini predicts them incorrectly when given only a single modality, but when provided with both the transcript and the video, it correctly identifies the pair of claims.}
\label{tab:scandal-example3-part2}
\end{table*}

\definecolor{darkgreen}{rgb}{0,0.45,0}
\definecolor{darkred}{rgb}{0.7,0,0}
\definecolor{lightred}{rgb}{1,0.7,0.7}


\begin{table*}[t]
\centering
\begin{tabular}{p{0.94\linewidth}}
\toprule
\textbf{Example 4 (Part 1) - Movie: \textit{"Tight Spot"}} \\
\midrule

\begin{center}
\includegraphics[width=0.45\linewidth]{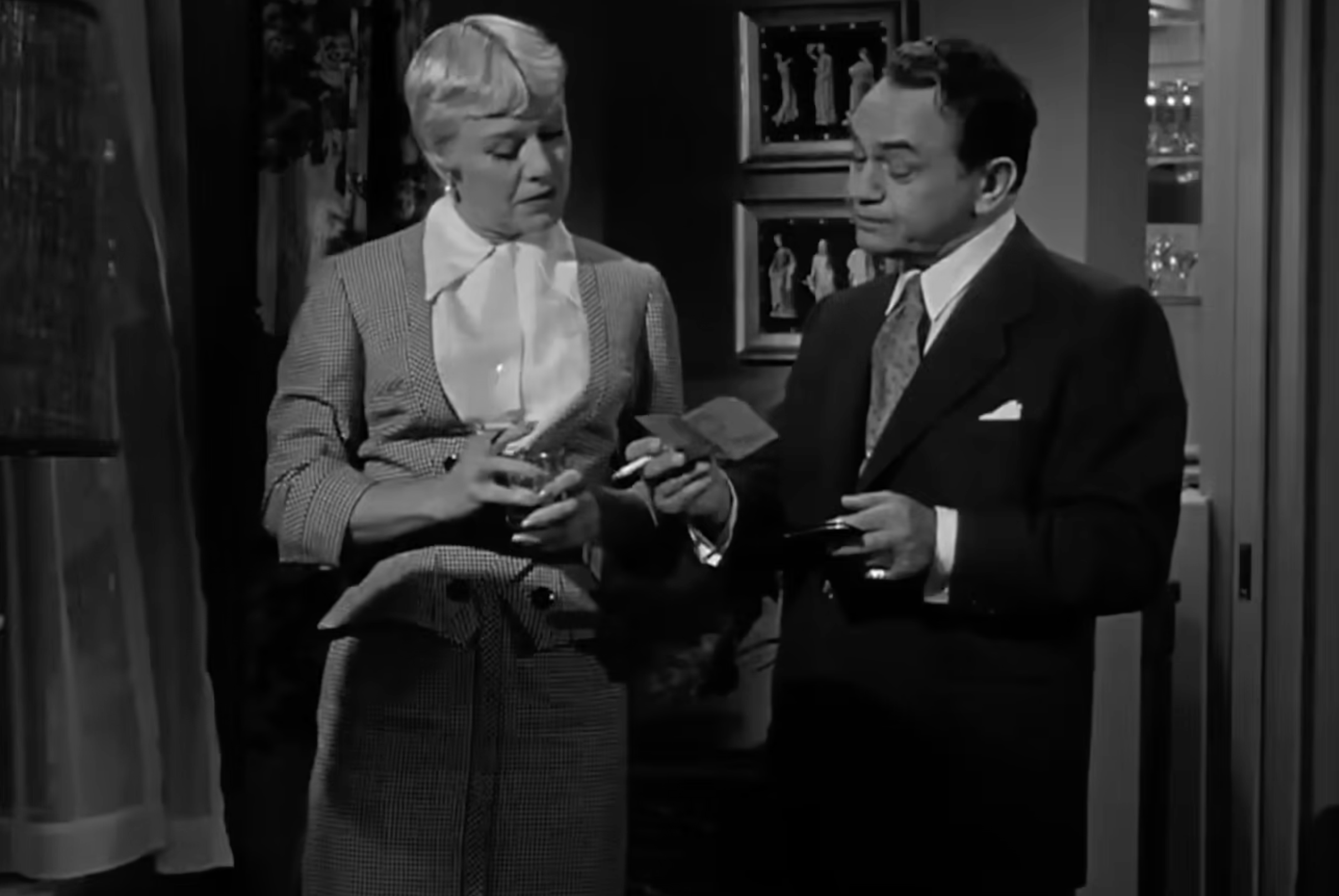}
\hfill
\includegraphics[width=0.40\linewidth]{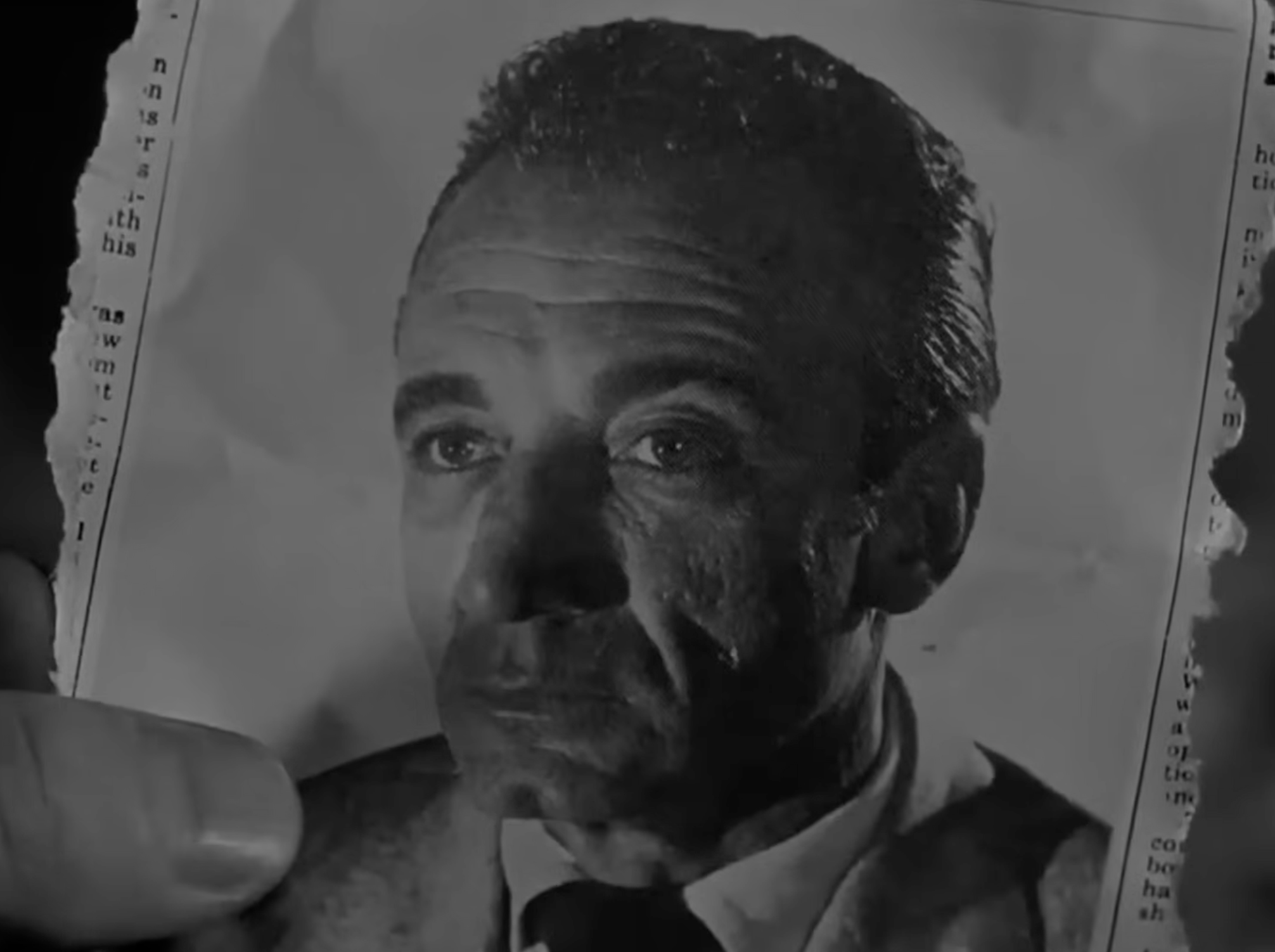}
\end{center} 

\textbf{Fact} (\cmark):  When Hallet shows Conley a photograph of an individual who was on the yacht, she immediately recognizes the person. \\
\textbf{Fib} (\xmark): When Hallet shows Conley a photograph of an individual who was on the yacht, she claims not to recognize the person. \\\\

\textbf{Granularity}: Single-scene
\\\\
\textbf{Context (\textcolor{darkred}{Spoiler alert !}):}  This pair of claims to be answered correctly requires \textit{both} the \textit{visual} and \textit{text} cues. Hallet shows Conley a photograph (visual information) and asks whether she recognizes the person (Senior Maracelli). Conley recognizes the person. \textit{When the model is given only the subtitles, it misclassifies both the fact and the fib (as visual cues are absent). In contrast, when provided with both the video and the subtitles, it predicts each one correctly.}\\\\

\textbf{Gemini's Predictions:} \\\\
\colorbox{lightred}{\textit{Subs only setting}}\\\\
\textbf{Prediction on Fact}: The statement claims that ``When Hallet shows Conley a photograph of an individual who was on the yacht, she immediately recognizes the person.'' We need to check two things in the transcript: 1. Does Hallett show Conley a photograph of an individual from the yacht? 2. If so, does Conley immediately recognize the person? In the transcript, the relevant exchange is: HALLETT: ``And at one of these islands, this man came aboard the yacht.'' CONLEY: ``Signor Maracelli. The Italian garden expert. He taught me to say his name himself.'' [...] \textit{The transcript does not use the word ``photograph'' or describe the action of showing one in this specific interaction.} Since the statement's premise that Hallett shows Conley a photograph is not explicitly confirmed by the transcript, a part of the statement is not verifiably true based on the provided text. \textbf{FALSE} \textcolor{darkred}{\textit{(Ground-truth: TRUE)}}\\\\
\textbf{Prediction on Fib}: When Mr. Hallett mentions Signor Maracelli coming aboard the yacht, Conley immediately identifies him as ``The Italian garden expert'' and adds, ``He taught me to say his name himself.'' This indicates she clearly recognizes the person Hallett is referring to (and presumably showing a photograph of, though the act of showing the photo isn't explicitly stated, Hallett's line ``this man came aboard the yacht'' implies he's identifying someone specific). She does not claim not to recognize him. \textbf{FALSE} \textcolor{darkred}{\textit{(Ground-truth: FALSE)}}\\

\bottomrule
\end{tabular}
\caption{\textbf{Illustrative Example 4 from ``Tight Spot''}. Gemini fails to predict both claims correctly when given only the subtitles or only the visual cues. However, when provided with both modalities together, it correctly predicts both claims.}
\label{tab:tightspot-example4-part1}
\end{table*}

\begin{table*}[t]
\centering
\begin{tabular}{p{0.94\linewidth}}
\toprule
\textbf{Example 4 (Part 2) - Movie: \textit{"Tight Spot"}} \\
\midrule
\\
\colorbox{lightred}{\textit{Video only setting}}\\\\
\textbf{Prediction on Fact}: The statement asks whether Conley immediately recognizes the person in the photograph shown to her by Hallet. In the video, at 00:30, Hallet takes out a photograph. At 00:31, he shows it to Conley. Conley looks down at the photo, then at 00:32, she looks up at Hallet with an expression that seems more contemplative or questioning rather than one of instant, clear recognition. She then glances back at the photo before looking at Hallet again and starting to speak at 00:33. Her reaction involves a pause and a sequence of looks, which does not convey *immediate* recognition. ``Immediate'' suggests an instantaneous, reflexive understanding, which is not evident in her more measured response. \textbf{FALSE} \textcolor{darkred}{\textit{(Ground-truth: TRUE)}}\\\\
\textbf{Prediction on Fib}: The statement claims that when Hallet shows Conley a photograph of an individual from a yacht, Conley claims not to recognize the person. Observing the video, there is a scene from approximately 0:29 to 0:36 where a man (presumably Hallet, played by Edward G. Robinson) shows a photograph to a woman (presumably Conley, played by Ginger Rogers). The woman looks at the photograph (0:31-0:32). However, the video segment does not include any audible dialogue or clear visual cue (like shaking her head definitively or mouthing ``I don't know'') from the woman that would confirm she ``claims not to recognize the person.'' Her facial expression is somewhat pensive or surprised but not an explicit denial of recognition. She then proceeds to make a phone call (0:34). The context that the photograph is of ``an individual who was on the yacht'' is also not explicitly provided by the visual information in this clip. Since the crucial part of the statement - her claiming not to recognize the person - is not depicted or audible in the provided video evidence, the statement cannot be verified as true based on the available information. \textbf{FALSE} \textcolor{darkred}{\textit{(Ground-truth: FALSE)}}\\\\

\colorbox{lightred}{\textit{Video w/ subs setting}}\\\\
\textbf{Prediction on Fact}: When Hallett shows Conley the photograph (around 00:40), he identifies the man as ``Signor Maracelli.'' Conley immediately confirms recognition by saying, ``The Italian garden expert. He taught me to say his name himself.'' This demonstrates she knew who the person was. \textbf{TRUE} \textcolor{darkred}{\textit{(Ground-truth: TRUE)}}\\\\
\textbf{Prediction on Fib}: When Mr. Hallett mentions Signor Maracelli coming aboard the yacht, Conley immediately identifies him as ``The Italian garden expert'' and adds, ``He taught me to say his name himself.'' This indicates she clearly recognizes the person Hallett is referring to (and presumably showing a photograph of, though the act of showing the photo isn't explicitly stated, Hallett's line ``this man came aboard the yacht'' implies he's identifying someone specific). She does not claim not to recognize him. \textbf{FALSE} \textcolor{darkred}{\textit{(Ground-truth: FALSE)}}\\

\bottomrule
\end{tabular}
\caption{\textbf{Illustrative Example 4 from ``Tight Spot''}. Gemini fails to predict both claims correctly when given only the subtitles or only the visual cues. However, when provided with both modalities together, it correctly predicts both claims.}
\label{tab:tightspot-example4-part2}
\end{table*}

\definecolor{darkgreen}{rgb}{0,0.45,0}
\definecolor{darkred}{rgb}{0.7,0,0}
\definecolor{lightred}{rgb}{1,0.7,0.7}


\begin{table*}[t]
\centering
\begin{tabular}{p{0.94\linewidth}}
\toprule
\textbf{Example 5 (Part 1) - Movie: \textit{"Tight Spot"}} \\
\midrule

\begin{minipage}{0.33\linewidth}
    \centering
    \includegraphics[width=\linewidth]{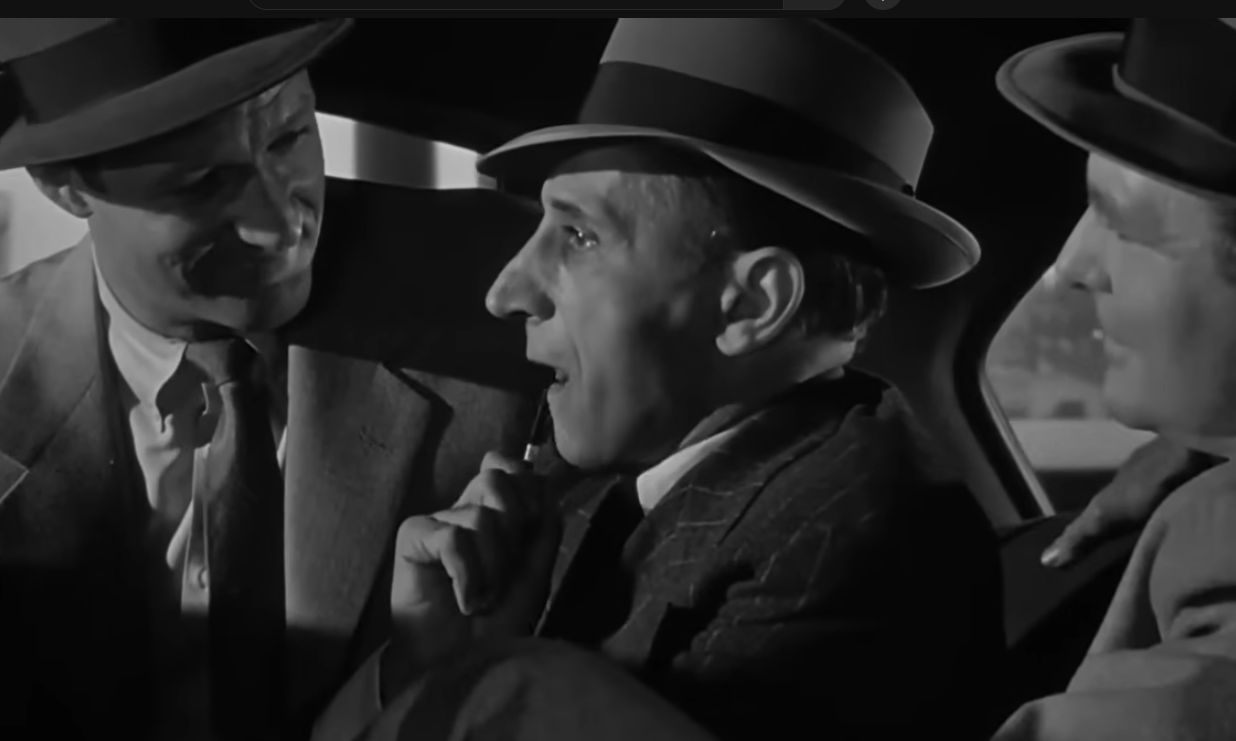}
    {\footnotesize Pete in taxi}
\end{minipage}\hfill
\begin{minipage}{0.33\linewidth}
    \centering
    \includegraphics[width=\linewidth]{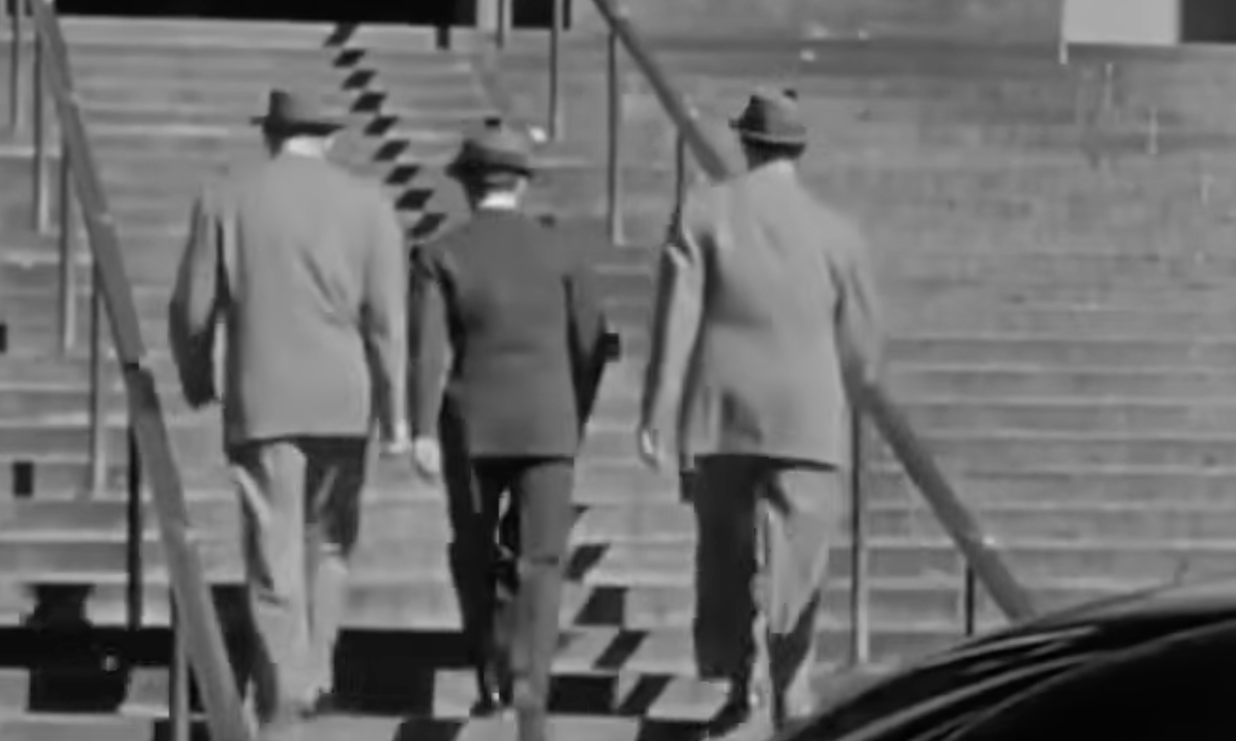}
    {\footnotesize Walking the stairs of the church}
\end{minipage}\hfill
\begin{minipage}{0.33\linewidth}
    \centering
    \includegraphics[width=\linewidth]{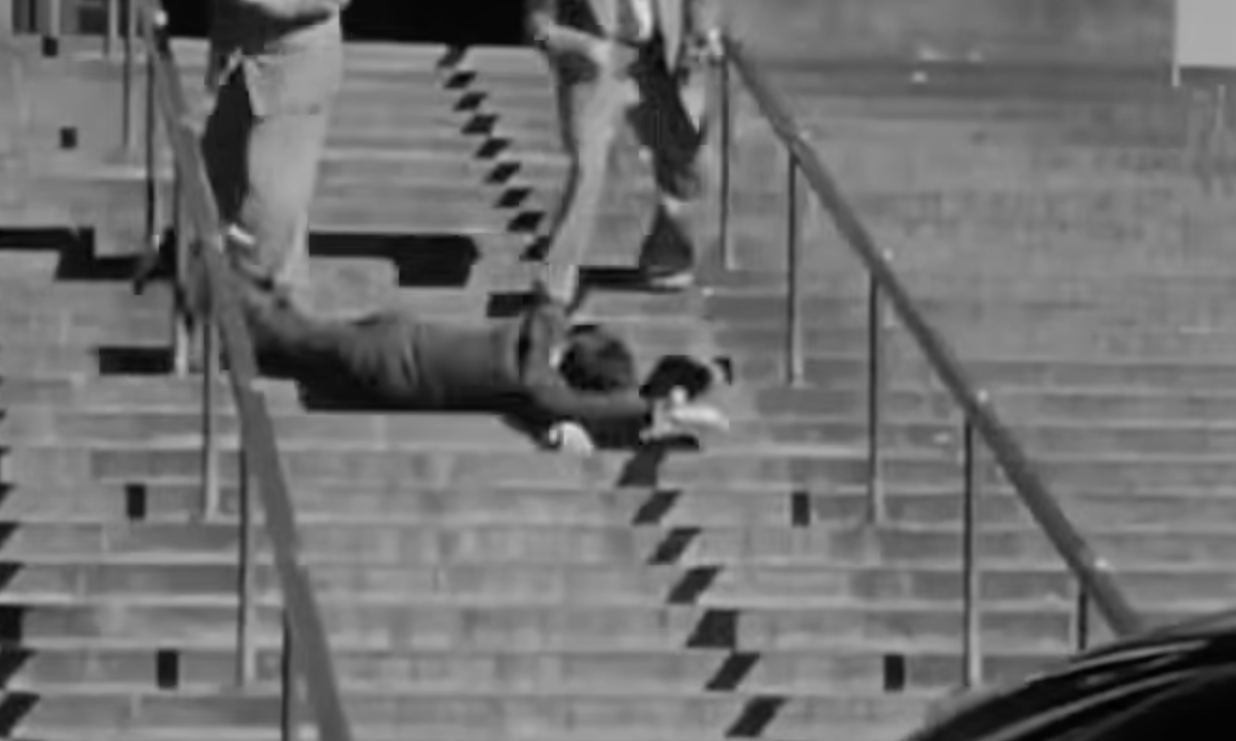}
    {\footnotesize Pete being shot}
\end{minipage}

\noindent \\
\textbf{Fact} (\cmark): Pete Tinelli was shot. \\

\textbf{Fib} (\xmark): Pete Tinelli was poisoned. \\\\

\textbf{Granularity}: Multi-scene
\\\\
\textbf{Context (\textcolor{darkred}{Spoiler alert !})}: Early in the film, we are introduced to Pete during a scene in which he rides in a taxi with two other men. Later, the group  arrives in front of a church, and as they walk up the steps, Pete is suddenly shot while the other two men look around, trying to identify the shooter. \textit{This pair of claims requires both visual and text cues to be resolved correctly, as the model needs to understand who is Pete (in the first scene) and connect it with his murder (being shot). Gemini fails to predict correctly the pair of claims in every modality setting}.\\\\

\textbf{Gemini's Predictions:} \\\\
\colorbox{lightred}{\textit{Subs only setting}}\\\\
\textbf{Prediction on Fact}: The transcript states that ``Pete's dead'' and that ``He was murdered by the Costain mob.'' [..] However, the transcript does not specify the method of murder. It does not explicitly say that Pete Tonelli was shot. Therefore, because the specific detail of ``shot'' is not confirmed in the transcript, the statement is false. \textbf{FALSE} \textcolor{darkred}{\textit{(Ground-truth: TRUE)}}\\\\
\textbf{Prediction on Fib}: The transcript states, ``Pete's dead... He was murdered by the Costain mob.'' It does not specify the method of murder. Since the transcript doesn't mention poisoning as the cause of Pete Tonelli's death, and only confirms he was murdered, the statement that he was poisoned cannot be confirmed as true based on the provided text. \textbf{FALSE} \textcolor{darkred}{\textit{(Ground-truth: FALSE)}}\\\\

\bottomrule
\end{tabular}
\caption{\textbf{Illustrative Example 5 from ``Tight Spot''}. Gemini fails to predict the pair of claims correctly, in every modality setting.}
\label{tab:tightspot-example5-part1}
\end{table*}

\begin{table*}[t]
\centering
\begin{tabular}{p{0.94\linewidth}}
\toprule
\textbf{Example 5 (Part 2) - Movie: \textit{"Tight Spot"}} \\
\midrule
\\
\colorbox{lightred}{\textit{Video only setting}}\\\\
\textbf{Prediction on Fact}: A man is visibly shot around the 1:57-1:58 mark in the video. In this scene, a man is held by two others, another man (resembling Edward G. Robinson) nods, a gunshot sound is heard, and the man being held slumps. However, throughout the provided video clip, there is no dialogue, on-screen text, or any other information that identifies the man who was shot as ``Pete Tinelli.'' [...] \textbf{FALSE} \textcolor{darkred}{\textit{(Ground-truth: TRUE)}}\\\\
\textbf{Prediction on Fib}: The video contains several scenes that lead to the conclusion that Pete Tinelli was poisoned. At 1:58, a group of men are shown in what appears to be a garage, and one man is forcibly made to drink something. He subsequently collapses at 2:01. Later, at 2:05, a woman asks, ``What happened to Pete Tinelli?'' and is told, ``He won't bother you anymore.'' Shortly after, at 2:08, one man informs another, ``Pete Tinelli's been taken care of.'' When asked ``How?'', the response is ``Poison.'' This sequence of events and dialogue directly confirms that Pete Tinelli was poisoned.\textbf{TRUE} \textcolor{darkred}{\textit{(Ground-truth: FALSE)}}\\\\
\colorbox{lightred}{\textit{Video w/ subs setting}}\\\\
\textbf{Prediction on Fact}: The provided transcript states at 00:54-00:55 that ``Pete's dead'' and ``He was murdered by the Costain mob.'' However, neither the transcript nor the accompanying video footage specifies the method of his murder. The statement ``Pete Tinelli was shot'' asserts a specific cause of death (being shot) which is not explicitly mentioned or shown in the provided materials. Since the method of murder is not confirmed, we cannot ascertain that he was shot based solely on the given information. \textbf{FALSE} \textcolor{darkred}{\textit{(Ground-truth: TRUE)}}\\\\
\textbf{Prediction on Fib}: The transcript states at 00:42, ``He was murdered by the Costain mob.'' While this confirms Pete Tonelli was murdered, it does not specify the method of murder. The statement claims he was poisoned. Since the available information does not mention poisoning as the cause of death, the specific claim that he was poisoned cannot be verified as true from the provided movie and transcript. \textbf{FALSE} \textcolor{darkred}{\textit{(Ground-truth: FALSE)}}\\

\bottomrule
\end{tabular}
\caption{\textbf{Illustrative Example 5 from ``Tight Spot''}. Gemini fails to predict the pair of claims correctly, in every modality setting.}
\label{tab:tightspot-example5-part2}
\end{table*}

\definecolor{darkgreen}{rgb}{0,0.45,0}
\definecolor{darkred}{rgb}{0.7,0,0}
\definecolor{lightred}{rgb}{1,0.7,0.7}

\begin{table*}[t]
\centering
\begin{tabular}{p{0.94\linewidth}}
\toprule
\textbf{Example 6 (Part 1) - Movie: \textit{``A Bucket of Blood''}} \\
\midrule

\begin{minipage}{0.23\linewidth}
    \centering
    \includegraphics[width=\linewidth]{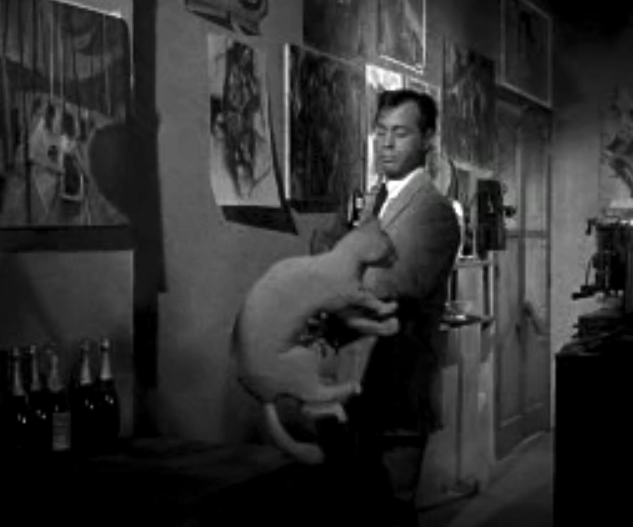}
    {\footnotesize Holding the cat sculpture}
\end{minipage}\hfill
\begin{minipage}{0.23\linewidth}
    \centering
    \includegraphics[width=\linewidth]{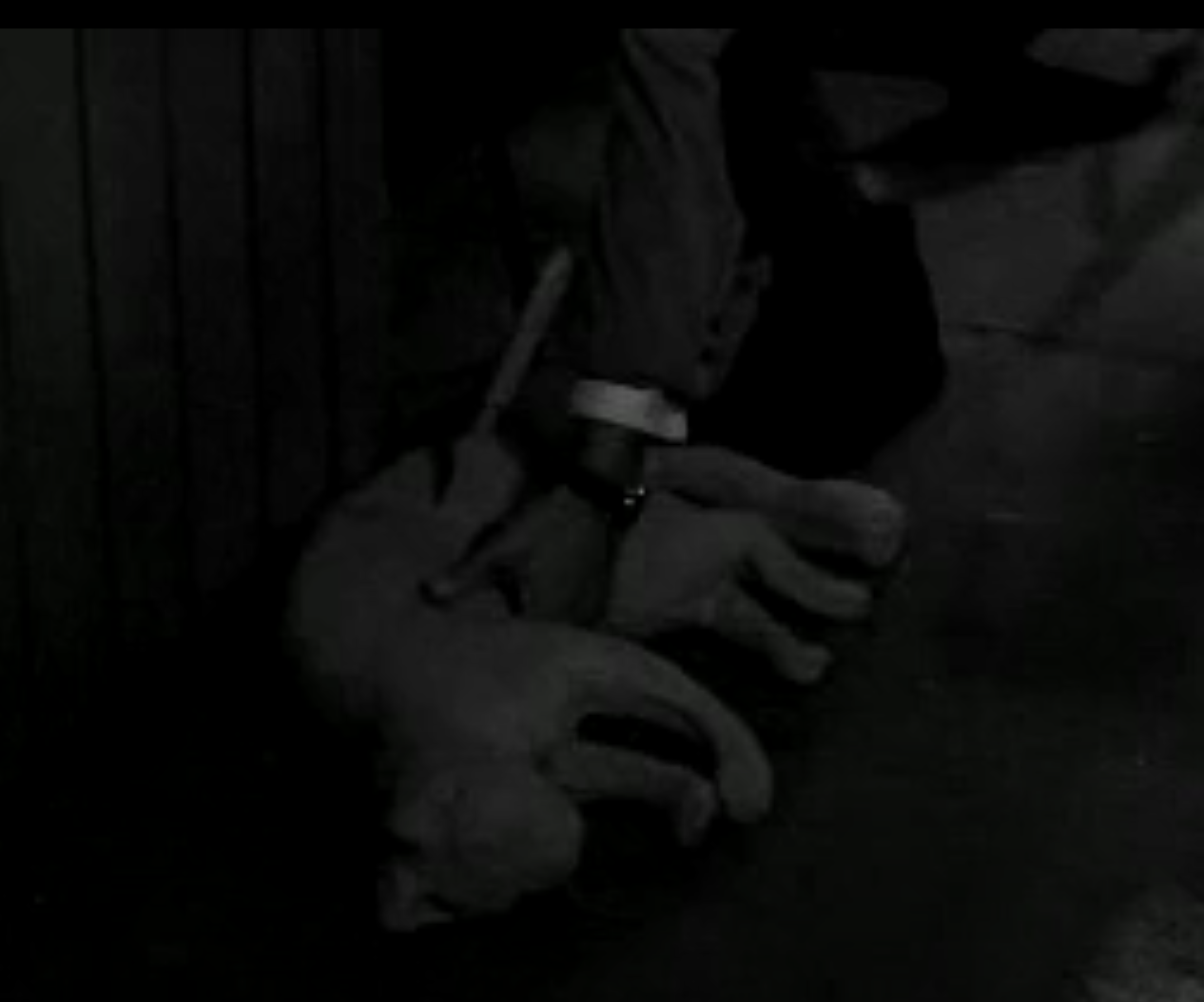}
    {\footnotesize Drops it}
\end{minipage}\hfill
\begin{minipage}{0.26\linewidth}
    \centering
    \includegraphics[width=\linewidth]{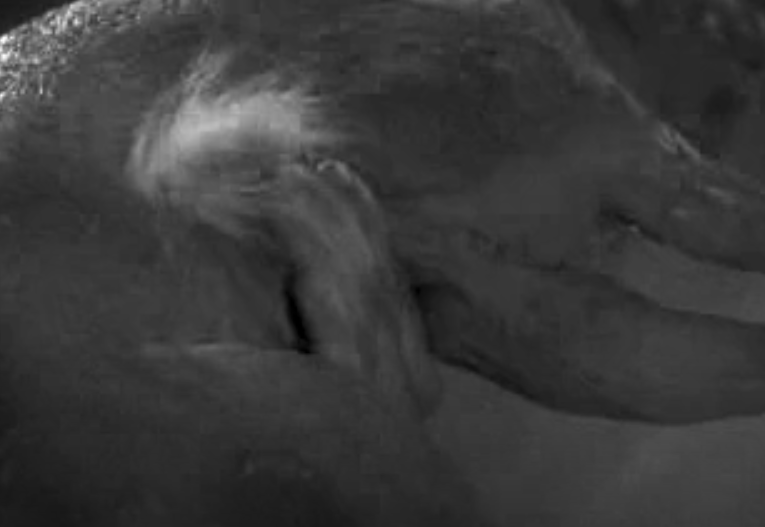}
    {\footnotesize The damage}
\end{minipage}\hfill
\begin{minipage}{0.25\linewidth}
    \centering
    \includegraphics[width=\linewidth]{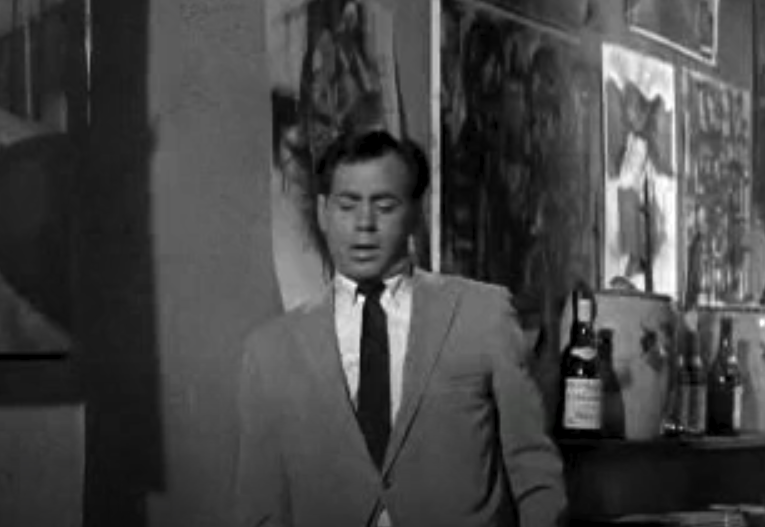}
    {\footnotesize Moment of realization }
\end{minipage}\hfill

\noindent \\
\textbf{Fact} (\cmark): The gallery owner dropped the cat sculpture, causing minor damage.\\\\

\textbf{Fib} (\xmark): The gallery owner dropped the cat sculpture, but it remained intact.\\\\

\textbf{Granularity}: Single-scene
\\\\
\textbf{Context (\textcolor{darkred}{Spoiler alert !})}: When the gallery owner accidentally drops the cat sculpture in the cafe and sees real flesh beneath the clay, he realizes that the sculpture is not a realistic imitation but an actual dead cat covered in clay. \textit{Gemini does not correctly predict the claim pair, as it fails to recognize that the sculpture was damaged.} \\\\

\textbf{Gemini's Predictions:} \\\\
\colorbox{lightred}{\textit{Subs-only setting}}\\\\
\textbf{Prediction on Fact}: The transcript details the creation of the ``Dead Cat'' sculpture by Walter and its reception at the Yellow Door, owned by Leonard De Santis. Leonard agrees to display the cat sculpture. Later, a customer takes the sculpture home to show his wife and then makes an offer to buy it for \$500. Leonard tells this customer, ``Someone has the cat just now, but I'll have him back in a few days. Then you can have it for the \$500.'' There is no mention in the transcript of Leonard, the gallery/club owner, dropping the cat sculpture or of it sustaining any damage. \textbf{FALSE}\\\\
\textbf{Prediction on Fib}: The transcript mentions that Mr. De Santis, the owner of the Yellow Door (where the art is displayed, akin to a gallery owner in this context), arranged for a potential buyer to take the cat sculpture home to show his wife. It states, ``A man wanted to pay me \$100 for the cat. In fact, he's taking it home to show to his wife.'' Later, the same man calls back to buy it, and Mr. De Santis says, ``Someone has the cat just now, but I'll have him back in a few days.'' There is no mention in the transcript of Mr. De Santis or anyone else dropping the cat sculpture. Therefore, the premise that the gallery owner dropped it is false. \textbf{FALSE}\\
\bottomrule
\end{tabular}
\caption{\textbf{Illustrative Example 6 from the movie: ``A Bucket of Blood''}. Gemini fails to predict the pair of claims correctly, in every modality setting.}
\label{tab:sabotage-example6-part1}
\end{table*}

\begin{table*}[t]

\centering
\begin{tabular}{p{0.94\linewidth}}
\toprule
\textbf{Example 6 (Part 2) - Movie: \textit{``A Bucket of Blood''}} \\
\midrule
\\
\colorbox{lightred}{\textit{Video-only setting}}\\\\
\textbf{Prediction on Fact}: The statement is: ``The gallery owner dropped the cat sculpture, causing minor damage.'' To verify this, we need to identify a ``gallery owner,'' a ``cat sculpture,'' an act of ``dropping'' by the gallery owner, and resulting ``minor damage.'' 1.  **Gallery Owner:**    *   Leonard de Santis is the owner of "The Yellow Dog" cafe, which also functions as an art gallery where poetry is read and art is displayed [...]  Walter Paisley creates a sculpture of a cat by encasing his landlady's dead cat in clay (0:16-0:20). This is the most prominent cat sculpture.    *   At 0:29, Leonard and Carla are seen holding other, more stylized white cat sculptures, indicating these might be common art pieces in their circle. 3.  **Dropped the cat sculpture, causing minor damage:**     *   At 0:29, Leonard (the cafe/gallery owner) is shown holding one of the stylized white cat sculptures. He does not drop it. [...] No scene in the provided video shows a character identifiable as "the gallery owner" dropping a cat sculpture and causing minor damage. The closest event involves a police officer tapping Walter's cat sculpture, causing a tiny chip. Therefore, the statement is false. \textbf{FALSE}\\\\
\textbf{Prediction on Fib}: The statement says: ``The gallery owner dropped the cat sculpture, but it remained intact.'' Let's break this down: 1.  **Identity of the ``gallery owner''**: In the movie clip, the gallery owner is Maxwell Brock, often seen wearing a beret or a suit, and he runs ``The Yellow Door'' gallery. 2.  **Did the gallery owner drop a cat sculpture?**:    *   At 0:27, the main character, Walter Paisley (a busboy who becomes an artist), is carrying a large abstract sculpture (not a cat) when he is bumped by Carla. Walter drops this sculpture, and it shatters. Walter is not the gallery owner.    *   Walter Paisley creates a sculpture of a cat (by encasing his landlady's dead cat in clay) around 0:22. This sculpture is later handled but not dropped by the gallery owner.    *   At 0:50, Maxwell Brock (the gallery owner) is seen handling a *different* white, stylized cat sculpture. He picks it up, examines it carefully, and places it back down. He does *not* drop this sculpture. Since there is no instance in the provided video where the gallery owner (Maxwell Brock) drops any cat sculpture, the first part of the statement ("The gallery owner dropped the cat sculpture") is false. Therefore, the entire statement is false. \textbf{FALSE}\\\\
\colorbox{lightred}{\textit{Video w/ subs setting}}\\\\
\textbf{Prediction on Fact}: The video and transcript show the gallery owner, Leonard, handling the cat sculpture on a couple of occasions. First, when Walter initially brings it to ``The Yellow Door'' (around 13:36-13:55), Leonard is skeptical but eventually agrees to display it. He takes the sculpture from Walter and places it on a surface without dropping it. Later, after the sculpture gains attention (around 25:05), Leonard is seen picking up the cat sculpture, looking at it, and then placing it back down on a table, again without dropping or damaging it. The transcript contains no mention of the gallery owner dropping the sculpture. \textbf{FALSE}\\\\
\textbf{Prediction on Fib}:
The statement claims that the gallery owner dropped the cat sculpture. Reviewing the video and transcript, at 13:47, Leonard De Santis, the owner of the Yellow Door gallery/cafe, picks up the cat sculpture made by Walter Paisley. Shortly after, at 13:48, Walter Paisley himself is holding the sculpture. At 13:49, Walter Paisley (the artist, not the gallery owner) drops the cat sculpture onto the floor. The sculpture appears to remain intact. Since it was Walter, the artist, and not Leonard, the gallery owner, who dropped the sculpture, the first part of the statement is false. \textbf{FALSE}\\
\bottomrule
\end{tabular}
\caption{\textbf{Illustrative Example 6 from the movie: ``A Bucket of Blood''}. Gemini fails to predict the pair of claims correctly, in every modality setting.}
\label{tab:sabotage-example6-part2}
\end{table*}

\end{document}